\documentclass[1p,11pt]{elsarticle}

\usepackage{lineno}
\usepackage[hidelinks]{hyperref}
\usepackage{setspace}
\usepackage{amsmath}
\usepackage{amsfonts}
\usepackage{subcaption}
\usepackage{siunitx}
\usepackage{graphicx}
\usepackage{mathtools}
\usepackage{caption}
\usepackage{subcaption}
\usepackage{xcolor}
\usepackage{booktabs}
\usepackage{soul}
\usepackage{makecell}
\extrafloats{100}

\journal{Physica D: Nonlinear Phenomena}

\begin{document}

\begin{frontmatter}

\title{Fold Bifurcation Identification through Scientific Machine Learning}

\author[BUD,LEN]{Giuseppe Habib\corref{cor1}}
\ead{habib@mm.bme.hu}
\cortext[cor1]{Corresponding Author}
\author[BUD]{\'Ad\'am Horv\'ath}
%\author[EMMEECS]{Mia Loccufier} Add or not?
\address[BUD]{Department of Applied Mechanics, Faculty of Mechanical Engineering, Budapest University of Technology and Economics, M\H uegyetem rkp. 3., H-1111,Budapest, Hungary}
\address[LEN]{MTA-BME Lend\"ulet ``Momentum" Global Dynamics Research Group, Budapest University of Technology and Economics, M\H uegyetem rkp. 3., H-1111,Budapest, Hungary}

\begin{abstract}
This study employs scientific machine learning to identify transient time series of dynamical systems near a fold bifurcation of periodic solutions. The unique aspect of this work is that a convolutional neural network (CNN) is trained with a relatively small amount of data and on a single, very simple system, yet it is tested on much more complicated systems. This task requires strong generalization capabilities, which are achieved by incorporating physics-based information. This information is provided through a specific pre-processing of the input data, which includes transformation into polar coordinates, normalization, transformation into the logarithmic scale, and filtering through a moving mean. The results demonstrate that such data pre-processing enables the CNN to grasp the important features related to transient time-series near a fold bifurcation, namely, the trend of the oscillation amplitude, and disregard other characteristics that are not particularly relevant, such as the vibration frequency. The developed CNN was able to correctly classify transient trajectories near a fold for a mass-on-moving-belt system, a van der Pol-Duffing oscillator with an attached tuned mass damper, and a pitch-and-plunge wing profile. The results contribute to the progress towards the development of similar CNNs effective in real-life applications such as safety monitoring of dynamical systems.
\end{abstract}

\begin{keyword}
  Scientific machine learning \sep physics-informed neural network \sep convolutional neural network \sep fold bifurcation \sep saddle-node bifurcation \sep bifurcation prediction
\end{keyword}
\end{frontmatter}

%\linenumbers

\section{Introduction}

\begin{figure}
\begin{center}
\setlength{\unitlength}{\textwidth}
\includegraphics[width=0.5\textwidth]{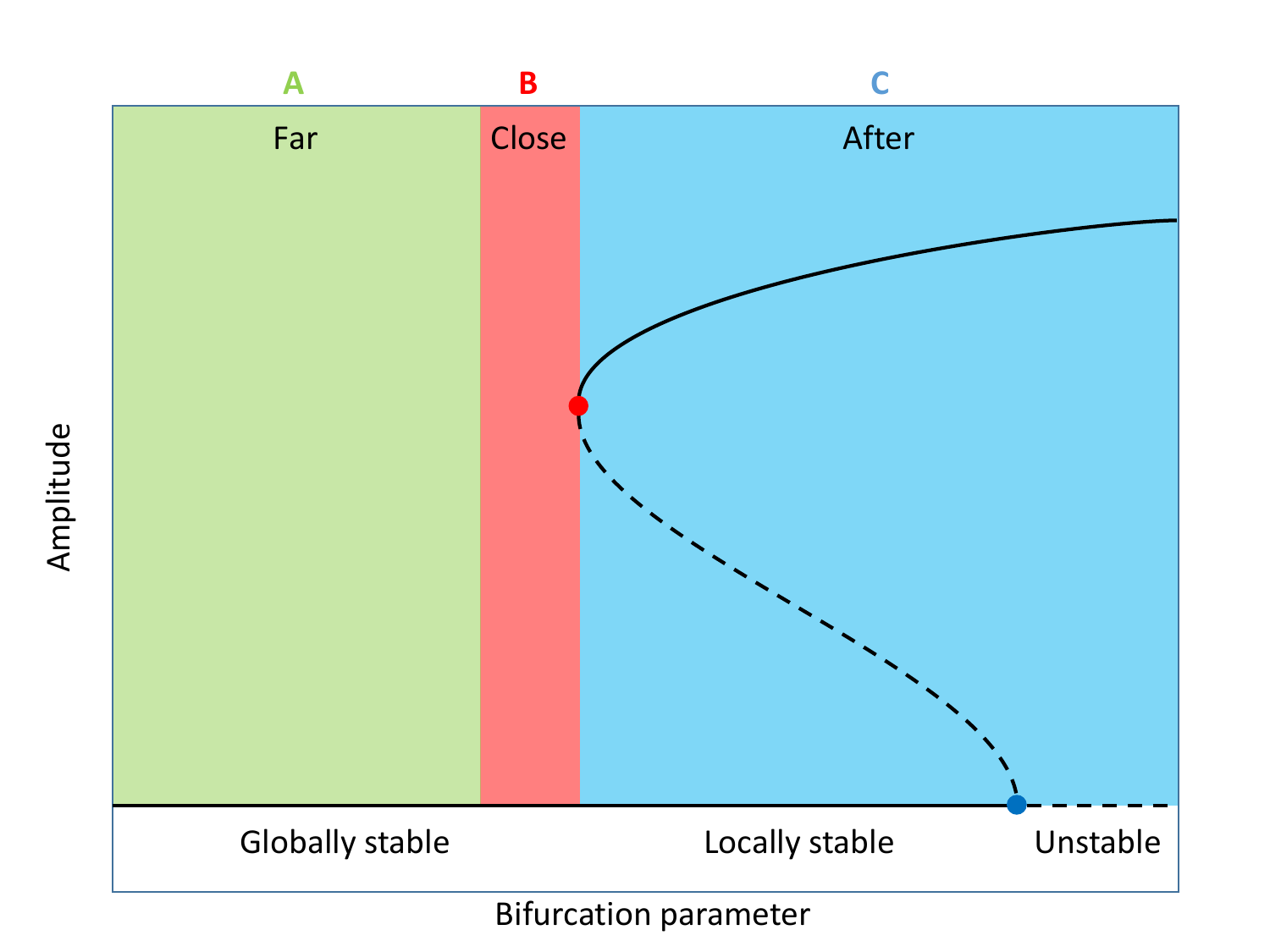}
\end{center}
\caption{\label{fig_generic_fold}Illustrative example of a fold bifurcation (red dot) originated by a subcritical Andronov-Hopf bifurcation (blue dot). Green, red and blue colors mark regions {\em far} (A), {\em close} (B) and {\em after} (C) the fold, respectively.}
\end{figure}

Fold, or saddle-node, bifurcations are a type of bifurcation related to the merging of a saddle and a node steady-state solutions; hence the name saddle-node bifurcation \cite{kuznetsov1998elements}.
The saddle and node solutions can have different natures, typically equilibrium points or periodic solutions.
In both cases, if a point of each solution is represented in a plane for variations of a parameter, the branch of the two solutions generally assumes a folded shape, as shown in Fig.~\ref{fig_generic_fold}, from which the other name fold bifurcation is derived. Accordingly, fold bifurcations mark the appearance of two steady-state solutions, one stable and one unstable.
In this study, we refer uniquely to fold bifurcations of periodic solutions.

Fold bifurcations have an important role in engineering practice.
Let us consider the bifurcation scenario in Fig.~\ref{fig_generic_fold} and assume that the upper branch of stable steady-state solutions represents the working condition of a system \cite{jiang2020bursting, rothe1992multiple}.
If the bifurcation parameter decreases below the value of the fold bifurcation, the desired steady-state ceases to exist, leading the system to suddenly diverge from its state, usually catastrophically.
Accordingly, this scenario is a mathematical catastrophe \cite{arnold1981singularity}.

On the other hand, if the desired solution is not directly related to the fold (for instance, the trivial equilibrium in Fig.~\ref{fig_generic_fold}), then the fold does not affect the system's local dynamics around that equilibrium.
In fact, this solution does not change its stability properties when the bifurcation parameter crosses the value of the fold bifurcation.
This scenario suggests that the fold bifurcation can be neglected in this case, which is often done in practice.
However, the fold can have an important effect on the system's global dynamics.
Let us consider the scenario in Fig.~\ref{fig_generic_fold}, assuming that no other solution exists; in regions A and B, before the fold, the trivial solution is globally stable, while after the fold, it is only locally stable.
This means that, in the locally stable region, while a small perturbation will only generate a transient effect, a large enough perturbation might lead the system to converge towards the large amplitude solutions.
In practice, this scenario is particularly elusive; in fact, even after extensively testing the system, initial conditions leading to the upper branch of solutions might never be encountered, leaving the fold bifurcation undetected.
However, during regular operations, it might be enough to have such perturbation only once to cause an accident.
This phenomenon is usually referred to as the dynamical integrity of a stable solution, which is unbounded only for globally stable solutions \cite{lenci2019global}.

Several systems of engineering relevance suffer from limited dynamical integrity and local stability. Examples include flutter instability \cite{ghadami2016bifurcation, nitti2021spatially, habib2021dynamical}, machining processes \cite{dombovari2019experimental, szaksz2024dynamical}, brake squeal \cite{papangelo2017subcritical, hu2020friction}, robot control \cite{habib2013nonlinear, habib2022bistability}, wheel shimmy \cite{horvath2022stability, habib2023towed}, pressure relief valves \cite{kadar2023nonlinear}, turbulent flows \cite{cherubini2015nonlinear, kerswell2018nonlinear}, electric blackouts \cite{gajduk2014stability, ren2015early}, human brain epilepsy \cite{suffczynski2004dynamics, lytton2008computer}, human balance \cite{zakynthinaki2010modeling, smith2017basins} and prey-predator ecosystems \cite{saleh2022basins, garai2023coexistence}, to name a few.

Quite often, regions of local stability are related to subcritical bifurcations. For example, subcritical Andronov-Hopf bifurcations generate a branch of unstable periodic solutions that coexists with the stable one; subcritical Neimark-Sacker bifurcations generate a similar scenario \cite{kuznetsov1998elements}, but with quasiperiodic solutions.
However, in some cases, limited dynamical integrity is due to isolated branches of periodic limit cycles, which makes their detection even harder \cite{habib2018isolated}.
These are encountered in several systems, such as chemical reactors \cite{uppal1976classification}, traffic models \cite{martinovich2023nonlinear}, bridges under wind excitation \cite{zulli2012bifurcation}, and shimmy of rigid wheels \cite{takacs2008isolated}, for example.

Several methods exist to identify and predict fold bifurcations.
Analytical methods typically provide an approximate solution to the system dynamics, usually in the form of a system of algebraic equations.
This applies to the multiple-scale \cite{nayfeh2008nonlinear}, averaging \cite{verhulst2023toolbox}, or harmonic balance methods, to mention a few.
Once a system of algebraic equations for the system's solution is available, a fold can be found either by plotting the whole solutions in a given parameter range or by studying the system singularities through implicit derivatives of the system of algebraic equations \cite{golubitsky1985singularities}.
Clearly, implementing these methods requires knowledge of the system's equations of motion.

Numerical methods to identify fold bifurcations mostly consist of pseudo-arclength algorithms coupled with shooting techniques or collocation-based methods.
By varying the bifurcation parameter, one of the two branches of periodic solutions is continued until the fold is reached \cite{nayfeh2008applied}.
Alternatively, through direct numerical integration of the equations of motion and by performing a sweep-up and a sweep-down of the bifurcation parameter, the fold is defined by a sudden jump-down and settling of the periodic solution onto an equilibrium (for a bifurcation scenario similar to the one depicted in Fig.~\ref{fig_generic_fold}).
Although these methods typically exploit the system's equations of motion, they can also be implemented through a black-box model of the system.

Similarly, experimental methods for fold bifurcation identification usually implement a sweep-up and a sweep-down of the bifurcation parameter while the system operates.
The fold is identified by a sudden jump-down of a periodic solution.
Additionally, control-based continuation techniques enable tracking unstable solutions until a fold is reached or even track folds in the parameter space \cite{renson2019numerical}.

Several methods to predict the presence of fold bifurcations directly from time series, without actually reaching it, were proposed in the literature.
Suppose the system is assumed on the branch of periodic solutions leading to the fold. In that case, these methods usually exploit early-warning signals related to statistical properties of the system's response to small perturbations \cite{liu2015identifying, ren2015early}.
These methods can recognize if a system is close to a fold (in this context, usually referred to as tipping point \cite{lenton2020tipping} or critical transition \cite{scheffer2009early, lade2012early}), possibly enabling operators to control the system and avoid reaching the fold, which would be a non-reversible transition.

If the system in operation is on the branch of solutions not directly related to the fold, small perturbations can hardly be used to predict the fold; in fact, the fold has no local effect on the solution.
However, large perturbations can still be used to forecast the fold without actually reaching it.
Epureanu and coworkers developed similar methods in a series of papers \cite{ghadami2016bifurcation, lim2011forecasting, chen2018forecasting, garcia2024data}; the methods they developed allow one to predict a full bifurcation diagram from the analysis of the displacement decay of a few trajectories obtained in the pre-bifurcation scenario.
A similar technique, specifically targeting fold bifurcations, was recently developed in \cite{habib2023predicting} and tested experimentally in \cite{kadar2024model}.

In this study, we try to perform a similar task utilizing machine learning.
In particular, we utilize so-called scientific convolutional neural networks (NNs).
NNs have been long used to analyze time series and identify anomalies.
Examples include but are not limited to, human activity recognition \cite{ordonez2016deep, xu2019innohar}, electrocardiogram classification \cite{zihlmann2017convolutional, hannun2019cardiologist}, speech recognition \cite{hinton2012deep, nassif2019speech}, natural language processing \cite{goldberg2016primer, goldberg2022neural}.
NNs have the ability to handle long-term dependencies and temporal context \cite{yu2019review}.
They can recognize patterns and are good at performing classifications \cite{li2021survey}.
All these properties make them a potentially useful tool for identifying if a time series is near a fold bifurcation in the parameter space.
Many studies demonstrated the ability of NNs to classify and process time series related to dynamical systems.
For example, they are used for chatter detection in machining \cite{lamraoui2015chatter, rahimi2021line}, chaos identification \cite{boulle2020classification}, fault detection and diagnosis in mechanical systems \cite{chen2021data, junior2022fault}, anomaly detection \cite{li2019deep}, and many other similar tasks.

Recently, NNs have also been implemented for bifurcation predictions. For example, in \cite{deb2022machine}, a sort of convolutional NN (CNN), combined with a long-short term memory (LSTM) architecture, is exploited to predict tipping points. The network, trained only with synthetic data from several systems, could predict tipping points from real-life data.
A similar result is illustrated in \cite{dylewsky2023universal}, applied to climate systems.
In \cite{fan2021anticipating}, the synchronization of chaotic systems is predicted through NNs. Other studies addressed the problems of bifurcation prediction through NNs in more general terms. For instance, in \cite{bury2021deep}, deep learning is exploited to predict a large range of different bifurcations, training a CNN-LSTM with synthetic data from differential equations and using it on real-life ones. In \cite{bury2023predicting}, a similar result is obtained using discrete time equations, also providing promising results. Both studies \cite{bury2021deep, bury2023predicting} required using a large number of randomly generated models (in the order of tens of thousands).

Although NNs are excellent interpolators, being able to deal with large inputs, they are generally unable to extrapolate results, meaning that if an input is outside of the range of the input used for training, a NN tends to provide a wrong output \cite{zhu2023reliable}.
One method to overcome, or at least limit, this problem is to design and train the network, taking into account the physics of the problem as well, and not only data; in other words, by providing the network with some physics-based information. Such NNs are usually referred to as scientific or physics-informed NNs.
The main idea behind them is to introduce an appropriate inductive, learning, or observational bias that can steer the learning process towards identifying physically consistent solutions \cite{karniadakis2021physics}.

Inductive bias consists of tailoring the NN architecture in such a way that it implicitly satisfies some physical laws; this approach is usually implemented through some sort of invariant symmetries and rotations; this is done, for instance, by CNNs \cite{lecun1995convolutional}, providing excellent results in computer vision and many other fields.

Learning bias is usually provided through the loss function, which also includes a penalty if a given mathematical function is not satisfied. This mathematical function can have any desired form, and it should represent a physical principle that we want the NN to approximately satisfy, such as, for example, conservation of mass or energy in a mechanical system \cite{raissi2019physics}.

Observational bias consists of providing the network with data embodying the underlying physics of the problem or using specifically designed data augmentation procedures. For instance, an observational bias can be obtained by conveniently balancing experimental and synthetic data. This method was successfully applied in a variety of fields, such as, for example, fracture propagation \cite{chen2023physics}.

In this work, we investigate the ability of a scientific CNN to identify trajectories that are close to a fold bifurcation.
The NN is physics-informed through an observational bias obtained through an appropriate input normalization, which hides some signal properties (the frequency) and highlights other properties (the oscillation amplitude trend).
Several studies in the literature have already approached the same problem through machine learning \cite{bury2021deep, patel2023using}; however, up to the authors' knowledge, it is the first time that the prediction is performed from trajectories not converging to the branch related to the fold itself.
Our results show that such an NN can classify trajectories with excellent accuracy and possesses remarkable extrapolation properties.

\section{Objective}

This study aims to train a CNN to identify if a time series belongs to a parameter set near a fold bifurcation.
This objective is formalized in a classification task, as detailed below.
The developed CNNs will be trained to distinguish between trajectories referring to a point in the parameter space that is {\em far} from a fold (region A in Fig.~\ref{fig_generic_fold}), {\em close} to a fold (region B), or {\em after} a fold (region C).
Considering that the fold bifurcation generates two branches of periodic solutions, regions of the parameter space where the branches of periodic solutions do not exist are interpreted as before the fold (either \emph{far} from it or \emph{close} to it). In contrast, regions where they exist are considered as \emph{after}.
The bifurcation parameter is constant for each trajectory.

We consider trajectories that are always initiated from relatively large initial conditions, such that they capture the transient dynamics related to the fold (although it might not be the case, as illustrated in \cite{habib2023predicting}).
Other methods for forecasting fold bifurcations have, in fact, the same requirement \cite{lim2011forecasting, habib2023predicting}.

The CNNs are not asked to distinguish among the region between the fold and the Andronov-Hopf bifurcation and the region after the Andronov-Hopf bifurcation. Although they are very different from an engineering perspective, their differences are qualitative only in the relative vicinity of the trivial solution, while at high amplitude, close to the stable periodic solutions, they are similar.
As the time series fed to the CNNs are initiated at high amplitudes, very weak information distinguishing the two regions is available to the CNNs.
Therefore, they are both classified as \emph{after}.
Nevertheless, this aspect is of practical relevance and will be considered in future developments of this study.

All utilized trajectories relative to the bistable region converged to the periodic solution because of their large initial conditions. This was purposely done in order to simplify the computation; in fact, trajectories starting from small amplitude and converging to the equilibrium, even if in the bistable region, can easily be confused with trajectories far from the fold.
This simplification does not reflect any specific engineering aspect in terms of application, but it is a necessary step to approach the problem at hand gradually. Future studies will also address small amplitude oscillations, aiming at identifying features distinguishing low amplitude trajectories.

The major challenge of the study is that the CNNs will be trained on a single simple system (single-degree-of-freedom (DoF) oscillator with nonlinear damping) and tested on very different and more complicated systems. Accordingly, a CNN must discern and use the general features that characterize a system near a fold bifurcation to be effective. Conversely, all features relevant to the specific training system should be discarded.
As extrapolating is notoriously hard for NNs, this task is intrinsically challenging.

The CNNs will be trained with a specific set of parameter values (except for the bifurcation parameter) of the oscillator with nonlinear damping, as described below.
Then, the following test will be performed to assess the effectiveness of the network:
\begin{enumerate}
\item verify the accuracy in classifying time series from the same system and same range of parameter values as for the training test
\item verify the accuracy in classifying time series from the same system but a different and wider range of parameter values
\item verify the accuracy in classifying time series from other systems. In particular, three other systems are considered: a mass-on-moving-belt system, a van der Pol-Duffing oscillator with an attached dynamic vibration absorber, and a pitch-and-plunge wing profile. The systems are described in Sect.~\ref{sect_models}.
\end{enumerate}

If the training data are assorted enough, the first task is achievable through any sufficiently complex NN, even a classical fully connected one. It is, in fact, a simple validation of the training.
The second task already requires some sort of extrapolation from the network, but to a limited extent, considering the simplicity of the system.
Conversely, the third task can be accomplished only if the network grasps the essential features of time series near fold bifurcations. Any other property specific to the training system will disturb the prediction.
As shown later, this will be only possible through a proper normalization of the data provided to the network based on a physical understanding of the phenomenon.

We notice that, while the distinction between regions B and C (\emph{close} and \emph{after} the fold) is precisely defined by the bifurcation, the boundary between trajectories far and close to the fold is completely arbitrary.
To overcome this problem, trajectories used to train the CNNs of the two groups are very well separated, as explained in Sect.~\ref{sec_training}.
Then, the trained CNNs are tested on a large range of values of the bifurcation parameter in order to evaluate their capabilities.

\section{Mathematical models} \label{sect_models}

Four different dynamical systems are considered for the analysis: a single-DoF system with nonlinear damping, a non-smooth mass-on-moving-belt system, a van der Pol-Duffing oscillator with an attached vibration absorber, and a pitch-and-plunge wing profile.
The first one is the only one used for the training.

\subsection{Oscillator with nonlinear damping}

A single-DoF oscillator having a nonlinear damping characteristic is considered, whose dynamics is described by the equation of motion
\begin{equation}
\ddot x+x+c_1 \dot x-c_3\dot x^3\left(1-\dot x^2\right)=0.\label{eq_EOM_NLD}
\end{equation}

\begin{figure}[h]
\begin{center}
\setlength{\unitlength}{\textwidth}
\begin{subfigure}[b]{0.325\textwidth}
    \includegraphics[width=\textwidth]{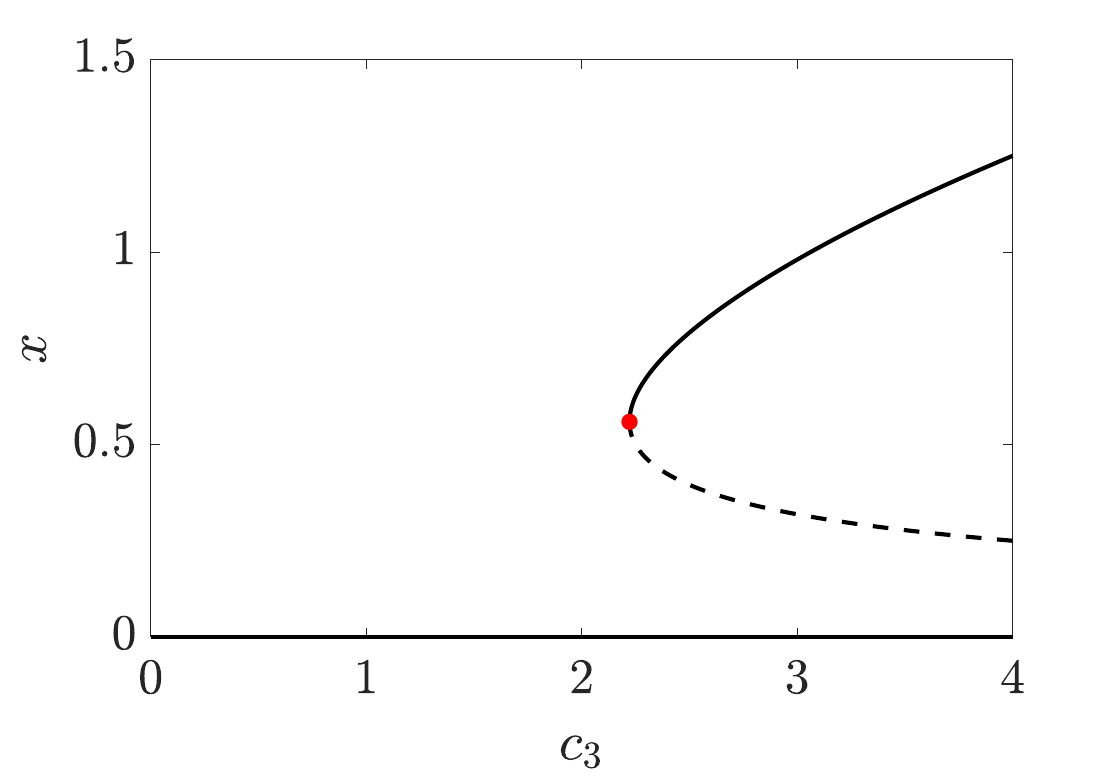}
    \caption{Nonlinear damping system}
    \label{fig_bif_NLD}
  \end{subfigure}
\begin{subfigure}[b]{0.325\textwidth}
    \includegraphics[width=\textwidth]{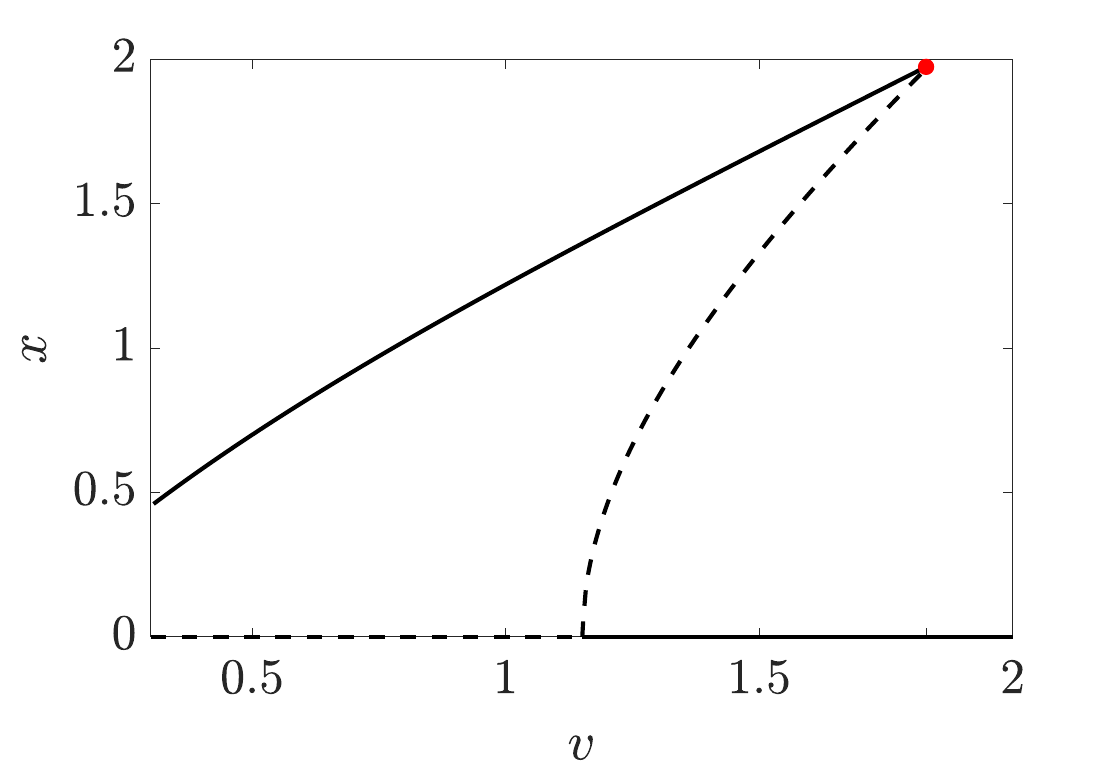}
    \caption{Mass-on-moving-belt}
    \label{fig_bif_MOB}
  \end{subfigure}\\
\begin{subfigure}[b]{0.325\textwidth}
    \includegraphics[width=\textwidth]{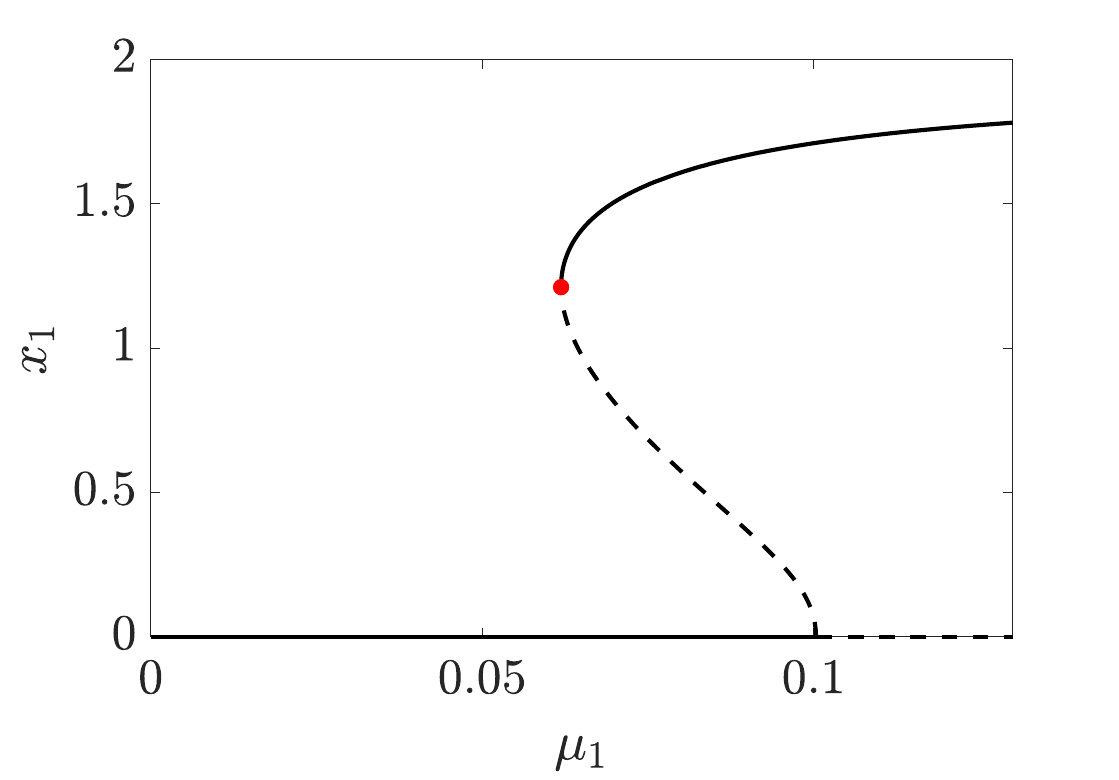}
    \caption{Van der Pol-Duffing oscillator}
    \label{fig_bif_VDP}
  \end{subfigure}
\begin{subfigure}[b]{0.325\textwidth}
    \includegraphics[width=\textwidth]{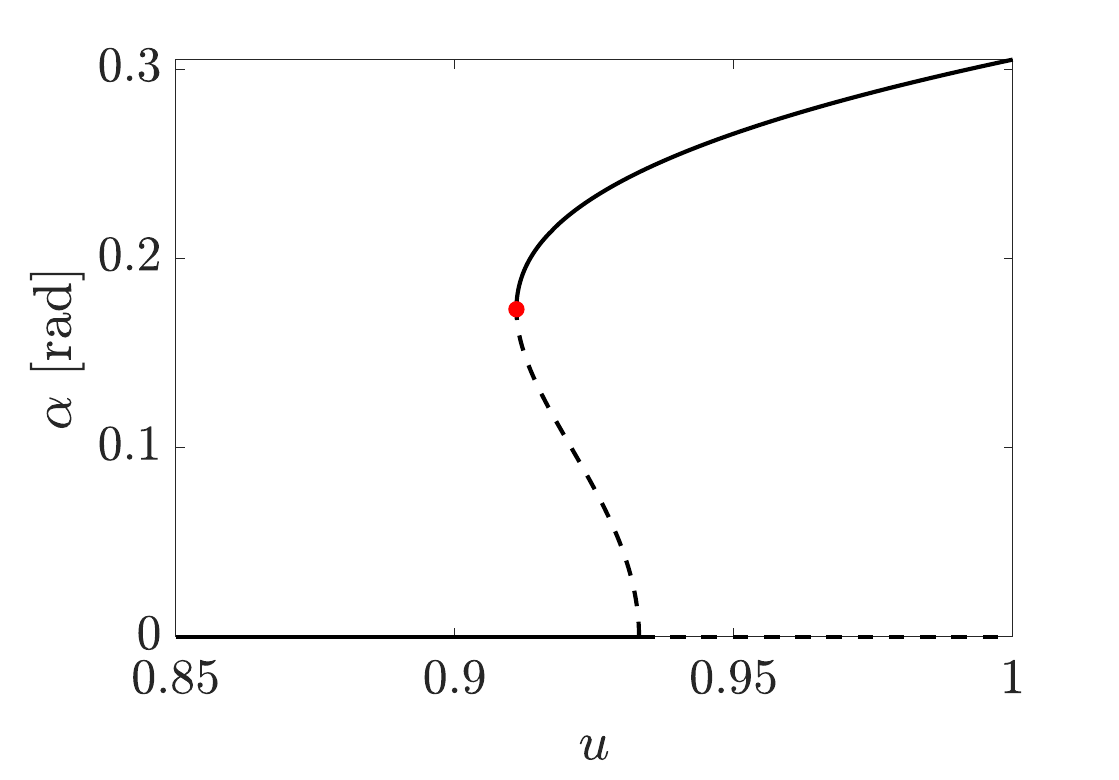}
    \caption{Pitch-and-plunge wing}
    \label{fig_bif_PnP}
  \end{subfigure}
\end{center}
\caption{\label{fig_bif_all}Bifurcation diagrams of the investigated systems; solid lines: stable solutions, dashed lines: unstable solutions; red dots: fold bifurcations. The lines mark the maximal amplitude of the steady-state solutions for variations of the bifurcation parameter. (a) System in Eq.~(\ref{eq_EOM_NLD}) for $c_1=0.5$; (b) system in Eq.~(\ref{eq_EoM_MoB}); (c) system in Eq.~(\ref{eq_EOM_VDP}) for the parameter values $r=0.05$, $\gamma=0.97$, $\mu_2=0.12$ and $\alpha=0.3$; (d) system in Eq.~(\ref{eq_EOM_PnP}).}
\end{figure}

For $c_1>0$, the trivial solution is always stable.
For $c_3=0$, the system is linear, while for $c_3\neq0$ the damping has a nonlinear characteristic. If $c_3>0$, for increasing oscillation amplitudes, first damping decreases and then increases because of the cubic and fifth order terms.
As thoroughly investigated in \cite{habib2018isolated}, for $c_3=c_3^*:=40c_1/9$ the system presents a fold bifurcation, which generates two branches of periodic solutions existing for $c_3>c_3^*$.
Increasing $c_3$, the branch of unstable periodic solutions approaches the trivial solution, reducing its dynamical integrity.
However, as far as $c_1>0$, they never merge.
The corresponding bifurcation diagram is illustrated in Fig.~\ref{fig_bif_NLD}.

Although the differential equation in Eq.~(\ref{eq_EOM_NLD}) does not model any specific physical system, several mechanical systems present a nonlinear damping force, which can lead to similar dynamics.

\subsection{Mass-on-moving-belt}

\begin{figure}[h]
\begin{center}
\setlength{\unitlength}{\textwidth}
\begin{subfigure}[b]{0.4\textwidth}
    \includegraphics[width=\textwidth]{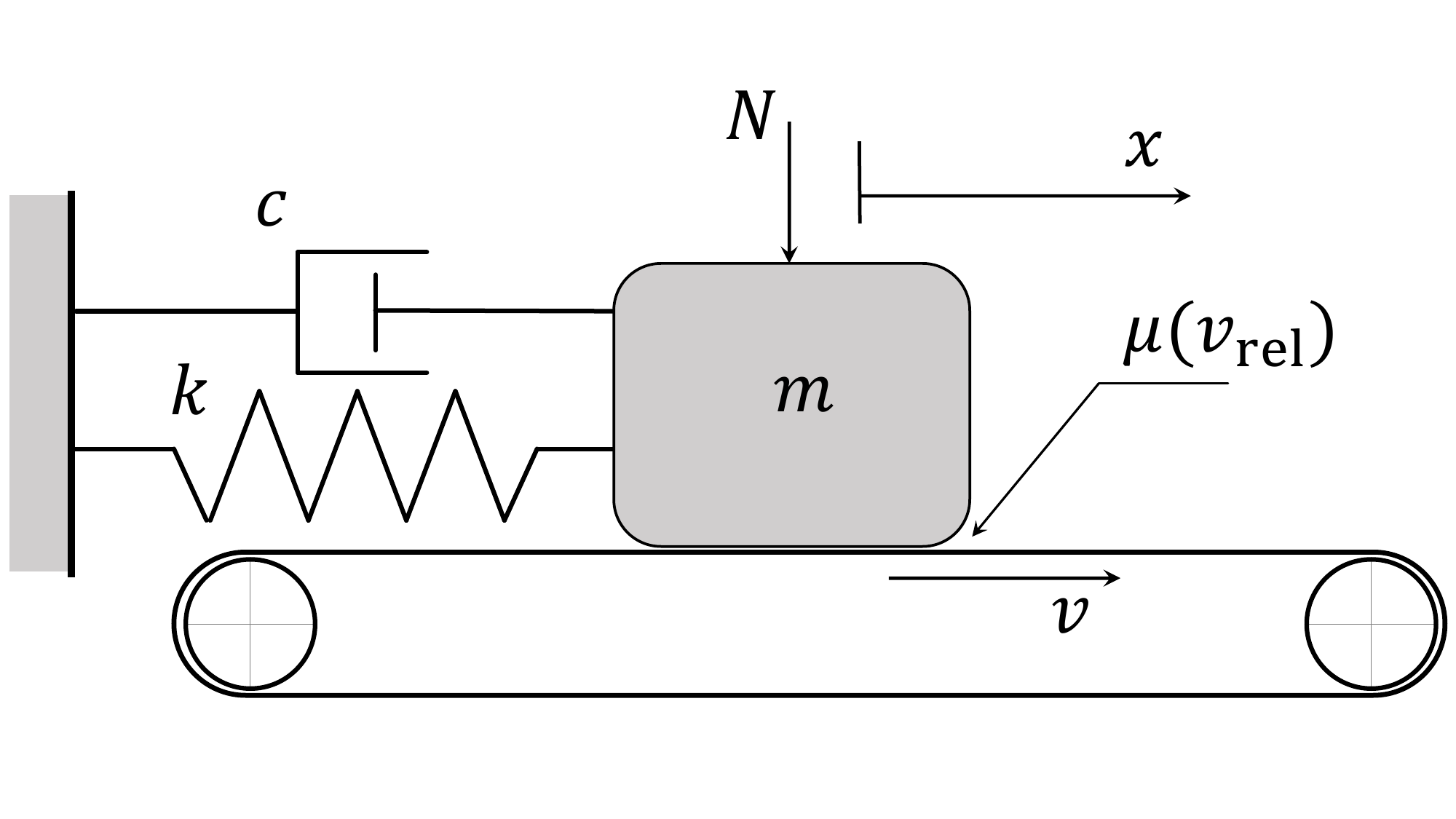}
    \caption{Mass-on-moving-belt system (Eq.~(\ref{eq_EoM_MoB}))}
    \label{MOB_model}
  \end{subfigure}
\begin{subfigure}[b]{0.45\textwidth}
    \includegraphics[width=\textwidth]{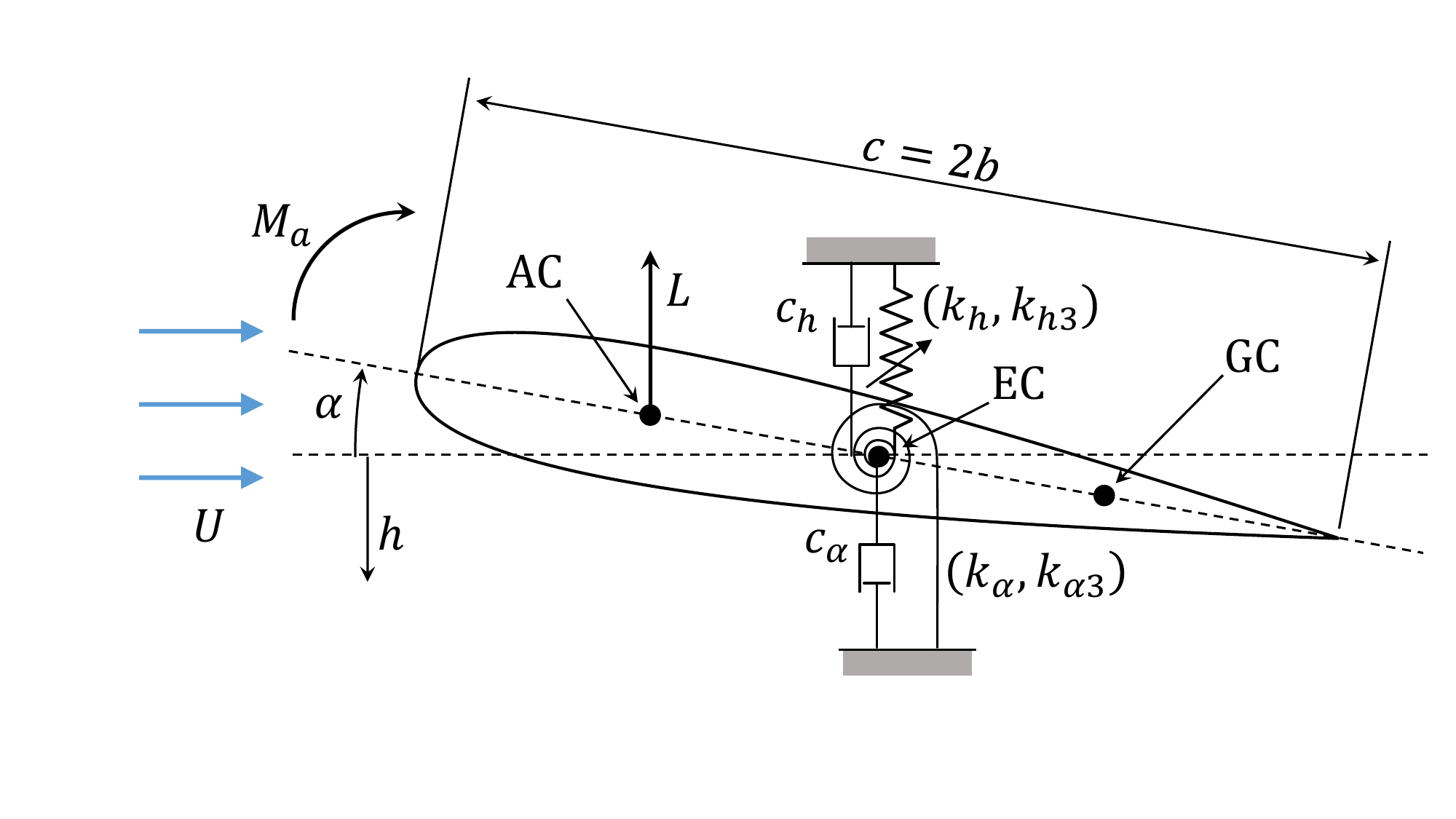}
    \caption{Pitch-and-plunge airfoil (Eq.~(\ref{eq_EOM_PnP}))}
    \label{PnP_model}
  \end{subfigure}
\end{center}
\caption{\label{fig_models}Mechanical models of the mass-on-moving-belt system (a) and pitch-and-plunge airfoil (b).}
\end{figure}

We consider the classical mass-on-moving-belt system illustrated in Fig.~\ref{MOB_model}. This system is an archetypal model for friction-induced vibrations \cite{papangelo2017subcritical, hu2020friction}.
It is typically used for studying violin string dynamics \cite{leine1998stick} and brake squeal \cite{kinkaid2003automotive}.

The non-dimensional equation of motion of the system is as follows \cite{hu2020friction}:
\begin{equation}
\ddot x+2\zeta\dot x+x=F_f, \label{eq_EoM_MoB}
\end{equation}
where 
\begin{equation}
\left\{\begin{array}{ll}
F_{\text f}=\mu\left( v_{\text{rel}}\right)&\quad v_{\text{rel}}\neq0\\
|F_{\text f}|\leq\mu_{\text s}&\quad v_{\text{rel}}=0\,
\end{array}\right.
\end{equation}
and $\zeta$ is the damping ratio.
The friction coefficient $\mu$ is given by the exponential decaying function \begin{equation}
\mu (v_{\text{rel}}) = \left(\mu_{\text d}+\left(\mu_{\text s}-\mu_{\text d}\right) \text e^{-\frac{|v_{\text{rel}|}}{v_0}}\right) \text{sign} \left(v_{\text{rel}}\right) \,,
\end{equation}
where $v_{\text{rel}} = v - \dot x $ is the relative velocity.
The friction force presents a discontinuity for $v_{\text{rel}}=0$.
The adopted parameter values are $\mu_{\text s}=1$, $\mu_{\text d}=0.5$, $v_0=0.5$ and $\zeta=0.05$, which are the same parameter values utilized in \cite{papangelo2017subcritical, hu2020friction, habib2023predicting}.

For any belt velocity, the system presents an equilibrium for which the spring and the friction force are equal and opposite.
However, for belt velocities $v$ larger than a specific $v_{\text{cr}}$, the equilibrium is stable, while for $v<v_{\text{cr}}$ it is unstable.
The instability is marked by a subcritical Andronov-Hopf bifurcation \cite{papangelo2017subcritical}.
At the Andronov-Hopf bifurcation, a branch of unstable periodic solution arises, which then merges with a branch of stable periodic solutions through a saddle-node bifurcation, as illustrated in Figure \ref{fig_bif_MOB}.
The velocity corresponding to the saddle-node bifurcation is denoted here with $v^*$.
For the adopted parameter values, $v_{\text{cr}}=1.151$ and $v^*=1.83$.
The branches of stable and unstable periodic solutions merge in a non-smooth way because of the system's discontinuity.
The stable periodic solutions are characterized by a stick-slip motion.

The NNs will be asked to distinguish between trajectories for $v$ significantly larger than $v^*$ (\emph{far}), slightly larger than $v^*$ (\emph{close}) and smaller than $v^*$ (\emph{after}).
Testing trajectories have uniformly randomly chosen initial conditions in the range $x(0)\in\left(2,\ 3\right)$ and $\dot x(0)\in\left(1,\ 2\right)$. Trajectories were interrupted after 1000 time unites or when oscillation amplitude was below $10^{-8}$. Trajectories are then re-sampled in 1000 steps.

\subsection{Van der Pol-Duffing oscillator with an attached dynamic vibration absorber}

The third model we consider is a van der Pol-Duffing oscillator with an attached dynamic vibration absorber.
This system was thoroughly studied in the literature \cite{gattulli2001simple, gattulli2003one}, since the van der Pol oscillator is an archetypal system for self-excited oscillations \cite{candido2020non}, while the Duffing term provides a nonlinearity in the stiffness, which makes the system natural frequency amplitude-dependent \cite{kovacic2011duffing}.
Accordingly, several studies investigated the suppression and mitigation of these kinds of self-excited oscillations through dynamic vibration absorbers \cite{habib2015suppression}.

The equations of motion governing the system's dynamics are \begin{equation}
\mathbf M\ddot{\mathbf x}+\mathbf C\dot{\mathbf x}+\mathbf{Kx}+\mathbf b\left(\mathbf x,\dot{\mathbf x}\right)=\mathbf 0,\label{eq_EOM_VDP}
\end{equation}
where
\begin{equation}
\begin{split}
&\mathbf x=\left[\begin{array}{c}
x_1\\x_2\end{array}\right],\,\mathbf{M}=\left[\begin{array}{cc}
1&0\\0&r\end{array}\right],\,\mathbf{K}=\left[\begin{array}{cc}
1+\gamma^2r&-\gamma^2r\\-\gamma^2r&\gamma^2r\end{array}\right],\\
&\mathbf{C}=\left[\begin{array}{cc}
-2\left(\mu_1+\gamma\mu_2r\right)&-2\gamma\mu_2r\\-2\gamma\mu_2r&2\gamma\mu_2r\end{array}\right],\,\mathbf{b}=\left[\begin{array}{c}
\alpha x_1^3+2\mu_1 x_1^2\dot x_1\\0\end{array}\right].
\end{split}
\end{equation}
$x_1$ and $x_2$ are the state variables, respectively marking the displacement of the van der Pol-Duffing oscillator and of the vibration absorber, $r$ is the mass ratio, $\gamma$ is the natural frequency ratio, $\mu_1$ is the van der Pol negative damping coefficient, $\mu_2$ is the absorber's pseudo damping ratio and $\alpha$ the nonlinear restoring force coefficient.
For a better description of the system and the derivation of the equations of motion, we address the interested reader to \cite{habib2015suppression}.

For the present study we fixed parameter values at $r=0.05$, $\gamma=0.97$, $\mu_2=0.12$, $\alpha=0.3$, while $\mu_1$ is the bifurcation parameter.
This set of values provides the bifurcation diagram illustrated in Fig.~\ref{fig_bif_VDP}, which presents a fold bifurcation for $\mu_1^*=0.062$.
The CNN will have to distinguish between trajectory obtained for $\mu_1$ significantly smaller (\emph{far}), slightly smaller (\emph{close}), or larger (\emph{after}) than $\mu_1^*$.
Testing trajectories have uniformly randomly chosen initial conditions in the range $x_1(0)\in\left(1,\ 2\right)$, $\dot x_1(0)\in\left(1,\ 2\right)$, $x_2(0)=x_1(0)$ and $\dot x_2(0)=0$.
All trajectories last 1000 time unites and have a sampling frequency of 1 Hz.
We note that this system presents a significant interaction between its two modes, which can make the classification of the trajectories more difficult \cite{ghadami2016bifurcation, habib2023predicting}.

\subsection{Pitch-and-plunge wing profile undergoing flutter instability}

The fourth system considered is a pitch-and-plunge wing profile exposed to an airflow and undergoing flutter oscillations (Fig.~\ref{PnP_model}); the system is similar to the one studied in \cite{malher2017flutter}.
The pitch-and-plunge wing model considered was implemented in various previous studies \cite{dowell2014modern, lee2007suppressing1, lee2007suppressing2} with only slight variations from the one considered here, regarding the system nonlinearities.
Non-dimensional equations of motion governing the dynamics of the system are \begin{equation}
\mathbf M\ddot{\mathbf x}+\mathbf C\dot{\mathbf x}+\mathbf{Kx}+\mathbf b\left(\mathbf x\right)=\mathbf 0,\label{eq_EOM_PnP}
\end{equation}
where
\begin{equation}
\begin{split}
&\mathbf x=\left[\begin{array}{c}
y\\\alpha
\end{array}\right],\,\mathbf K=\left[\begin{array}{ccc}
\Omega^2&\beta u^2\\
0&r_\alpha^2-\nu u^2
\end{array}\right],\,\mathbf C=\left[\begin{array}{ccc}
\zeta_h+\beta u &0\\
-\nu u&\zeta_\alpha
\end{array}\right],\\
&\mathbf M=\left[\begin{array}{ccc}
1&x_\alpha\\x_\alpha &r_\alpha^2\end{array}\right],\,\mathbf b=\left[\begin{array}{c}
0\\
\xi_{\alpha3}\alpha^3+\xi_{\alpha5}\alpha^5
\end{array}\right],
\end{split}
\end{equation}
$\alpha$ marks the pitch rotation, and $y$ indicates the heave displacement, non-di\-men\-sion\-al\-ized in relation to the semichord of the airfoil, while $u$ is the non-dimensional flow velocity.
For the physical meaning of all the other parameters, we address the interested reader to \cite{malher2017flutter}; a similar nomenclature is utilized here for facilitating the comparison.
The adopted parameter values are $x_\alpha=0.2$, $r_\alpha=0.5$, $\beta=0.2$, $\nu=0.08$, $\Omega=0.5$, $\zeta_\alpha=0.01$, $\zeta_h=0.01$, $\xi_{\alpha3}=-1$ and $\xi_{\alpha5}=20$.

For the considered parameter values, the bifurcation diagram for variations of the flow velocity $u$ is shown in Figure \ref{fig_bif_PnP}.
The equilibrium position loses stability through a subcritical Andronov-Hopf bifurcation; the emerging branch of unstable periodic solutions turns back at a fold for $u=u^*$ and becomes stable.
For the chosen parameter values, the Andronov-Hopf bifurcation occurs at $u=0.933$, while the fold is at $u=u^*=0.911$.
The CNN will have to classify trajectories obtained for $u$ significantly smaller (\emph{far}), slightly smaller (\emph{close}), or larger (\emph{after}) than $u^*$.
Testing trajectories have uniformly randomly chosen initial conditions in the range $y(0)\in\left(0.002,\ 0.01\right)$, $\dot y(0)\in\left(0.002,\ 0.005\right)$, $\alpha(0)\in\left(0.1,\ 0.3\right)$, and $\dot \alpha(0)\in\left(0.1,\ 0.3\right)$.
Trajectories were interrupted after 800 time unites or when oscillation amplitude was below $10^{-8}$, computed independently for pitch and heave DoF. Trajectories are then re-sampled in 1000 steps.
We note that this system presents a significant coupling between pitch and plunge motions near the flutter instability, which complicates the classification of the time series \cite{habib2023predicting}.

\section{Methods}

CNNs are a particular type of machine learning architecture that has the ability to automatically extract relevant features and patterns from raw data.
This ability is obtained through a series of convolutional layers that function as feature detectors.
They were first introduced in the 1980s and 1990s by several researchers in the field of computer vision.
One of the earliest forms of CNN is the Neocognitron, developed by Fukushima in 1980 \cite{fukushima1980neocognitron}, which was inspired by the structure and function of the visual cortex and was designed to recognize patterns in visual data.
Another important milestone towards the development of modern CNN is represented by the LeNet-5 architecture, developed by LeCun et al. in 1998 \cite{lecun1998gradient}, which was initially used for check recognition systems for the banking industry.
Despite these developments, CNNs did not become widely adopted until the 2010s, mainly because of the lack of efficient training algorithms, regularization techniques, and the availability of larger datasets.
A breakthrough in CNNs was marked by the AlexNet CNN, developed by Krizhevsky et al. \cite{krizhevsky2012imagenet} in 2012, which achieved a significant improvement in image classification accuracy and marked a turning point in the popularity and adoption of CNNs \cite{li2021survey}.

Today, CNNs are used in a wide range of applications, including image classification, object detection, image segmentation, natural language processing, video processing, and speech recognition. They are an optimal solution for time series classification tasks.

\subsection{Convolutional neural network architecture}\label{sect_architecture}

\begin{figure}
\begin{center}
\setlength{\unitlength}{\textwidth}
\includegraphics[width=0.6\textwidth]{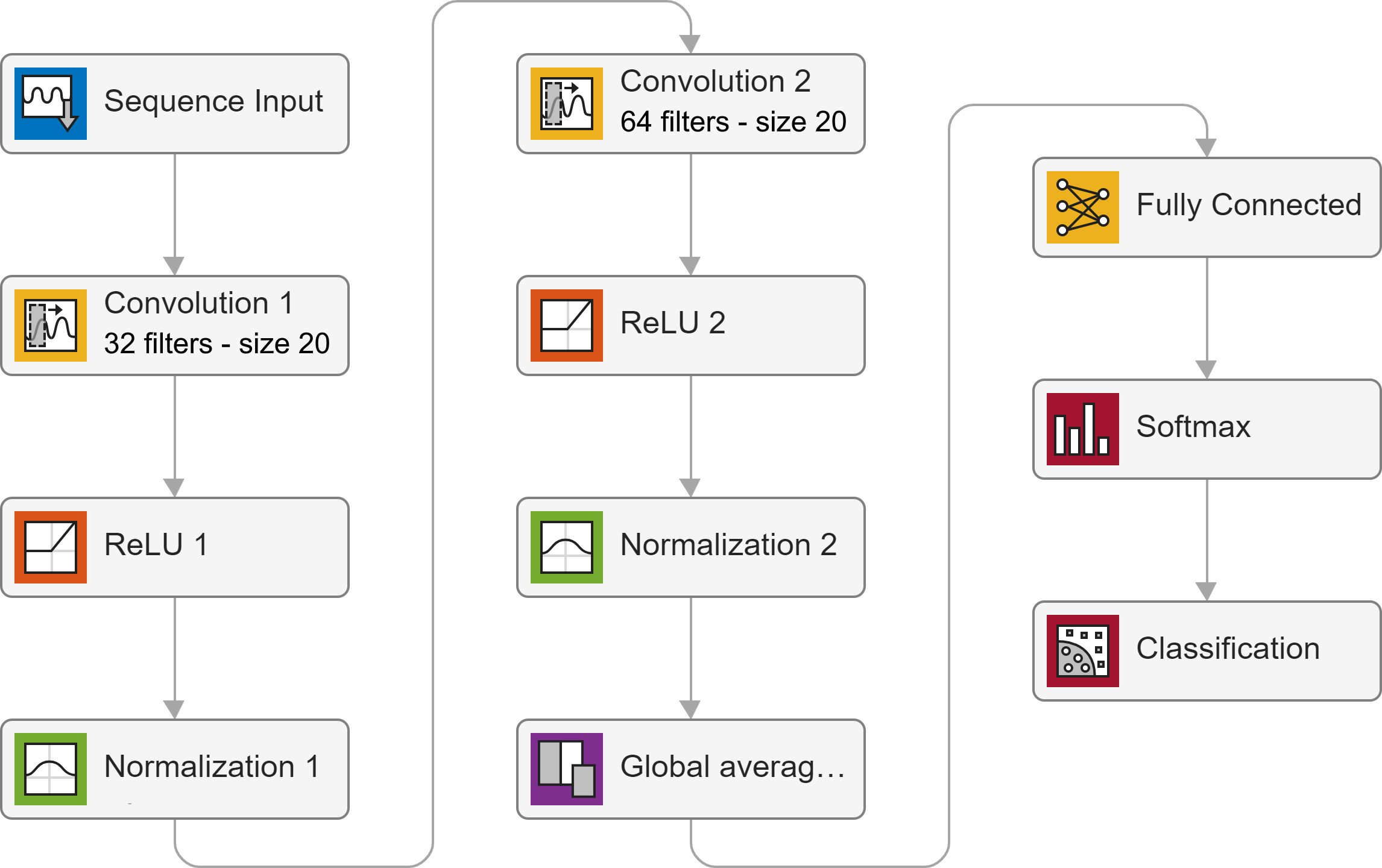}
\end{center}
\caption{\label{CNN_scheme}Graphical representation of the utilized CNN architecture.}
\end{figure}

The CNN utilized in this study is 1D, and its key features are:
\begin{itemize}
\item The network is tailored to process 1-dimensional input data.
\item It includes two convolutional layers. The first layer employs a filter size of 20 with 32 filters, while the second layer uses the same filter size but with 64 filters. ``Causal" padding is applied to account for past values in the input sequence.
\item Rectified linear units (ReLU) are used as activation functions after each convolution. Layer normalization is applied for better training stability.
\item A global average pooling layer is incorporated to calculate the average value of each feature map across the entire sequence.
\item A fully connected layer with an input size of 64 and output of 3 is employed for classification.
\item The network uses a softmax layer to convert its output into class probabilities.
\end{itemize}
The network was trained using the Adam optimization algorithm with a mini-batch size of 32. Training progressed over a maximum of 50 epochs. Initial learning rate is set to 0.001. Padding direction is set to `left'; however, all input signals completely filled the required signal length.
The architecture of the CNN is represented in Fig.~\ref{CNN_scheme}.
Hyperparameters were tuned in a trial-and-error manner and no analysis of the network sensitivity to the hyperparameters was performed.
However, the study does not aim to find the best possible network architecture but rather to analyze the importance of pre-processing the data through physically based assumptions, as explained in Sect.~\ref{sect_input}.
Indeed, the same network architecture was used for all types of data normalization tested.
All computation for this research was computed in the \texttt{MATLAB} environment.

\subsection{Physical information through input manipulation}\label{sect_input}

\begin{figure}[h!]
\begin{center}
\setlength{\unitlength}{\textwidth}
\begin{subfigure}[b]{0.325\textwidth}
    \includegraphics[width=\textwidth]{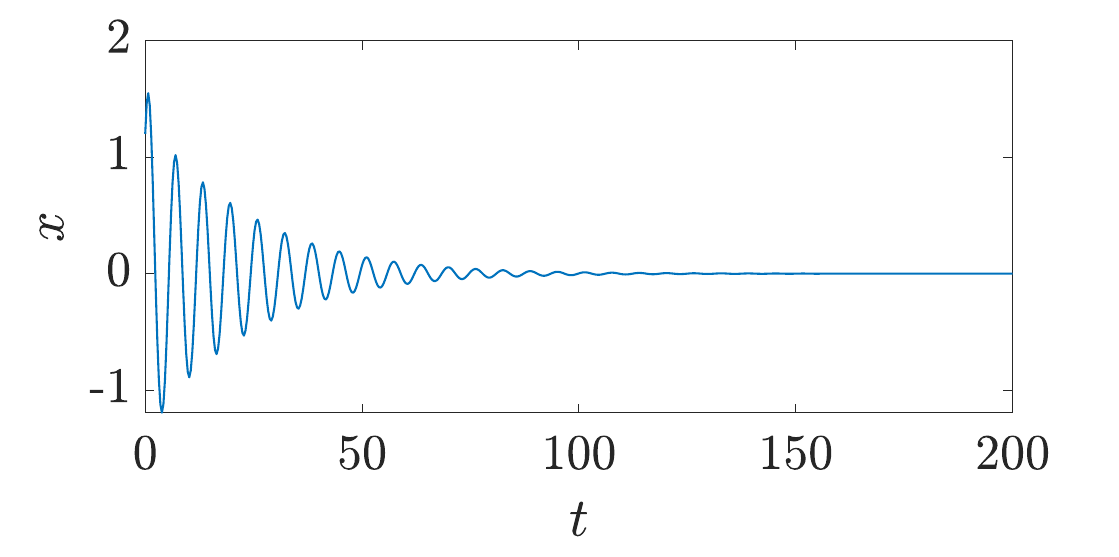}
    \caption{Original signal - far}
    \label{fig_norm_far_original}
  \end{subfigure}
\begin{subfigure}[b]{0.325\textwidth}
    \includegraphics[width=\textwidth]{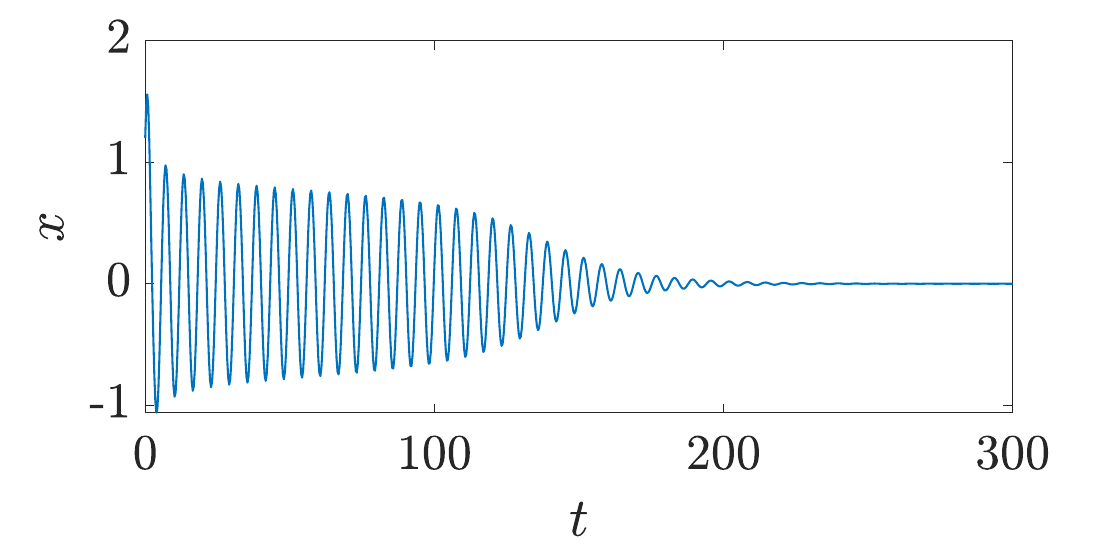}
    \caption{Original signal - close}
    \label{fig_norm_close_original}
  \end{subfigure}
\begin{subfigure}[b]{0.325\textwidth}
    \includegraphics[width=\textwidth]{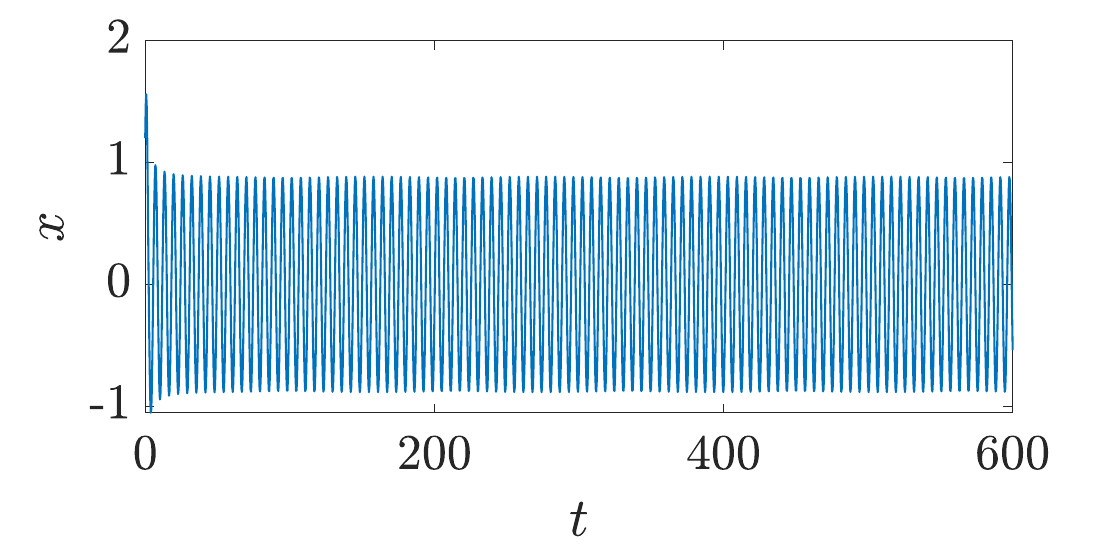}
    \caption{Original signal - after}
    \label{fig_norm_after_original}
  \end{subfigure}
\begin{subfigure}[b]{0.325\textwidth}
    \includegraphics[width=\textwidth]{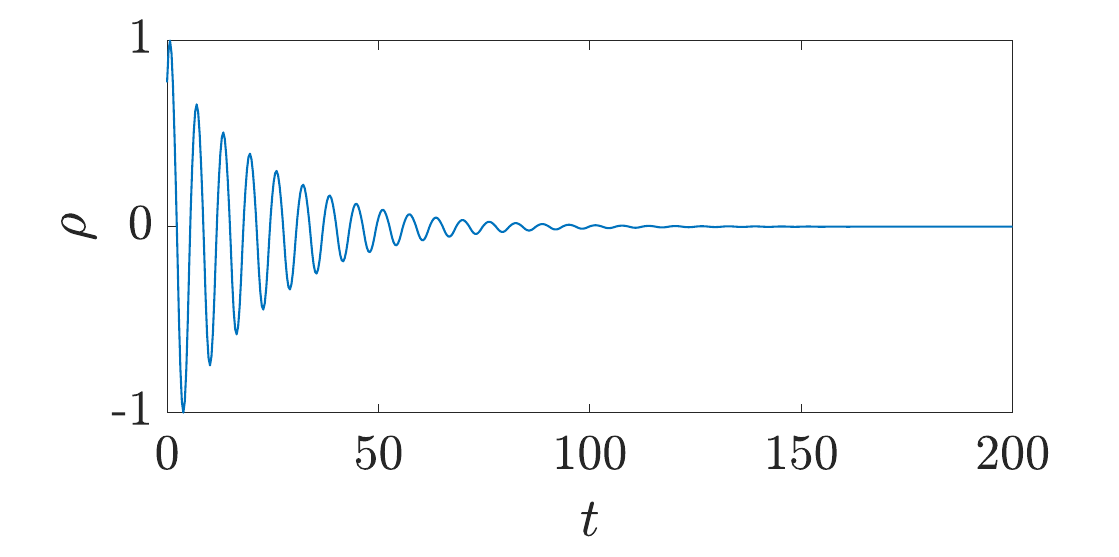}
    \caption{Min-Max - far}
    \label{fig_norm_far_minmax}
  \end{subfigure}
\begin{subfigure}[b]{0.325\textwidth}
    \includegraphics[width=\textwidth]{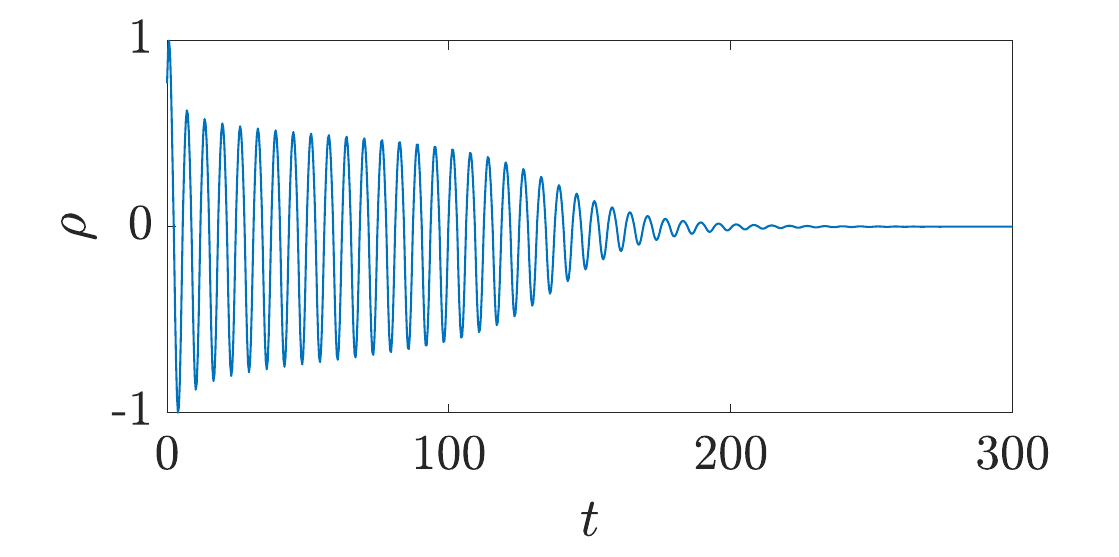}
    \caption{Min-Max - close}
    \label{fig_norm_close_minmax}
  \end{subfigure}
\begin{subfigure}[b]{0.325\textwidth}
    \includegraphics[width=\textwidth]{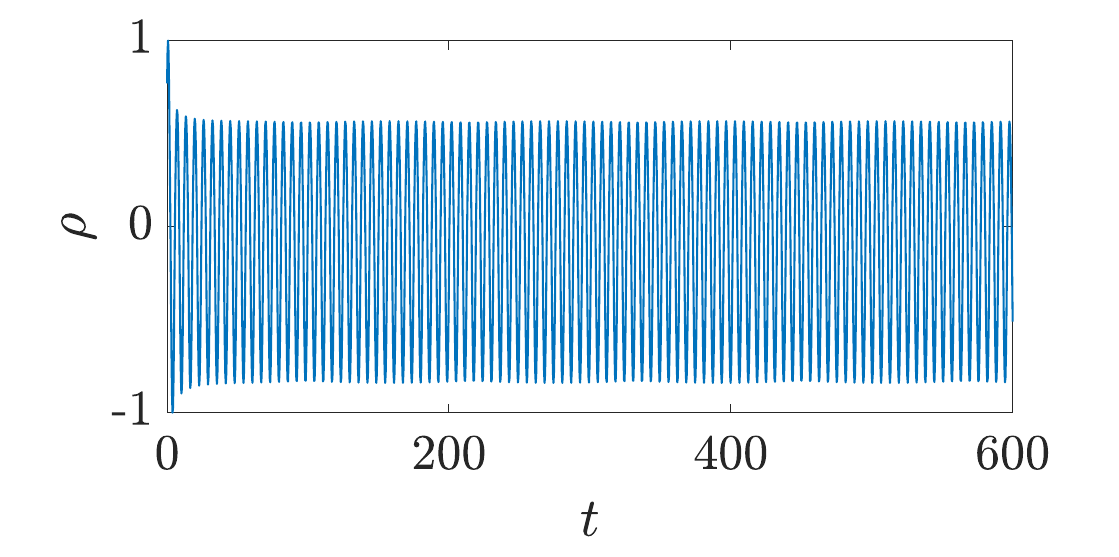}
    \caption{Min-Max - after}
    \label{fig_norm_after_minmax}
  \end{subfigure}
\begin{subfigure}[b]{0.325\textwidth}
    \includegraphics[width=\textwidth]{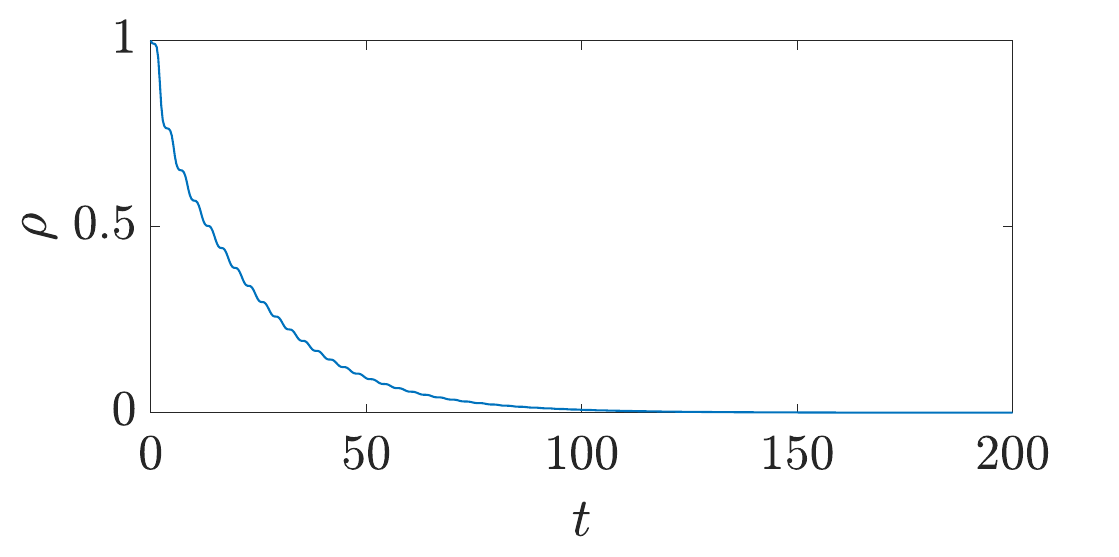}
    \caption{Polar - far}
    \label{fig_norm_far_polar}
  \end{subfigure}
\begin{subfigure}[b]{0.325\textwidth}
    \includegraphics[width=\textwidth]{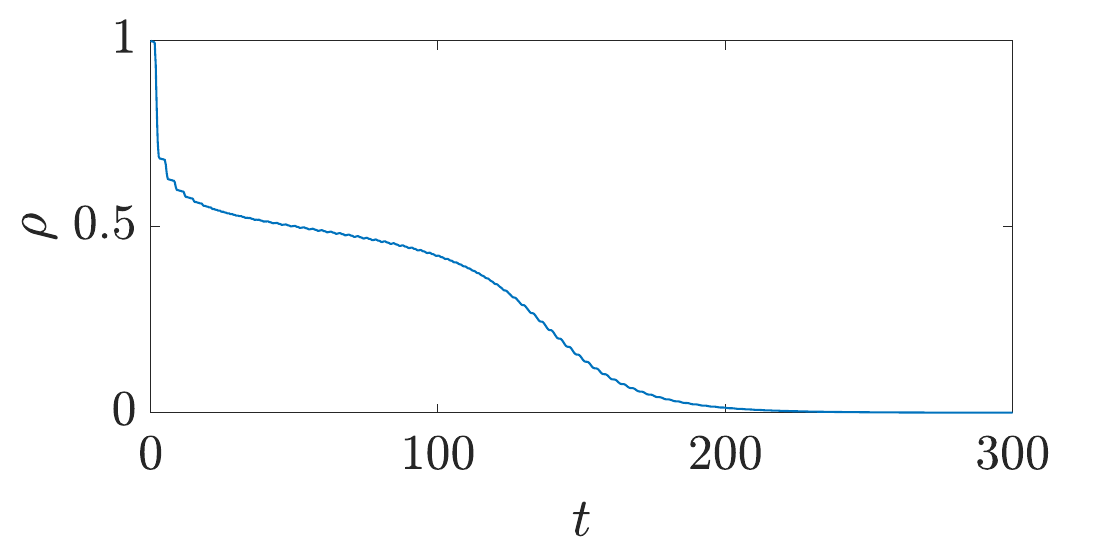}
    \caption{Polar - close}
    \label{fig_norm_close_polar}
  \end{subfigure}
\begin{subfigure}[b]{0.325\textwidth}
    \includegraphics[width=\textwidth]{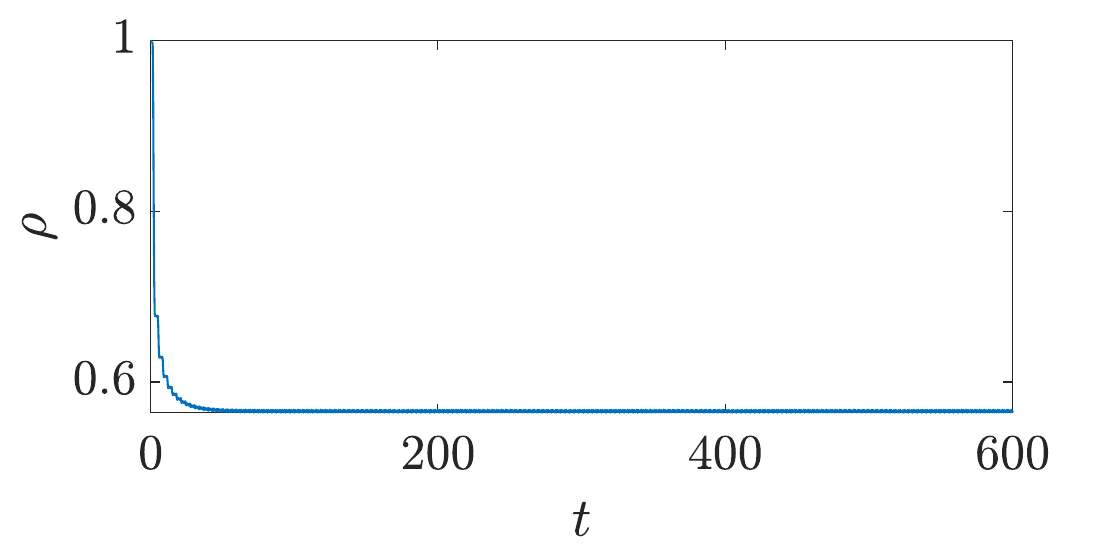}
    \caption{Polar - after}
    \label{fig_norm_after_polar}
  \end{subfigure}
\begin{subfigure}[b]{0.325\textwidth}
    \includegraphics[width=\textwidth]{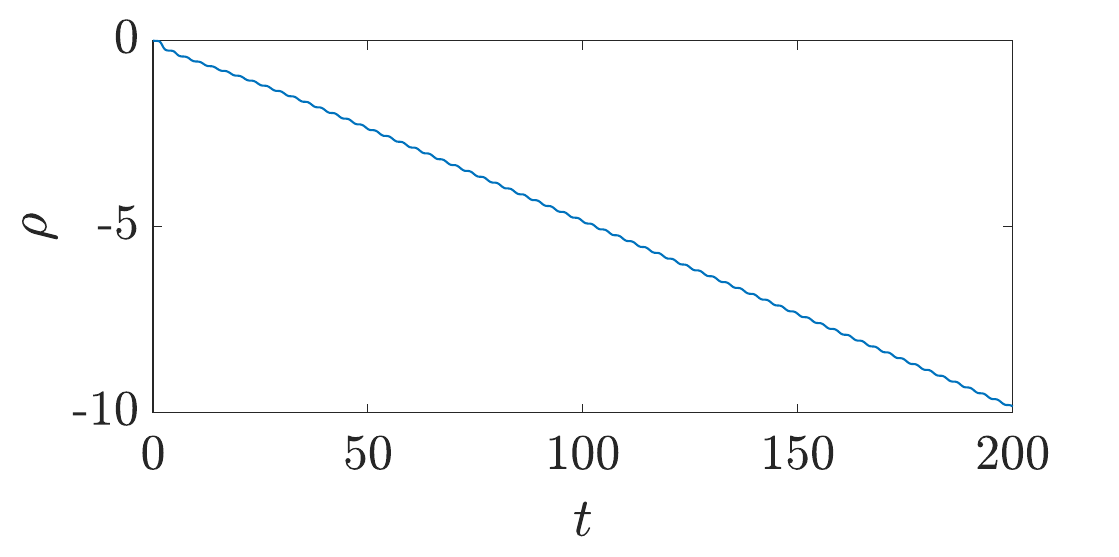}
    \caption{Polar log - far}
    \label{fig_norm_far_polar_log}
  \end{subfigure}
\begin{subfigure}[b]{0.325\textwidth}
    \includegraphics[width=\textwidth]{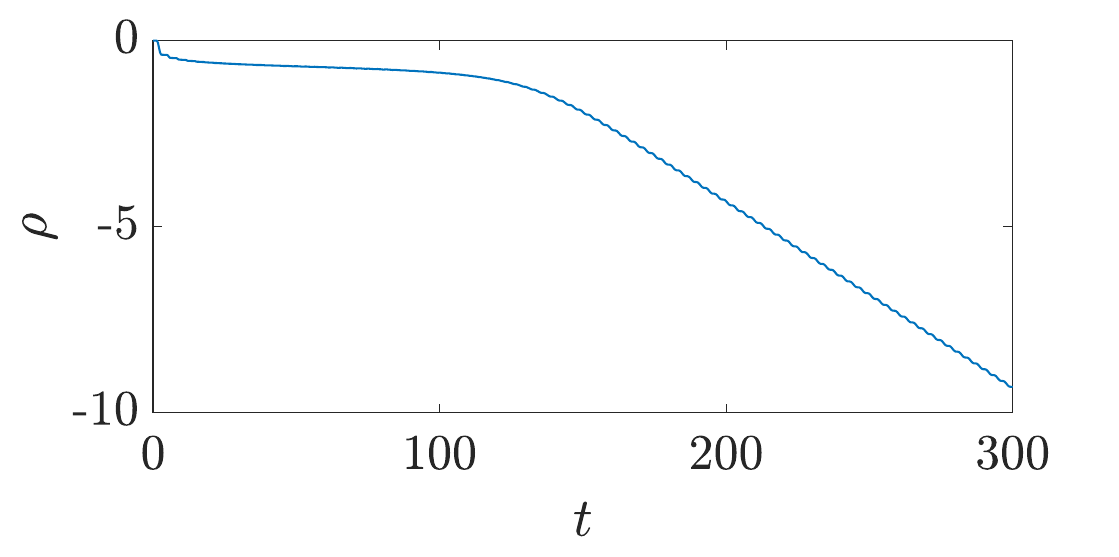}
    \caption{Polar log - close}
    \label{fig_norm_close_polar_log}
  \end{subfigure}
\begin{subfigure}[b]{0.325\textwidth}
    \includegraphics[width=\textwidth]{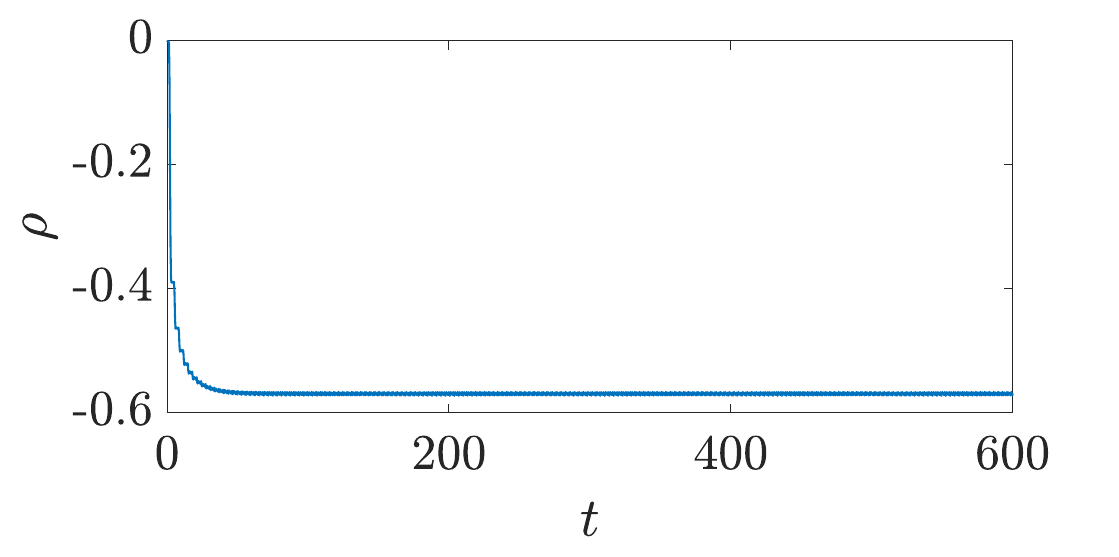}
    \caption{Polar log - after}
    \label{fig_norm_after_polar_log}
  \end{subfigure}
\end{center}
\caption{\label{fig_normalization}Various normalization of the signal. Left column: {\em far} from the fold, center column: {\em close} to the fold, right column: {\em after} the fold. Rows indicate, in order, the original signal, min-max normalization, transformation into polar form and normalization, transformation into polar form and logarithmic scale after normalization. Trajectories were obtained for the system in (\ref{eq_EOM_NLD}), with $c_1=0.1$, $c_3=0.0889$, $c_3=0.4222$ and $c_3=0.4889$; $c_{3\text{cr}}=0.444$.
 The length of the time series in the figure does not strictly reflect the length of the time series used for the training.}
\end{figure}

We implemented several normalization approaches to investigate the importance of providing data in the correct form. This is an essential step for providing physical insight to the CNN and enabling extrapolation capabilities.
In a recent work \cite{habib2023predicting}, we illustrated that time series close to a fold bifurcations dissipate energy in a particular way, which is not logarithmic as for linear systems.
Conversely, energy dissipation slows down in the vicinity of the fold.
This phenomenon can be understood comparing Figs.~\ref{fig_norm_far_original} and \ref{fig_norm_close_original}.
By depicting the logarithmic decrement of the oscillation amplitudes and measuring its minimal value during the decay, it is possible to predict the fold bifurcation, as described in \cite{habib2023predicting}.
According to this procedure, the oscillation frequency can be disregarded, while only the oscillation amplitude trend is relevant.
Besides, amplitude represented in a logarithmic scale better visualizes the phenomenon occurring near the fold.

In order to provide this information to the CNN, we performed various types of normalization, as described below:
\begin{enumerate}
\item {\bf Min-max normalization}: the signal is scaled such that the largest value is 1 and the smallest one is -1, as illustrated in Figs.~\ref{fig_norm_far_minmax}-\ref{fig_norm_after_minmax}. This normalization does not provide any physical insight to the CNN, but it helps deal with systems with very different amplitude ranges. It is a standard way of normalizing data and is unrelated to any physical insight of the phenomenon.
\item {\bf Polar transformation and normalization}: system coordinates are transformed into polar form according to the formula $\rho=\sqrt{x^2+\dot x^2}$, then $\rho$ is scaled such that its maximum value is 1. The $\rho$ signal provided to the network has a reduced information about the system frequency. This can be beneficial as frequency is not strictly related to the vicinity of the fold bifurcation.
Figures \ref{fig_norm_far_polar}-\ref{fig_norm_after_polar} illustrate the transformed signals. Far from the fold (Fig.~\ref{fig_norm_far_polar}), the signal decreases regularly until zero; close to the fold (Fig.~\ref{fig_norm_close_polar}), it presents a sort of bump before settling to 0; after the fold (Fig.~\ref{fig_norm_after_polar}) it rapidly converges to a constant value different from zero.
\item {\bf Polar transformation, normalization and moving mean}: similar to the previous normalization, with an additional moving mean to filter high-frequency variations of the signal. The adopted moving mean has a range of 5~\% of the entire signal.
\item {\bf Polar transformation, normalization and logarithmic scale}: system coordinates are transformed in polar form according to the formula $\rho=\sqrt{x^2+\dot x^2}$, then $\rho$ is scaled such that its maximum value is 1, and finally it is transformed to a natural logarithmic scale, as illustrated in Figs.~\ref{fig_norm_far_polar_log}-\ref{fig_norm_after_polar_log}. This normalization is similar to the previous one (without moving mean), with the additional feature that far from the bifurcation the signal is almost reduced to a straight line (it would be perfectly straight in the case of a linear single-DoF system). Besides, the signal does not converge to zero but to a negative number because of the logarithmic scale.
\item {\bf Polar transformation, normalization, logarithmic scale and moving mean}: similar to the previous case, with the additional moving mean filter to eliminate high-frequency content. The adopted moving mean has a range of 5~\% of the entire signal.
\end{enumerate}

\subsection{Training}\label{sec_training}

\begin{figure}
\begin{center}
\setlength{\unitlength}{\textwidth}
\includegraphics[width=0.6\textwidth]{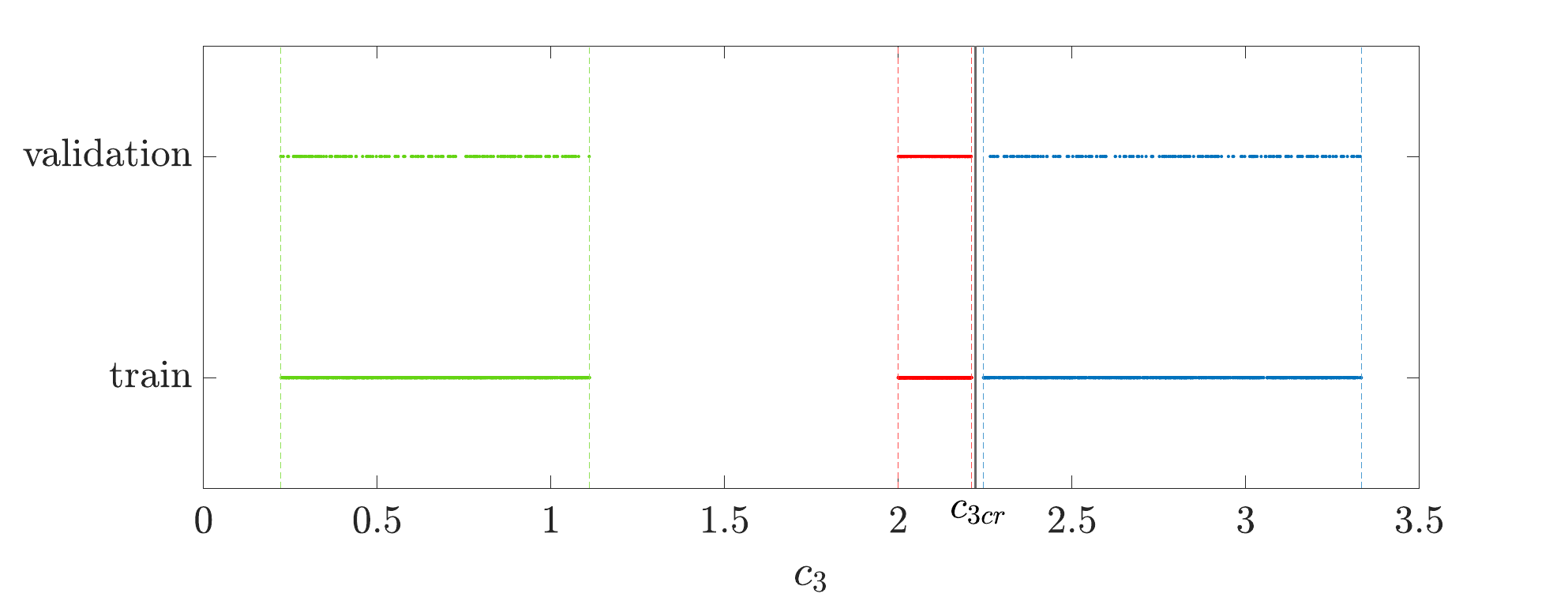}
\end{center}
\caption{\label{fig_data_info}Graphical representation of the parameter values utilized for training and validation. Vertical lines indicate boundary of regions, dots mark selected values for training and validation. Green: \emph{far}, red: \emph{close}, blue: \emph{after}.}
\end{figure}

As already mentioned, the only system used for the training was the single-DoF oscillator with nonlinear damping, whose dynamics is described by Eq.~(\ref{eq_EOM_NLD}).
Only 3000 time series were used for the training, 1000 for each class: {\em far}, {\em close}, and {\em after} the fold.
$c_1$ was fixed at 0.5, leading to $c_3^*=2.22$.
The value of $c_3$ for each class was randomly defined in the following ranges: far from the fold $c_3\in c_3^*\cdot(0.1,\,0.5)$, close to the fold $c_3\in c_3^*\cdot\left(0.9,\,0.995\right)$, and after the fold $c_3\in c_3^*\cdot\left(1.01,\,1.5\right)$.
Initial conditions were randomly selected in the range $x(0)\in\left(0.7,\,2\right)$ and $\dot x(0)\in\left(0.7,\,2\right)$.
All random quantities had a uniform distribution within the given interval.
Figure \ref{fig_data_info} offers a visual representation of the three categories. \emph{Far} and \emph{close} regions are purposely well separated since any boundary between them would be arbitrary. Besides, the separation facilitates the training of the NNs, enhancing the difference between the two categories.
Conversely, \emph{close} and \emph{after} regions are clearly separated by the fold bifurcation; therefore, they are very close to each other. A small gap between them is still present (between $0.995\,c_3^*$ and $1.01\,c_3^*$) because trajectories too close to the fold might take an excessive amount of time to reach the trivial solution, making them undistinguishable from trajectories after the fold.
Time series for training lasted a maximum of 700 time units; however, trajectories where interrupted earlier if for 7 subsequent time units their displacement absolute value was smaller than $10^{-8}$. The time series were then re-sampled in 1000 values regardless of their length.

We remark that the lack of normalization regarding the signal length makes it practically impossible for the NN to identify the signal's frequency. This choice reflects the practical difficulty in real applications of selecting signals with the same time length normalized with respect to the system's frequency. Nevertheless, as discussed in Sect. \ref{sect_discussion}, the signal frequency has almost no correlation with the classification proposed for the time series (\emph{far}, \emph{close} or \emph{after} the fold). In fact, for all normalization proposed, except min-max, the information about the system frequency is purposely removed. The only normalization that might be significantly affected by the lack of signal length normalization is the min-max; however, performing the same analysis discussed below with time-length normalized signal did not improve the min-max normalized network performance. The analysis is omitted here for the sake of brevity.

\section{Results}

Networks were trained with the same training data for all types of normalization.
In all cases, 100~\% accuracy was obtained during training and validation. This result suggests that the NN architecture and the amount of data are adequate for the task.
Indeed, 100~\% accuracy is quite unusual for any machine learning problem. In the authors' opinion, two main factors lead to this result. First of all, the validation set is relatively small, encompassing only 600 samples. However, considering that the training data included only 3000 time series, it seems appropriate. The second reason is that the trajectories are very similar to each other within each class, and very different between classes. Most probably, reducing the gap between \emph{far} and \emph{close} regions would reduce the accuracy of the NNs validation. Figure \ref{fig_data_info} illustrates that the validation data cover the whole $c_3$ range of each class.

\begin{table}[b]
	\centering
	\begin{tabular}{ccccc}
		\hline 
		\hline
		\makecell{\bf{Type of}\\ \bf{normalization}}
		 &  \makecell{\bf{Nonlinear}\\ \bf{damping}\\ \bf{$c_1=0.1$}} & \makecell{\bf{Mass-on}\\ \bf{moving-belt}} & \makecell{\bf{van der Pol}\\ \bf{-Duffing} }& \makecell{\bf{Pitch {\&}}\\ {\bf plunge} }\\ 
		\hline 
		Min-max & 66.6~\% & wrong & wrong & no {\em close}\\[-3mm]\\
		Polar & 87.5~\% & no {\em close} & good & good\\[-3mm]\\
		Pol-MovMean & 87.33~\% & no {\em close} & good & good\\[-3mm]\\
		Pol-log & 100~\% & wrong & good & inaccurate\\[-3mm]\\
		Pol-log-MovMean & 100~\% & good & good & good\\
		\hline 
		\hline
	\end{tabular} 
	\caption{Summary of the performance of the networks for different normalizations.}
	\label{tab_results}
\end{table}
Providing correct classification for other parameter ranges is significantly more challenging.
Different tests were defined, namely:
\begin{itemize}
\item Classify trajectories for the same system utilized during training but with $c_1=0.1$ instead of 0.5.
\item Classify trajectories for the same system utilized during training but with $c_1\in\left(0.05,\ 0.7\right)$. For this task, the time series duration differs from the training data, making the system different from the one used during training from the network perspective.
\item Classify trajectories for the mass-on-moving-belt system.
\item Classify trajectories for the van der Pol-Duffing oscillator with an attached dynamic vibration absorber.
\item Classify trajectories for the pitch-and-plunge wing profile.
\end{itemize}
For the first task, results can be quantified with a percentage of accuracy, while only qualitative assessments are given for the other tasks because of the arbitrarity of the boundary between \emph{far} and \emph{close} regions. Results are summarized in Table \ref{tab_results}, while their comprehensive discussion is provided below.
In the table, the following qualitative labels are provided.
If {\em far} and {\em after} trajectories are significantly wrongly classified, the network is judged as ``wrong"; if {\em close} trajectories are misclassified, then the label ``no {\em close}" is given, if the network is overall correct, but some errors are present it is marked as ``inaccurate"; while if the network provides correct classification, it is marked as ``good".
\begin{table}[t]
	\centering
	\begin{tabular}{cccccc}
		\hline 
		\hline
		\makecell{\bf{Type of}\\ \bf{normalization}}
		 &  \makecell{\bf{Mass-on}\\ \bf{mov.-belt}} & \makecell{\bf{VdP}\\ \bf{Duf.~$x_1$} } & \makecell{\bf{VdP}\\ \bf{Duf.~$x_2$} }& \makecell{\bf{Pitch {\&}}\\ {\bf plunge $y$} } & \makecell{\bf{Pitch {\&}}\\ {\bf plunge $\alpha$} }\\ 
		\hline 
		Min-max & - & - & - & - & -\\[-3mm]\\
		Polar & - & 0.234 & 0.359 & 0.690 & 0.234\\[-3mm]\\
		Pol-MovMean & - & 0.199 & 0.352 & 0.695 & 0.227\\[-3mm]\\
		Pol-log & - & 0.504 & 0.640 &  - & -\\[-3mm]\\
		Pol-log-MovMean & 0.368 & 0.552 & 0.577 & 0.965 & 0.948\\[-3mm]\\
		\makecell{Pol-log-MovMean \\ (noisy test data)}  & - & 0.993 & 0.804 & 0.846 & 0.857\\[-3mm]\\
		\makecell{Pol-log-MovMean \\ (noisy test and\\ training data)}  & 0.722 & - & 0.497 & 0.447 & 0.169\\
		\hline 
		\hline
	\end{tabular} 
	\caption{Ratio between the overlapping area of estimated \emph{close}-\emph{far} regions and \emph{close} region.}
	\label{tab_overlapping}
\end{table}

The table illustrates that the only network providing good results for all systems is the last one, where data were first transformed in polar coordinates, then normalized between 0 and 1, transformed in natural logarithmic scale, and finally filtered through a moving mean.
However, as better detailed later, also omitting the logarithmic scale transformation but only using polar coordinates provided acceptable results, except for the mass-on-moving-belt system, possibly because of the non-smoothness of the system.
Conversely, the simple min-max normalization provided very bad results; in that case, the network was completely unable to extrapolate classification capabilities to other systems.
These observations clearly show that the polar transformation is the main factor for enabling the network to extrapolate results.

Since the boundary between \emph{far} and \emph{close} regions is not well defined, and the shape of the time series depends on their initial conditions, the NN provides a region of overlapping between \emph{far} and \emph{close} region, resulting in a fuzzy boundary. In order to quantify the overlapping, the ratio between the overlapping region and the full extent of the estimated \emph{close} region is calculated, as indicated in Table \ref{tab_overlapping} for all types of normalization implemented. The smaller this number, the better the boundary between the two regions is defined.
Interestingly, we note that, although the final pre-processing strategy (transformation into polar coordinates, normalization, transformation into a logarithmic scale, and filtering via a moving average) was the only one able to correctly classify trajectories from all systems, it resulted in a much larger overlapping region, significantly larger than when the logarithmic transformation was not applied. This is clearly an undesirable outcome. As a result, a clear best normalization strategy cannot be defined, as discussed in Sect.~\ref{sect_discussion}.

\subsection{Min-max normalization}

\begin{figure}[h]
\begin{center}
\setlength{\unitlength}{\textwidth}
\begin{subfigure}[b]{0.35\textwidth}
    \includegraphics[width=\textwidth]{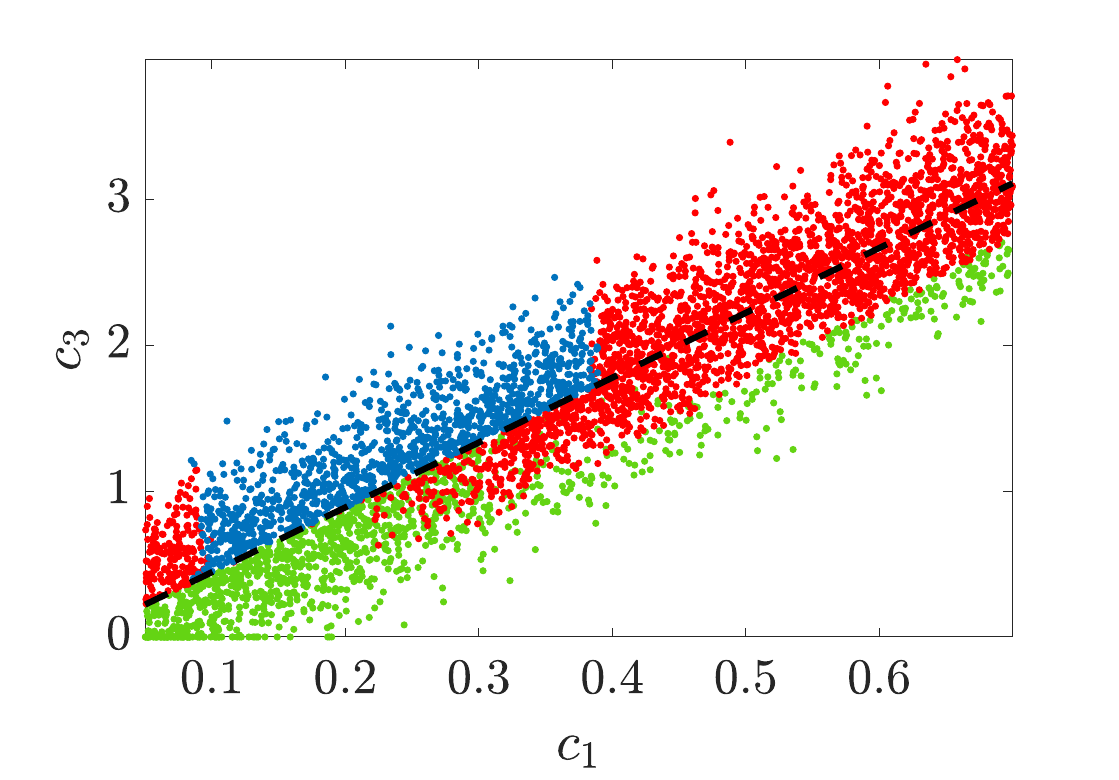}
    \caption{Nonlinear damping system}
    \label{fig_minmax_sparsi_NLD}
  \end{subfigure}
\begin{subfigure}[b]{0.35\textwidth}
    \includegraphics[width=\textwidth]{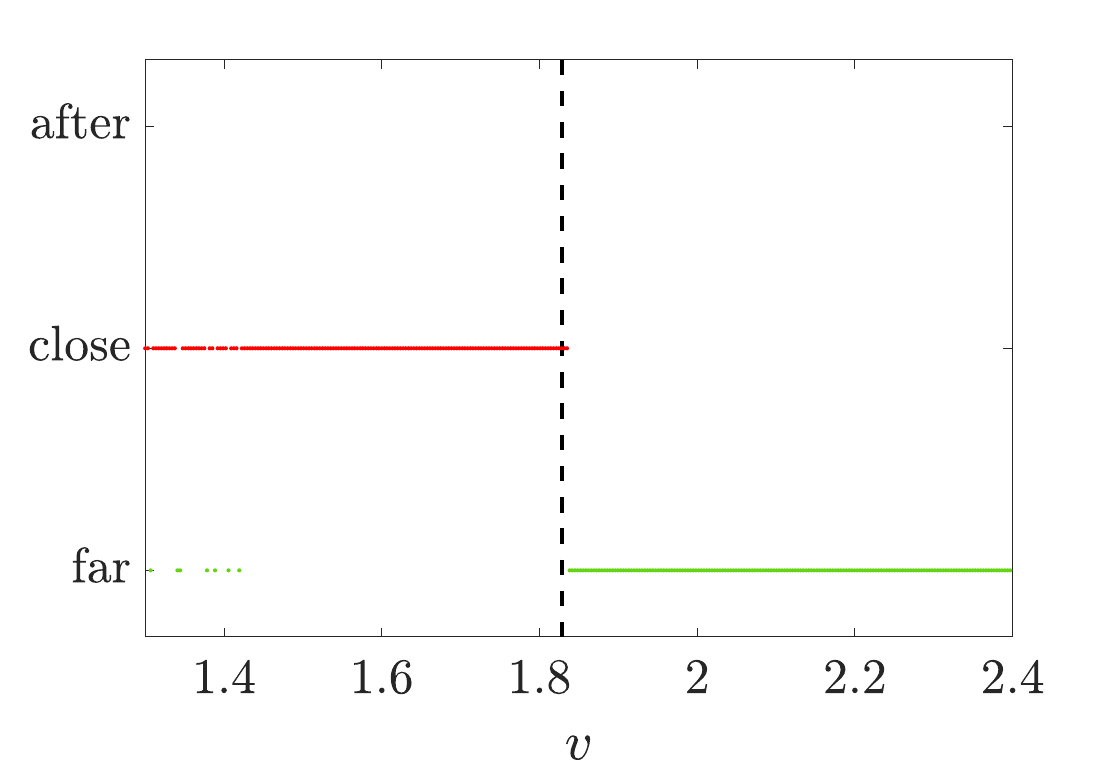}
    \caption{Mass-on-moving-belt}
    \label{fig_minmax_MOB}
  \end{subfigure}
\begin{subfigure}[b]{0.35\textwidth}
    \includegraphics[width=\textwidth]{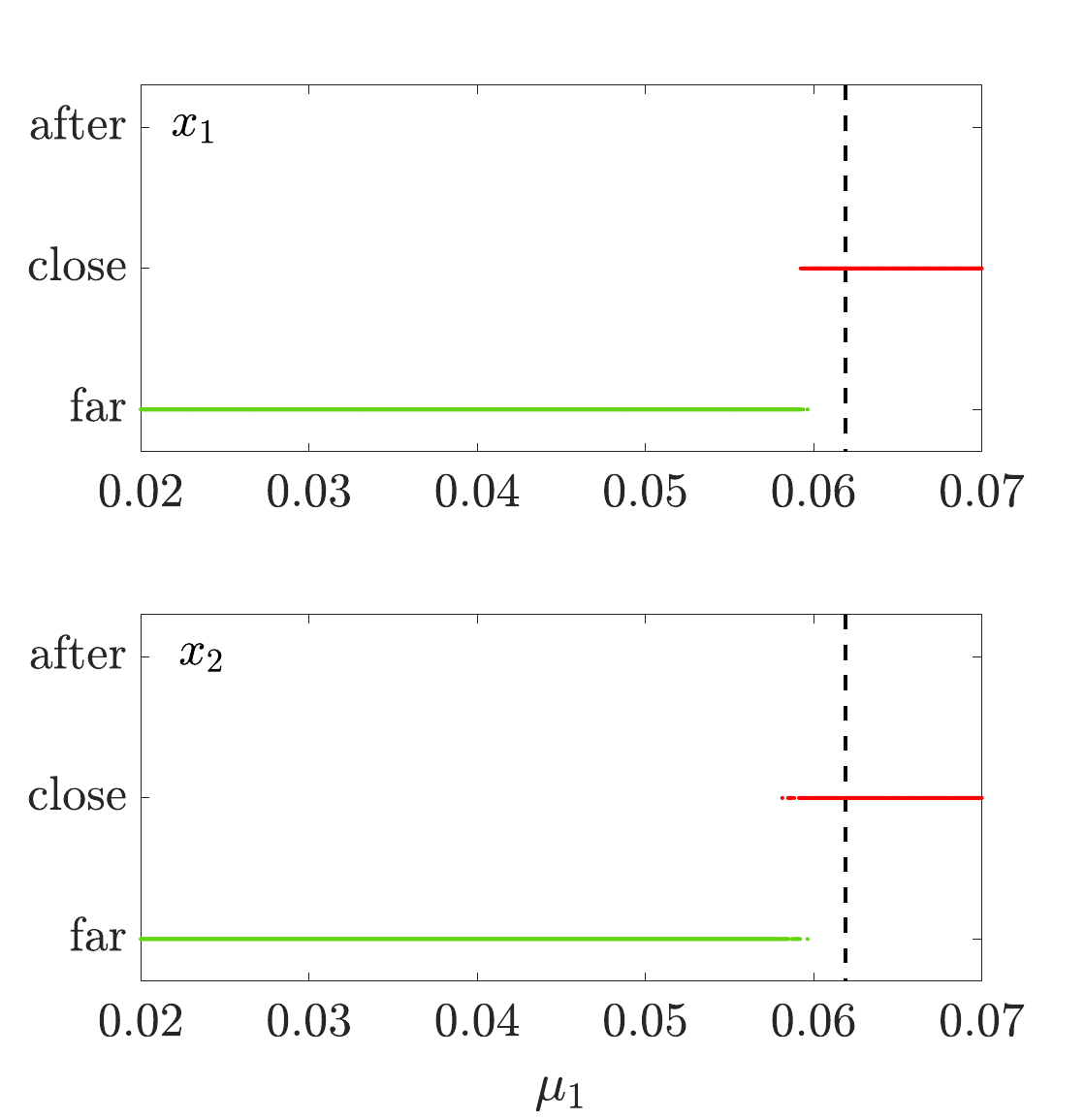}
    \caption{Van der Pol-Duffing oscillator}
    \label{fig_minmax_VDP}
  \end{subfigure}
\begin{subfigure}[b]{0.35\textwidth}
    \includegraphics[width=\textwidth]{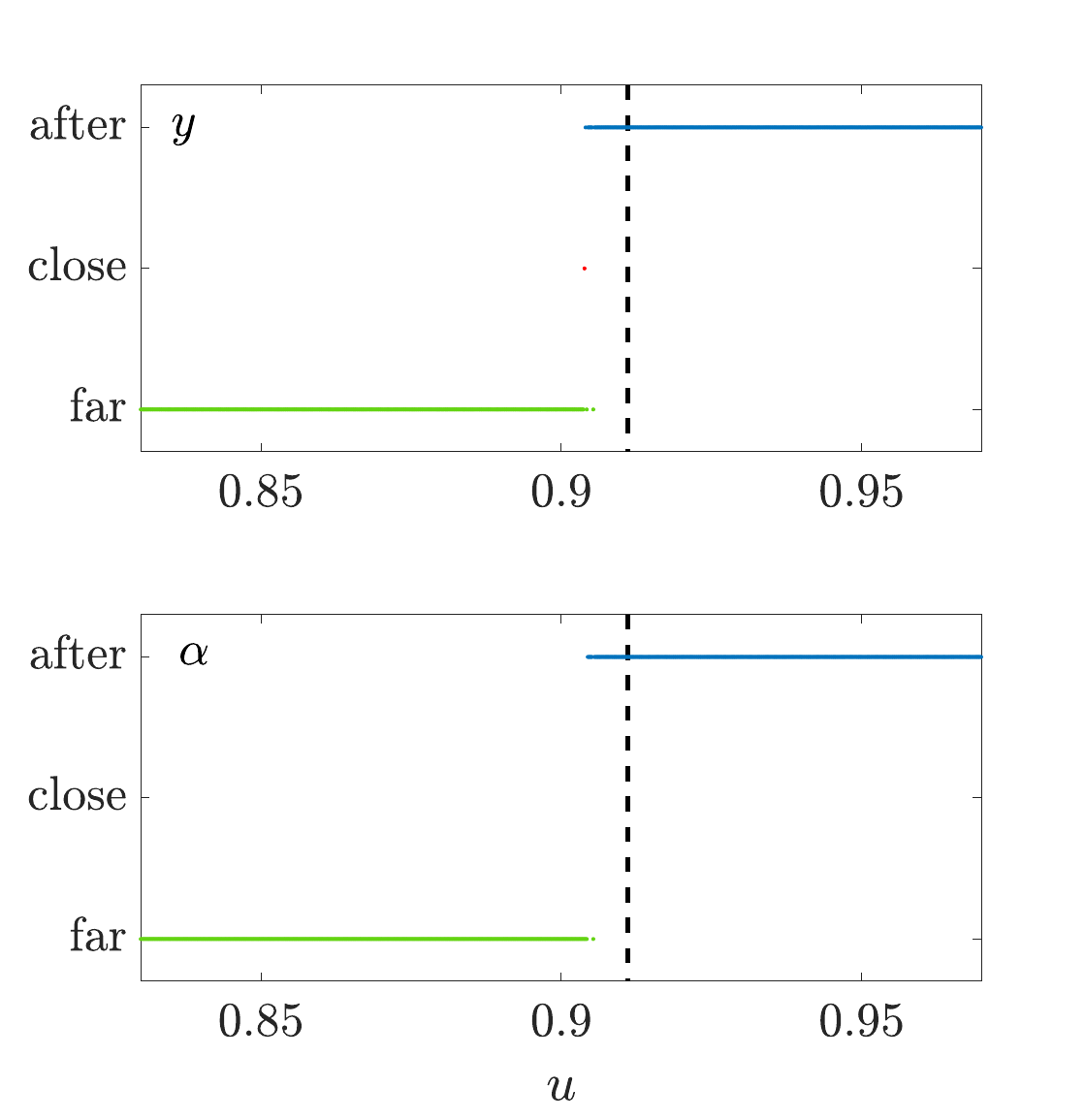}
    \caption{Pitch-and-plunge wing}
    \label{fig_minmax_PnP}
  \end{subfigure}
\end{center}
\caption{\label{fig_minmax_results}Prediction diagrams obtained by the CNN trained with min-max normalized data. Each point corresponds to a different time series. Dashed lines indicate the fold bifurcation. The colors mark the predicted classification.
Color interpretation: green-{\em far}, red-{\em close}, blue-{\em after}.}
\end{figure}

The network trained with simply normalized data was completely unable to extrapolate classification capabilities.
Although it obtained a 100~\% accuracy in the validation set, simple variations of the $c_1$ parameter prevented it from providing meaningful results (Fig.~\ref{fig_minmax_sparsi_NLD}).
Regarding other systems investigated, prediction was always inadequate; we note that {\em far} trajectories were generally correctly classified, while {\em after} and {\em close} trajectories were often systematically confused with each other (Fig.~\ref{fig_minmax_results}).
Considering the simple normalization provided, this poor result is not surprising.

\subsection{Polar transformation and normalization}\label{sect_res_pol}

\begin{figure}[h]
\begin{center}
\setlength{\unitlength}{\textwidth}
\begin{subfigure}[b]{0.35\textwidth}
    \includegraphics[width=\textwidth]{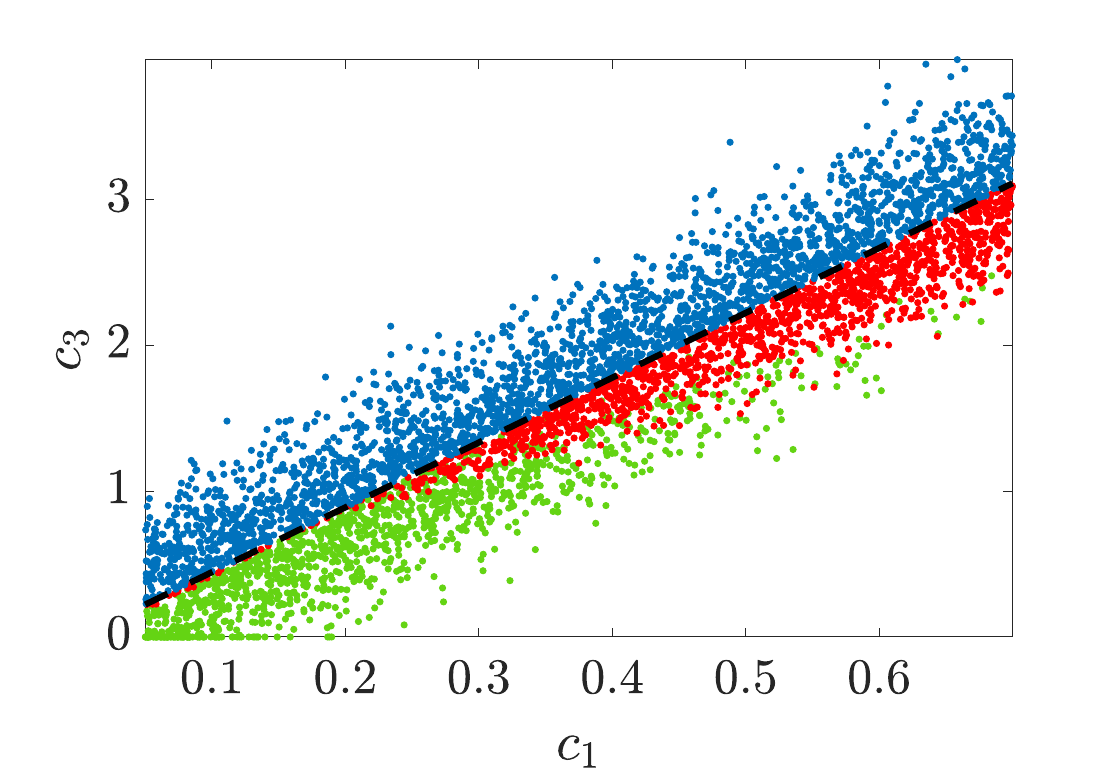}
    \caption{Nonlinear damping system}
    \label{fig_polar_01_sparsi_NLD}
  \end{subfigure}
\begin{subfigure}[b]{0.35\textwidth}
    \includegraphics[width=\textwidth]{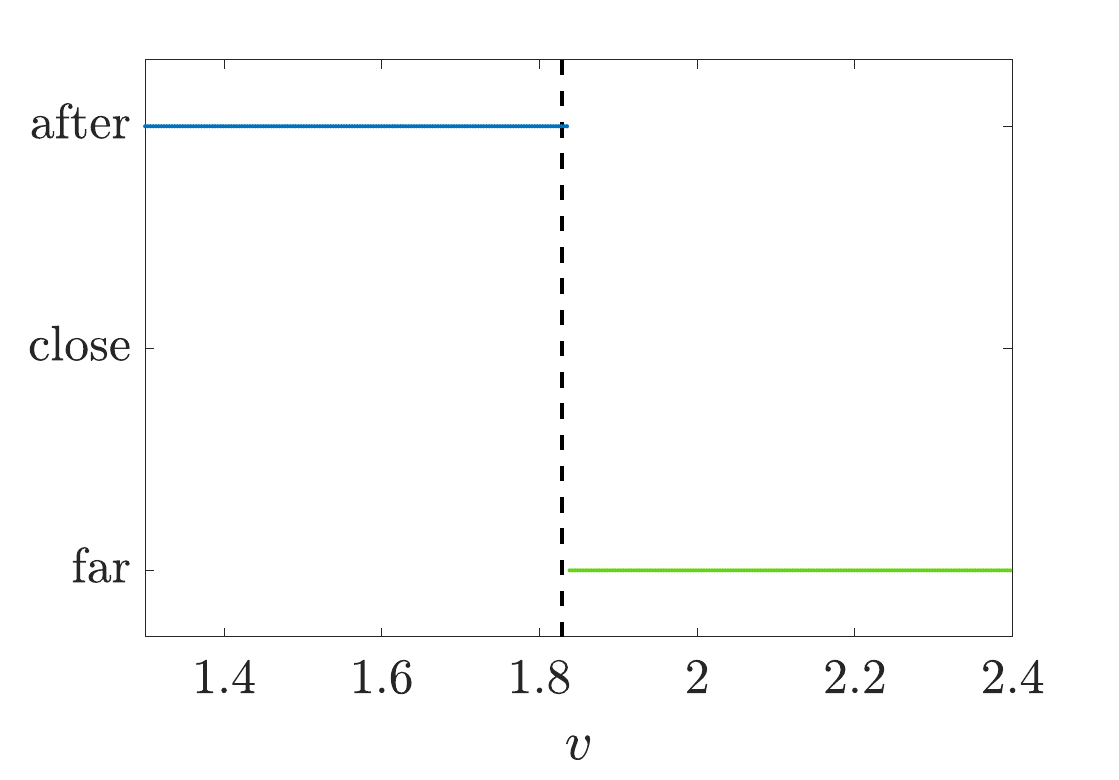}
    \caption{Mass-on-moving-belt}
    \label{fig_polar_01_MOB}
  \end{subfigure}
\begin{subfigure}[b]{0.35\textwidth}
    \includegraphics[width=\textwidth]{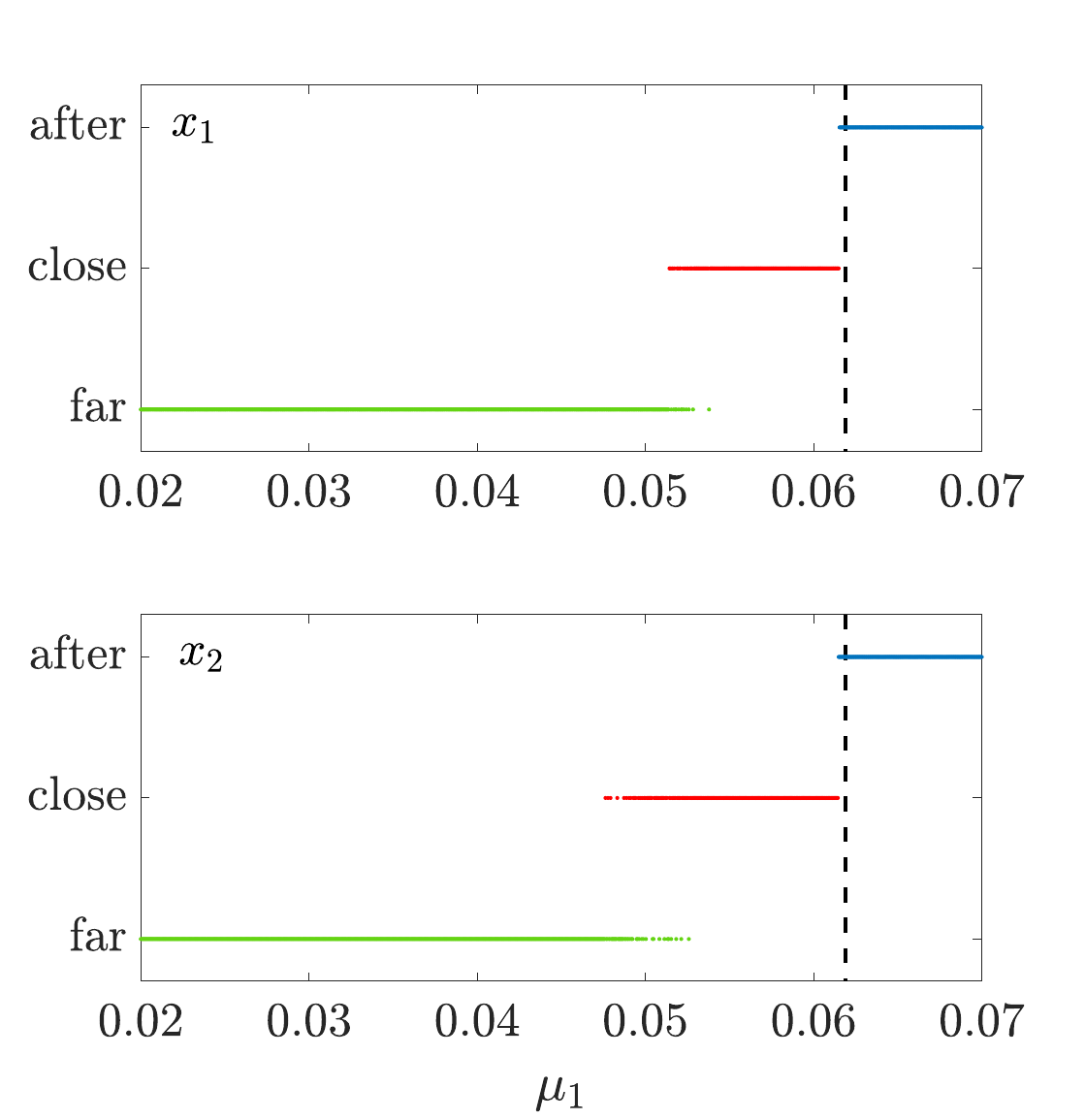}
    \caption{Van der Pol-Duffing oscillator}
    \label{fig_polar_01_VDP}
  \end{subfigure}
\begin{subfigure}[b]{0.35\textwidth}
    \includegraphics[width=\textwidth]{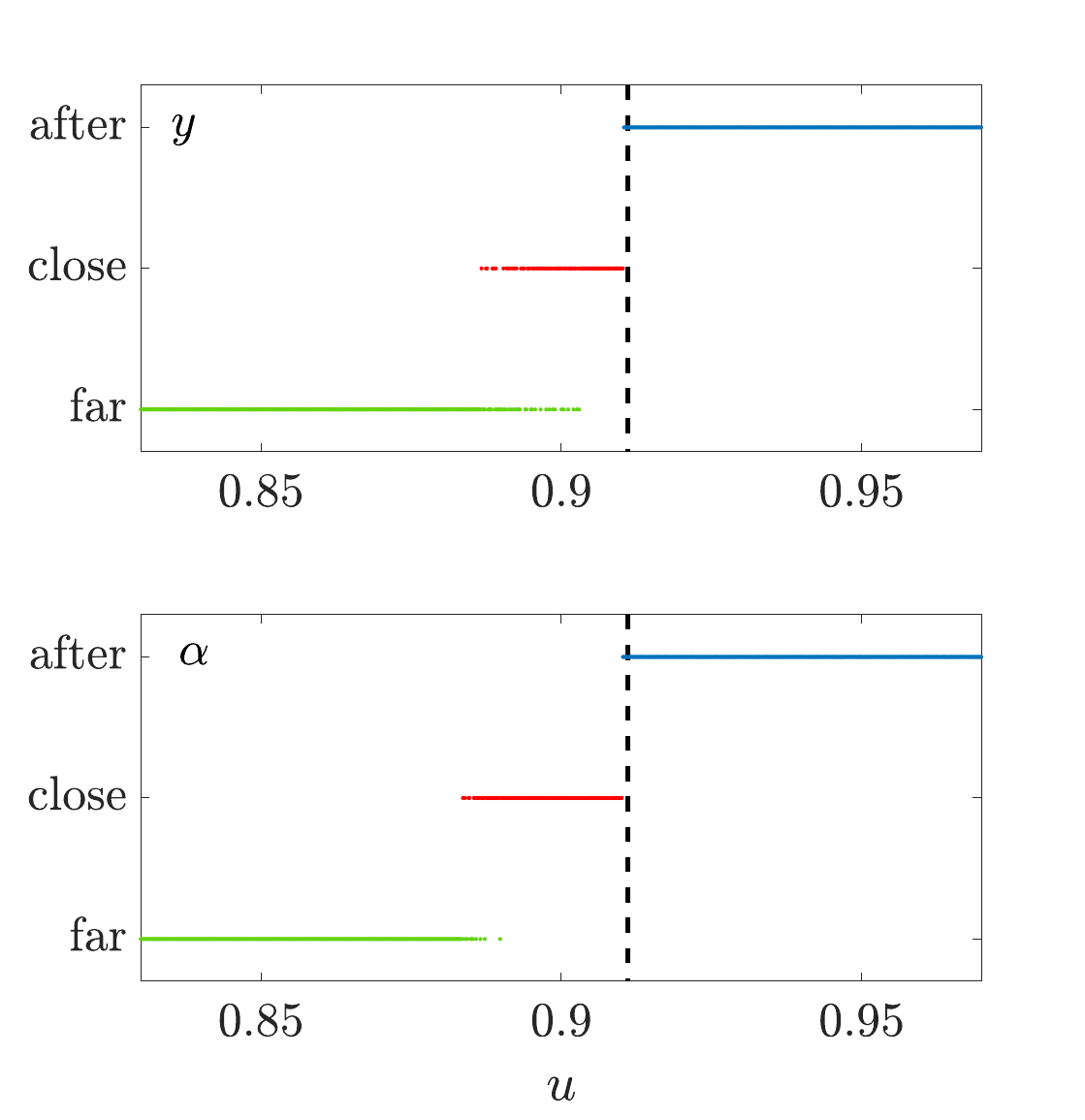}
    \caption{Pitch-and-plunge wing}
    \label{fig_polar_01_PnP}
  \end{subfigure}
\end{center}
\caption{\label{fig_polar_01_results}Prediction diagrams obtained by the CNN trained with data normalized in polar form. Each point corresponds to a different time series. Dashed lines indicate the fold bifurcation. The colors mark the predicted classification.
Color interpretation: green-{\em far}, red-{\em close}, blue-{\em after}.}
\end{figure}

The polar form transformation and amplitude normalization between 0 and 1 provided much better results.
This is mostly due to the almost complete elimination of the frequency of oscillation from the data, which is not related to general traits of system approaching a fold bifurcation.
We also note that the network never misclassified {\em far} and {\em after} trajectories.
Regarding the system with nonlinear damping (Fig.~\ref{fig_polar_01_sparsi_NLD}), reducing $c_1$ the network significantly reduced the trajectories classified as {\em close}, which are limited to those extremely close to the fold.
About the mass-on-moving-belt system (Fig.~\ref{fig_polar_01_MOB}), the network completely overlooked {\em close} trajectories, which were always classified as {\em far}. Regarding the van der Pol-Duffing oscillator (Fig.~\ref{fig_polar_01_VDP}) and the pitch-and-plunge wing profile (Fig.~\ref{fig_polar_01_PnP}) the classification was excellent with both degrees of freedom.
The overlapping region between \emph{far} and \emph{close} is also relatively small, except for the pitch-and-plunge if computed from $y$ (Table \ref{tab_overlapping}).

\subsection{Polar transformation, normalization and moving mean filter}\label{sect_res_pol_mvm}

\begin{figure}[h]
\begin{center}
\setlength{\unitlength}{\textwidth}
\begin{subfigure}[b]{0.35\textwidth}
    \includegraphics[width=\textwidth]{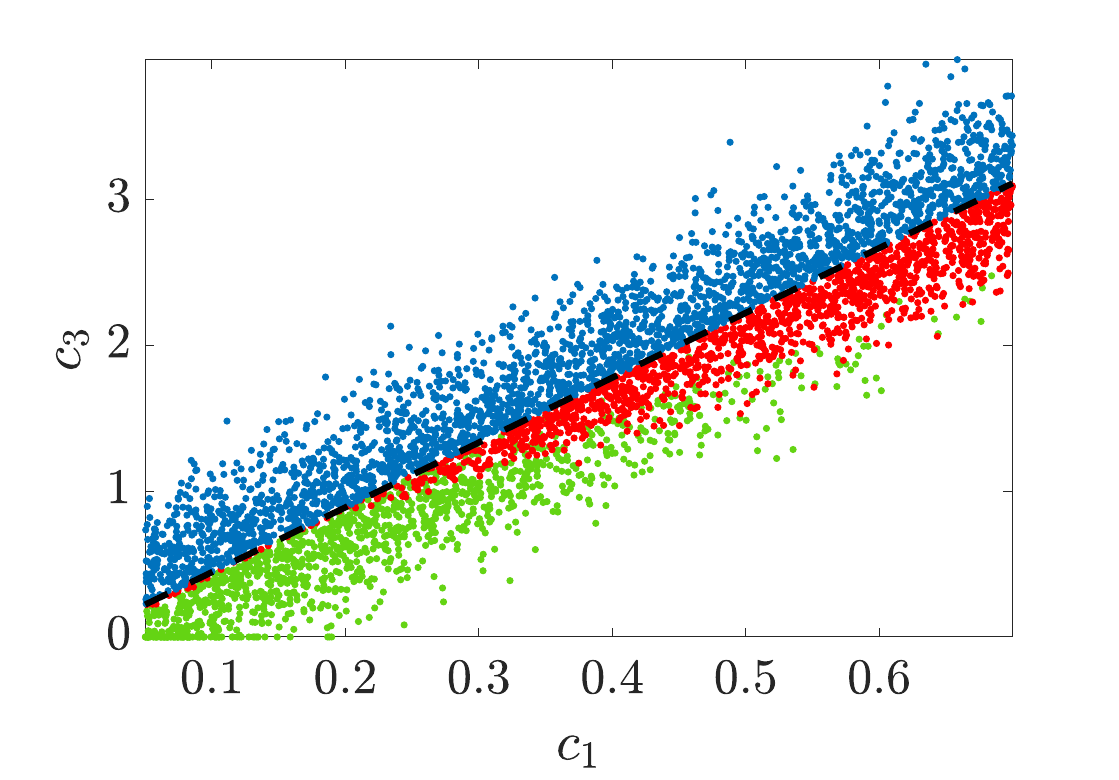}
    \caption{Nonlinear damping system}
    \label{fig_polar_01_mvm_sparsi_NLD}
  \end{subfigure}
\begin{subfigure}[b]{0.35\textwidth}
    \includegraphics[width=\textwidth]{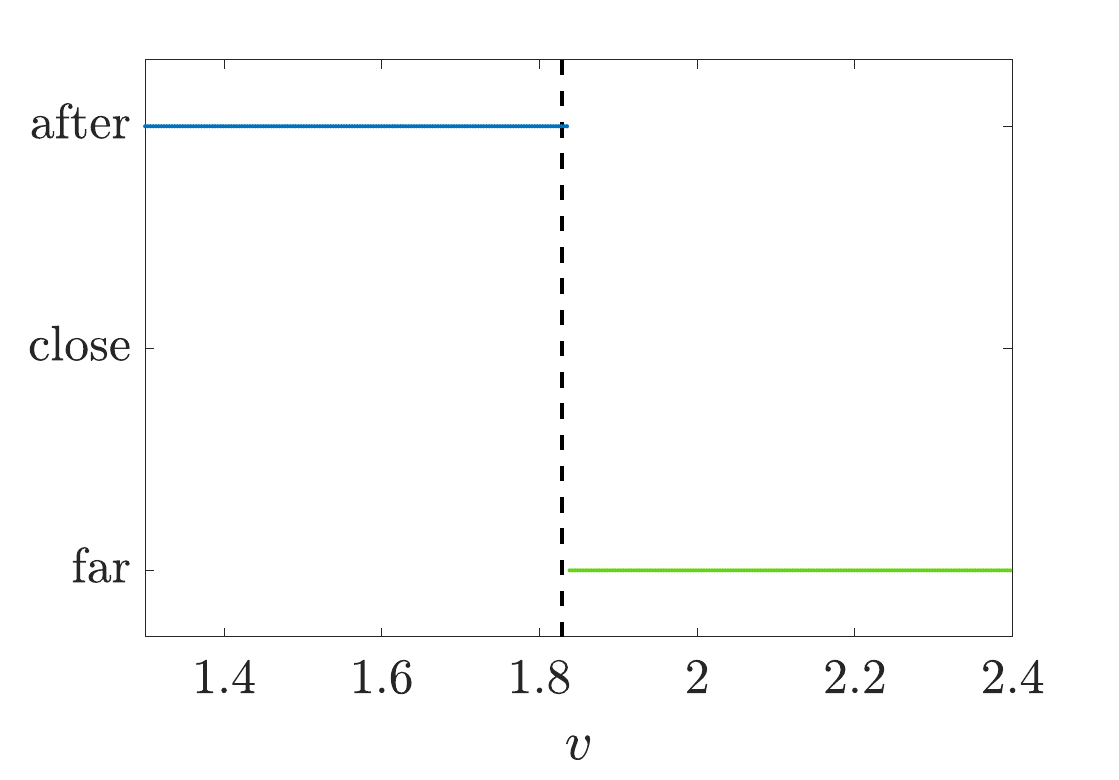}
    \caption{Mass-on-moving-belt}
    \label{fig_polar_01_mvm_MOB}
  \end{subfigure}
\begin{subfigure}[b]{0.35\textwidth}
    \includegraphics[width=\textwidth]{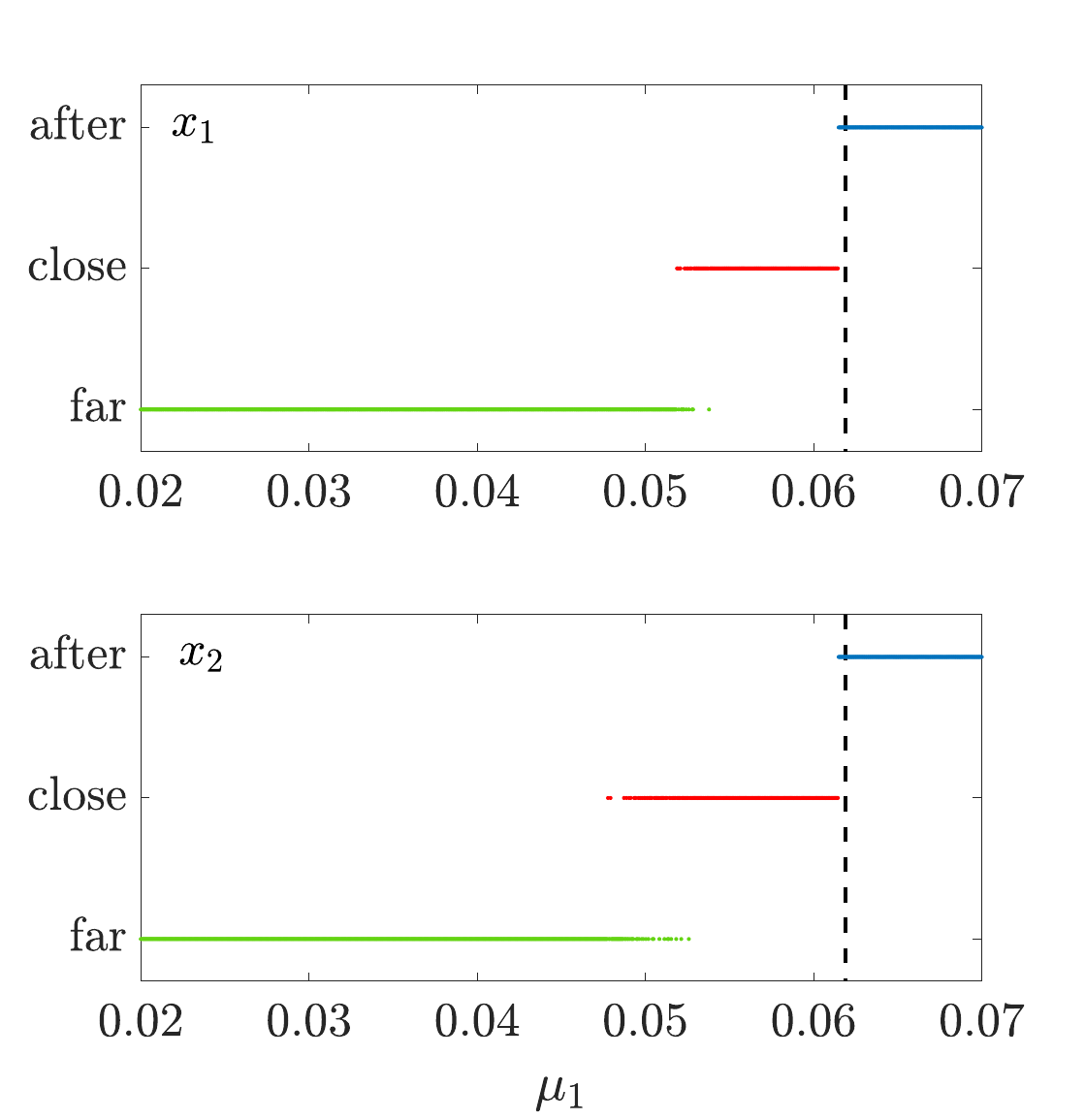}
    \caption{Van der Pol-Duffing oscillator}
    \label{fig_polar_01_mvm_VDP}
  \end{subfigure}
\begin{subfigure}[b]{0.35\textwidth}
    \includegraphics[width=\textwidth]{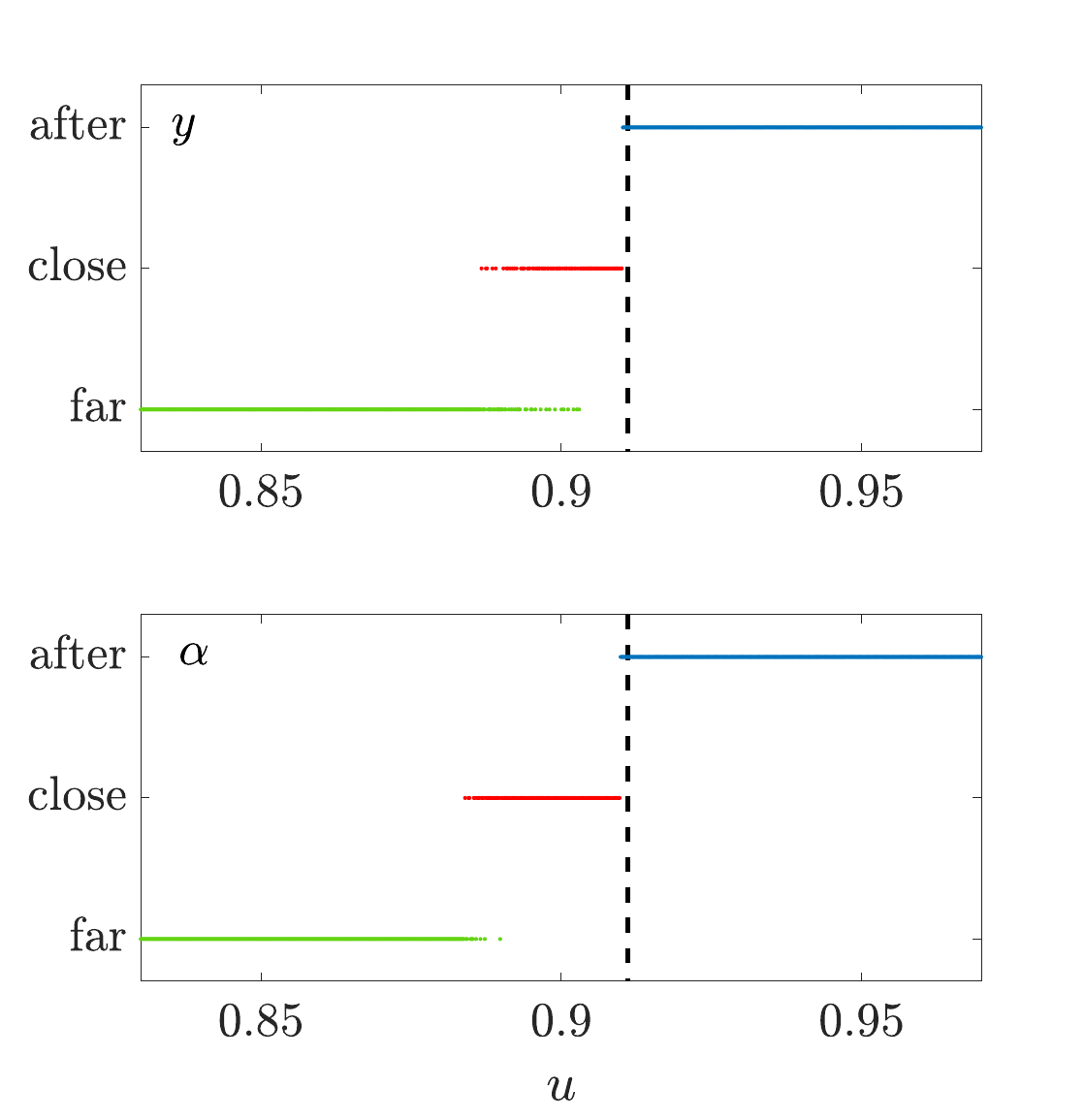}
    \caption{Pitch-and-plunge wing}
    \label{fig_polar_01_mvm_PnP}
  \end{subfigure}
\end{center}
\caption{\label{fig_polar_01_mvm_results}Prediction diagrams obtained by the CNN trained with data normalized in polar form filtered with a moving mean. Each point corresponds to a different time series. Dashed lines indicate the fold bifurcation. The colors mark the predicted classification.
Color interpretation: green-{\em far}, red-{\em close}, blue-{\em after}.}
\end{figure}

Adding a moving mean filter to the previous case (polar transformation and normalization) did not provide any advantage.
This network provided practically the same results as the previous one. Namely, acceptable results for the nonlinear damping oscillator (Fig.~\ref{fig_polar_01_mvm_sparsi_NLD}), missed {\em close} trajectories in the mass-on-moving-belt system (Fig.~\ref{fig_polar_01_mvm_MOB}), excellent results for the van der Pol-Duffing oscillator (Fig.~\ref{fig_polar_01_mvm_VDP}) and pith-and-plunge wing (Fig.~\ref{fig_polar_01_mvm_PnP}).
Indeed, this is not surprising considering that the CNN already includes filters that can provide a similar effect as the moving mean filter.
However, this is not always the case, as shown in Sect.~\ref{sect_res_pol_log_mvm}.

\subsection{Polar transformation, normalization and logarithmic scale}

\begin{figure}[h]
\begin{center}
\setlength{\unitlength}{\textwidth}
\begin{subfigure}[b]{0.35\textwidth}
    \includegraphics[width=\textwidth]{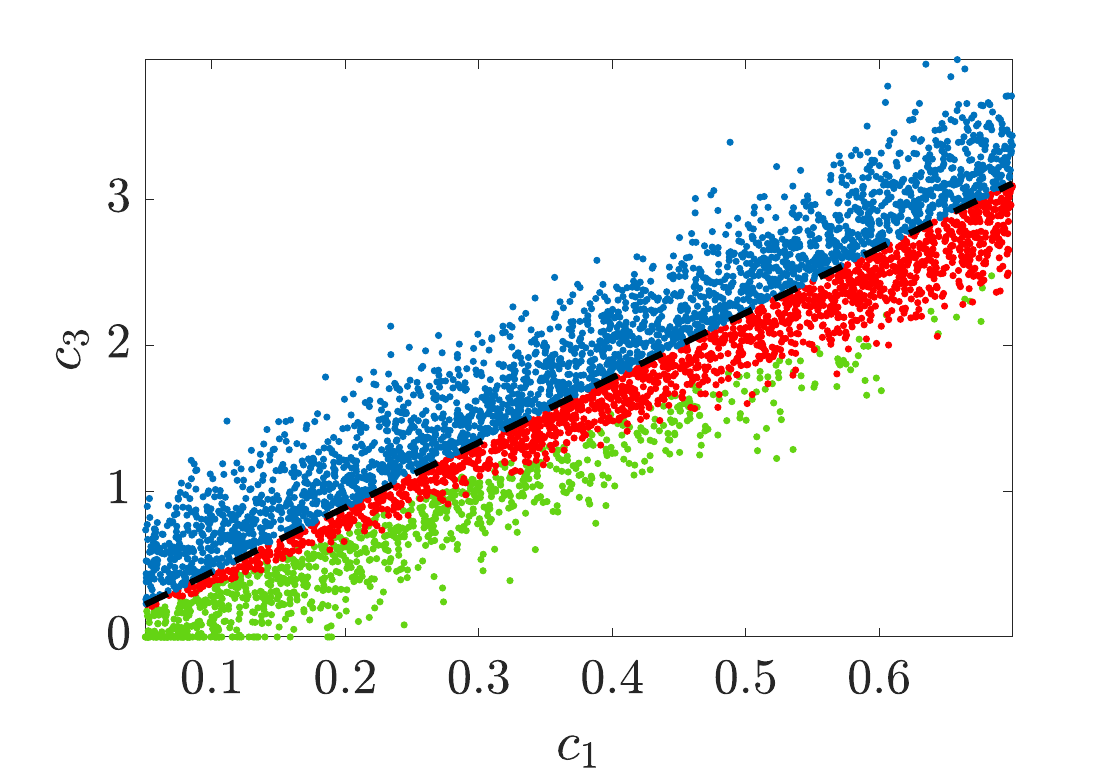}
    \caption{Nonlinear damping system}
    \label{fig_polar_01_log_sparsi_NLD}
  \end{subfigure}
\begin{subfigure}[b]{0.35\textwidth}
    \includegraphics[width=\textwidth]{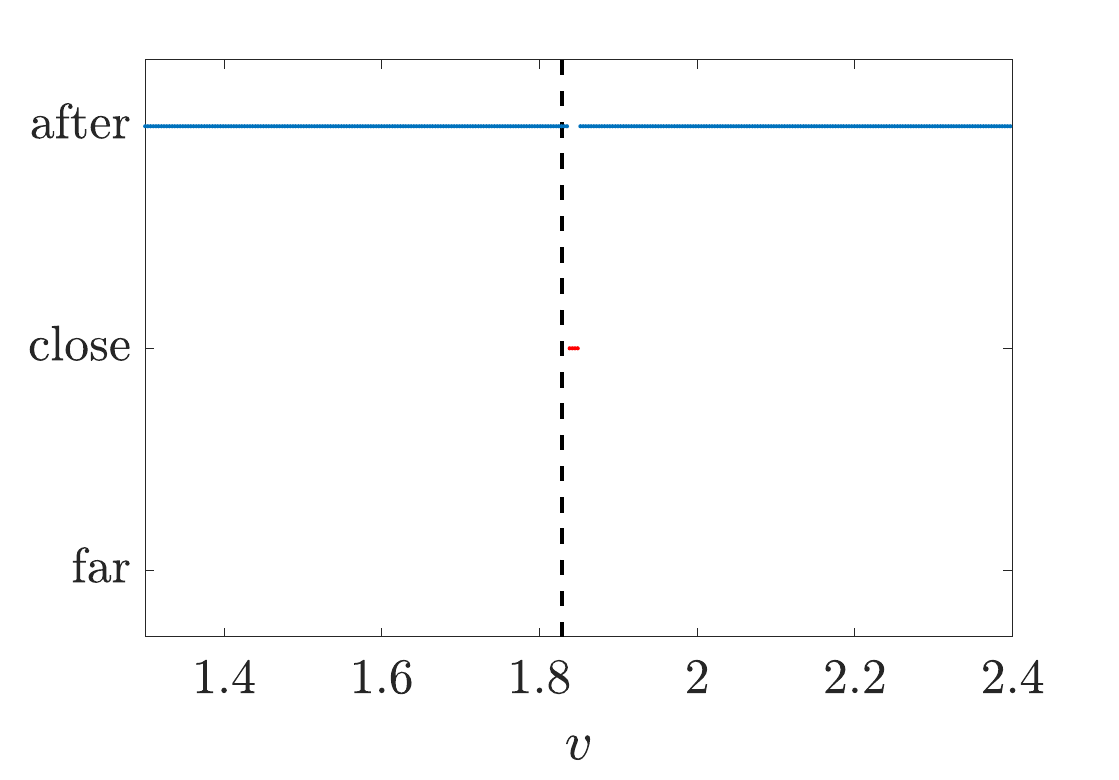}
    \caption{Mass-on-moving-belt}
    \label{fig_polar_01_log_MOB}
  \end{subfigure}
\begin{subfigure}[b]{0.35\textwidth}
    \includegraphics[width=\textwidth]{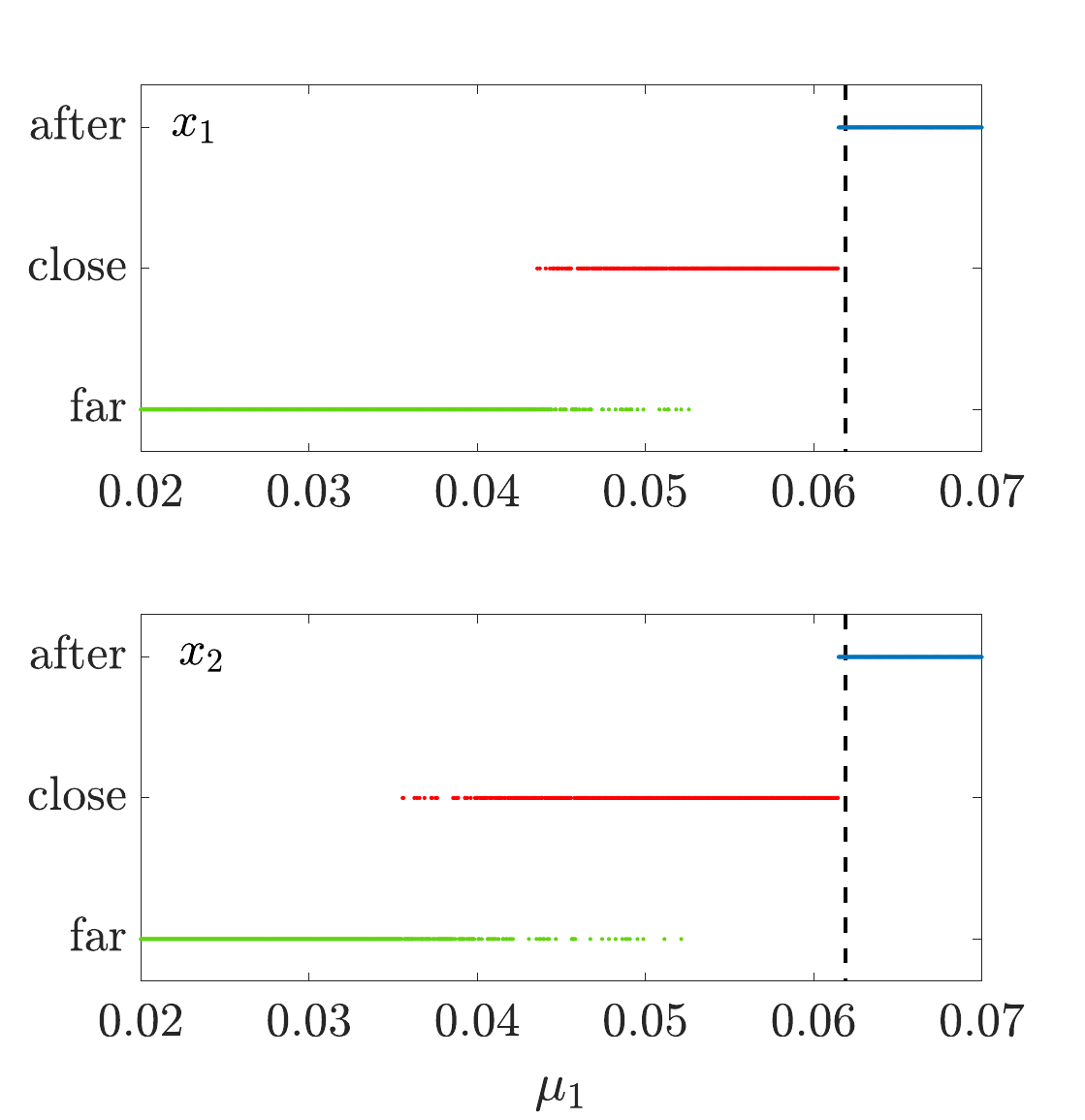}
    \caption{Van der Pol-Duffing oscillator}
    \label{fig_polar_01_log_VDP}
  \end{subfigure}
\begin{subfigure}[b]{0.35\textwidth}
    \includegraphics[width=\textwidth]{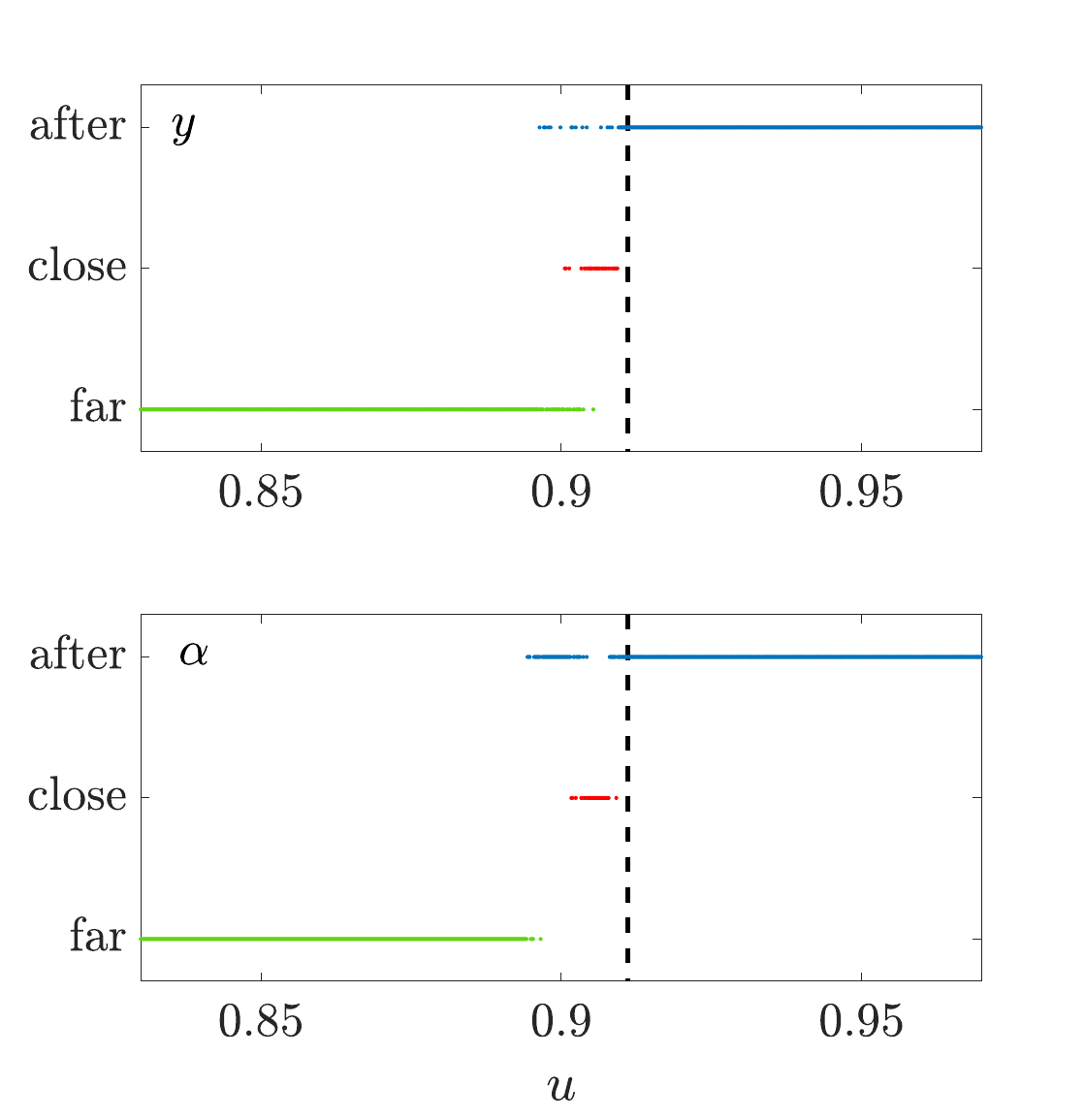}
    \caption{Pitch-and-plunge wing}
    \label{fig_polar_01_log_PnP}
  \end{subfigure}
\end{center}
\caption{\label{fig_polar_01_log_results}Prediction diagrams obtained by the CNN trained with data normalized in polar form and in logarithmic scale. Each point corresponds to a different time series. Dashed lines indicate the fold bifurcation. The colors mark the predicted classification.
Color interpretation: green-{\em far}, red-{\em close}, blue-{\em after}.}
\end{figure}

The transformation into a logarithmic scale reflects the observation that the decrement of the oscillation amplitude is logarithmic in a linear system.
Accordingly, if far from the bifurcation the nonlinearities are weak, the amplitude is expected to decrease approximately logarithmically (of course, modal interaction modifies this behavior), which in a logarithmic scale appears linear.
Training the network with data in polar coordinates, normalized and scaled in logarithmic coordinates provided better results than seen in the previous case for the nonlinearly damped oscillator (Fig.~\ref{fig_polar_01_log_sparsi_NLD}), which were excellent for all the investigated range of $c_1$.
However, the network provided a completely wrong classification for the mass-on-moving-belt system (Fig.~\ref{fig_polar_01_log_MOB}); for this system, all {\em far} cases were incorrectly classified as {\em after}. Interestingly, some {\em close} trajectories were correctly classified.
It is hard to understand the mechanism leading to such a wrong classification.
Regarding the van der Pol-Duffing oscillator (Fig.~\ref{fig_polar_01_log_VDP}), the network performed well; in this case, the region of {\em close} trajectories was much larger than for the networks whose data were not in logarithmic scale (Figs.~\ref{fig_polar_01_VDP} and \ref{fig_polar_01_mvm_VDP}); the prediction based on the primary system displacement ($x_1$) provided a smaller {\em close} region than if based on the vibration absorber displacement ($x_2$). However, the overlapping region is much larger than in the previous cases (Table \ref{tab_overlapping}).
About the pitch-and-plunge wing profile, the classification is acceptable; however, some time series in the {\em close} region were mistakenly classified as {\em after}; this problem is less significant if classification is based on plunge motion ($y$), rather than pitch ($\alpha$).

\subsection{Polar transformation, normalization, logarithmic scale and moving mean} \label{sect_res_pol_log_mvm}

\begin{figure}[h]
\begin{center}
\setlength{\unitlength}{\textwidth}
\begin{subfigure}[b]{0.35\textwidth}
    \includegraphics[width=\textwidth]{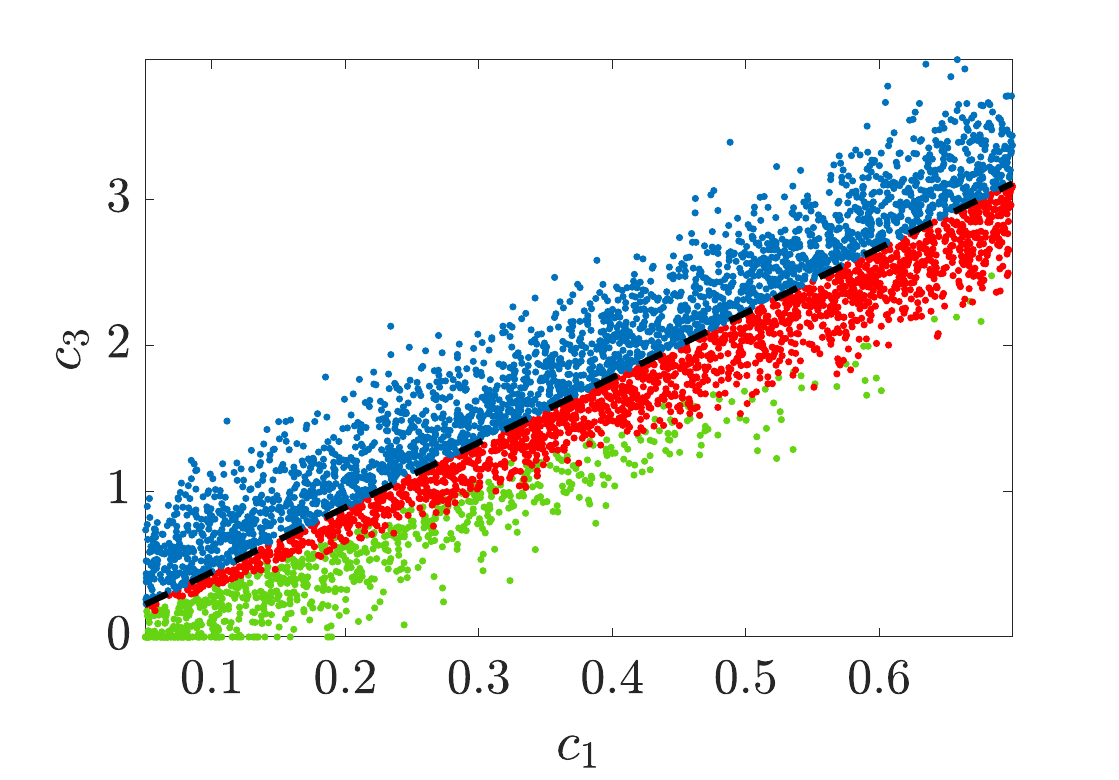}
    \caption{Nonlinear damping system}
    \label{fig_polar_01_log_mvm_sparsi_NLD}
  \end{subfigure}
\begin{subfigure}[b]{0.35\textwidth}
    \includegraphics[width=\textwidth]{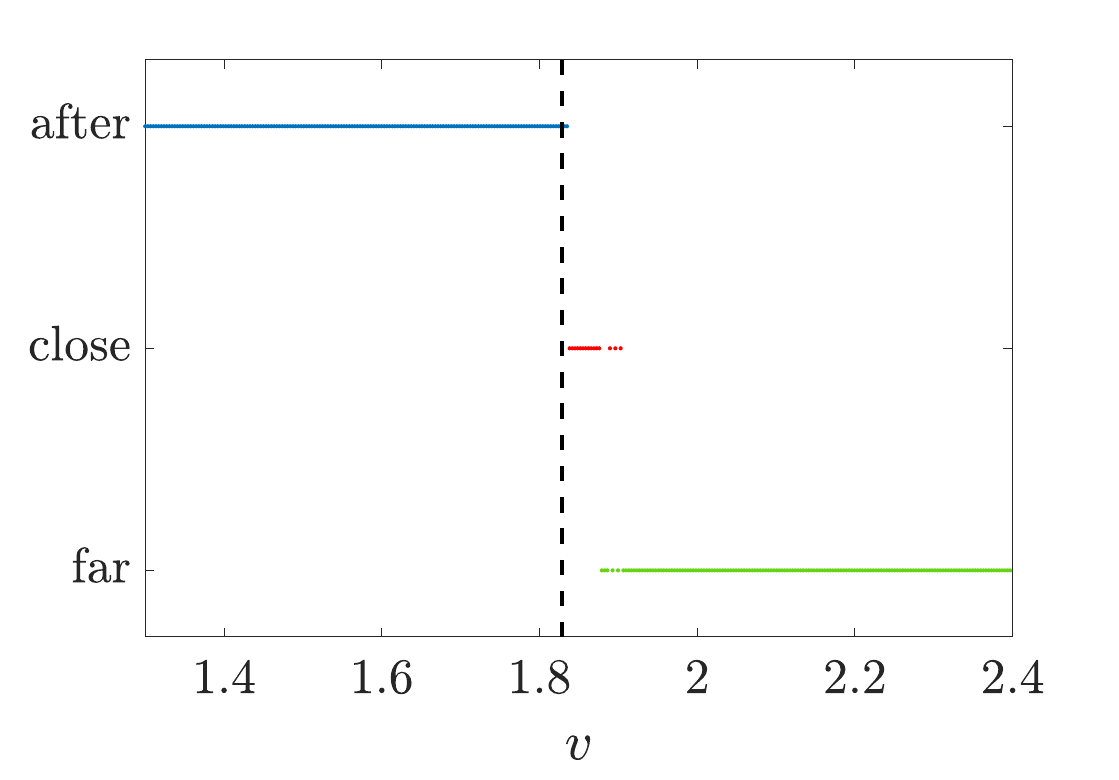}
    \caption{Mass-on-moving-belt}
    \label{fig_polar_01_log_mvm_MOB}
  \end{subfigure}
\begin{subfigure}[b]{0.35\textwidth}
    \includegraphics[width=\textwidth]{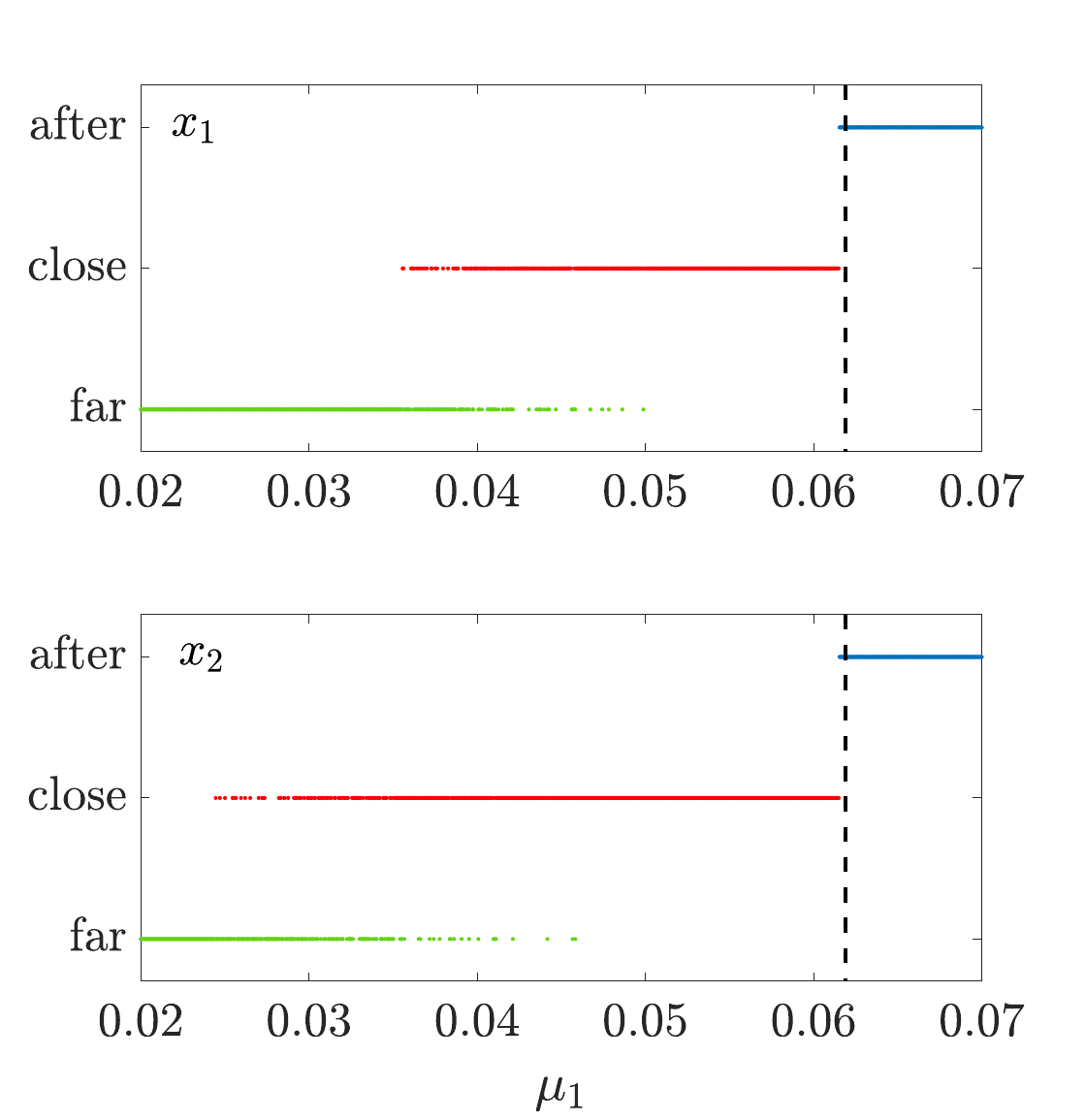}
    \caption{Van der Pol-Duffing oscillator}
    \label{fig_polar_01_log_mvm_VDP}
  \end{subfigure}
\begin{subfigure}[b]{0.35\textwidth}
    \includegraphics[width=\textwidth]{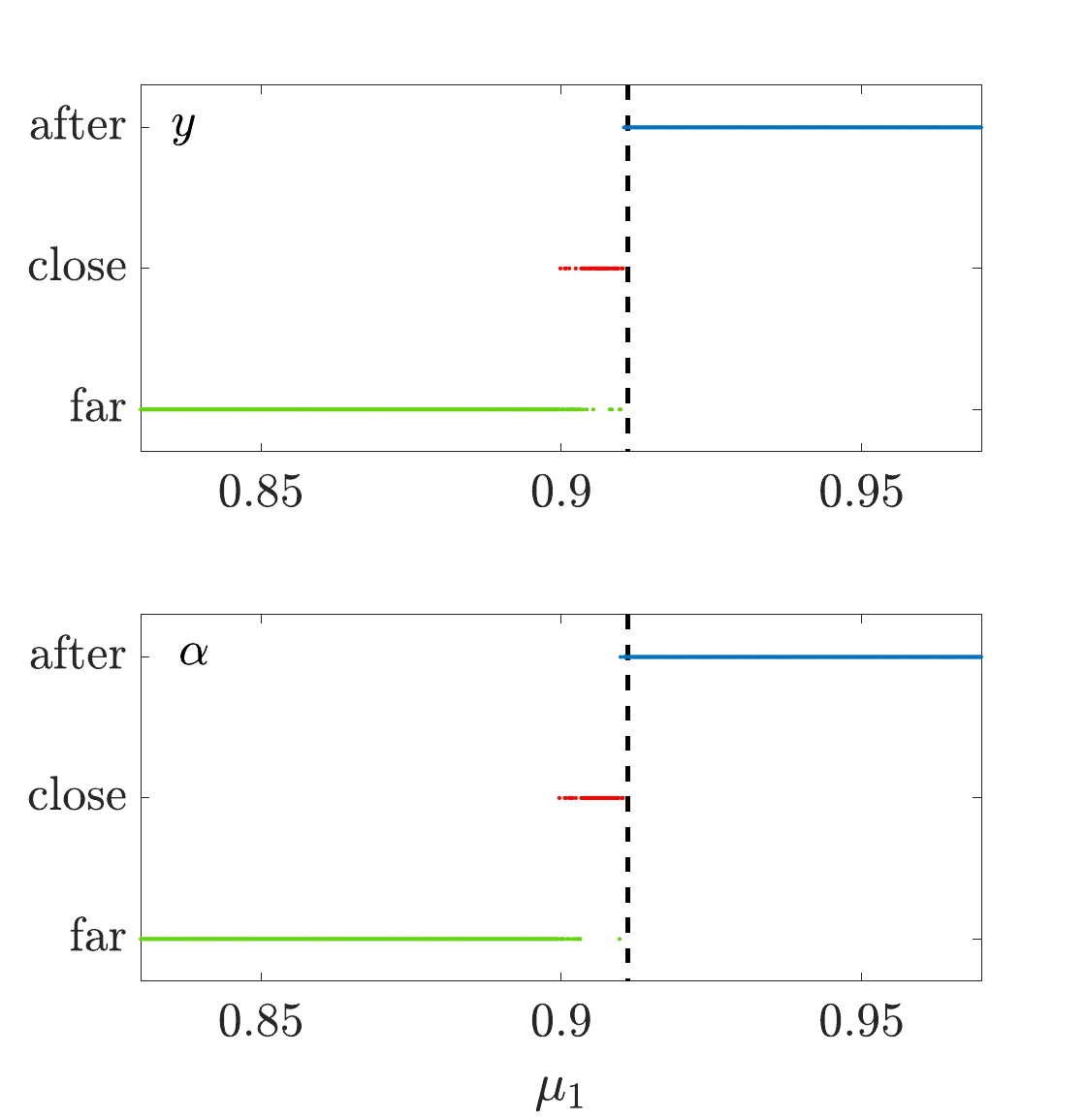}
    \caption{Pitch-and-plunge wing}
    \label{fig_polar_01_log_mvm_PnP}
  \end{subfigure}
\end{center}
\caption{\label{fig_polar_01_log_mvm_results}Prediction diagrams obtained by the CNN trained with data normalized in polar form and in logarithmic scale and then filtered with a moving mean. Each point corresponds to a different time series. Dashed lines indicate the fold bifurcation. The colors mark the predicted classification.
Color interpretation: green-{\em far}, red-{\em close}, blue-{\em after}.}
\end{figure}

The last case we considered is similar to the previous one, with an additional moving mean filter applied to the data to remove variations of the amplitude in polar coordinates during each complete oscillation.
The same filter was applied for the case not in logarithmic scale and produced no improvement (see Sect.~\ref{sect_res_pol_mvm}).
Conversely, in this case, it significantly enhanced the accuracy of the network.
All tested cases provided very good results.
In particular, trajectories of the nonlinearly damped oscillator were correctly classified for all the investigated range of $c_1$ (Fig.~\ref{fig_polar_01_log_mvm_sparsi_NLD}).
Excellent classification was also provided for the mass-on-moving-belt system (Fig.~\ref{fig_polar_01_log_mvm_MOB}); we remark that this was the only normalization that enabled the network to correctly classify {\em close} trajectories for this system.
The network was able to correctly distinguish between the three classes also for the van der Pol-Duffing oscillator (Fig.~\ref{fig_polar_01_log_mvm_VDP}) and pitch-and-plunge wing (Fig.~\ref{fig_polar_01_log_mvm_PnP}), using any of the coordinates. In terms of limitations, we note that the overlapping region is relatively large in all cases, especially for the pitch-and-plunge system (Table \ref{tab_overlapping}).

\section{Results with noisy signals}
\label{sect_noisy}

\begin{figure}[h]
\begin{center}
\setlength{\unitlength}{\textwidth}
\begin{subfigure}[b]{0.32\textwidth}
    \includegraphics[width=\textwidth]{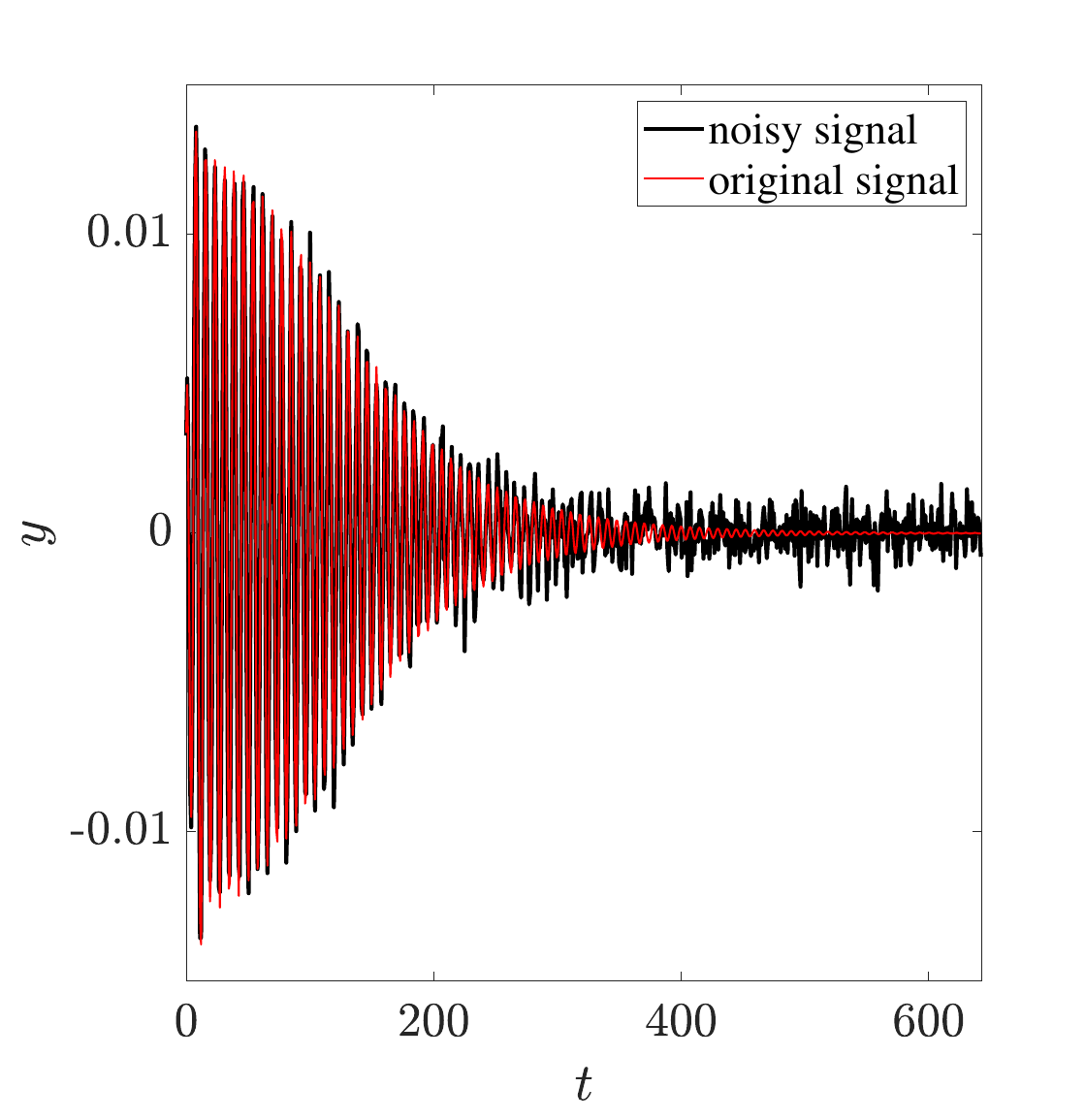}
    \caption{Original/noisy signals}
    \label{fig_clean_noise_original}
  \end{subfigure}
\begin{subfigure}[b]{0.32\textwidth}
    \includegraphics[width=\textwidth]{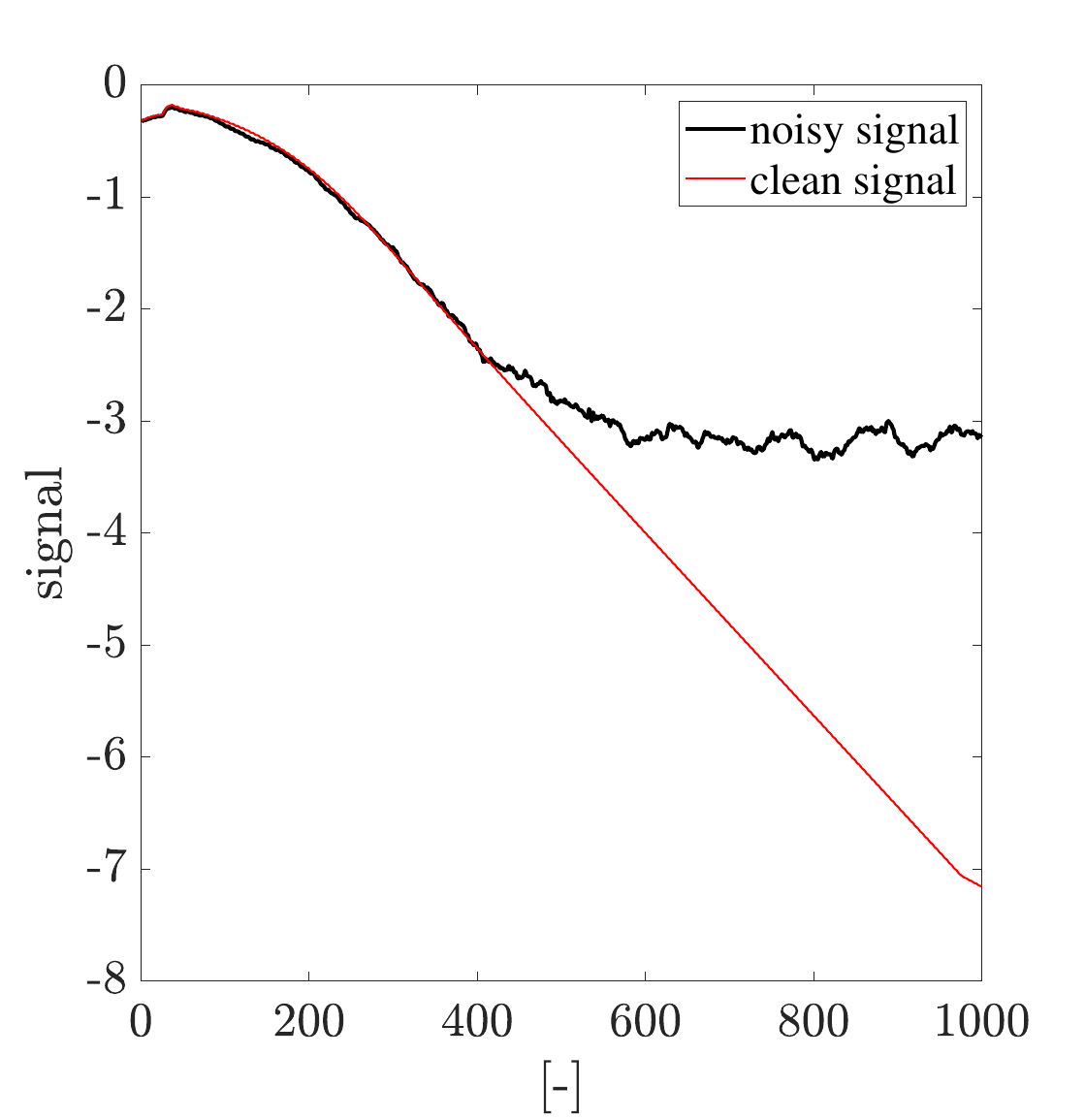}
    \caption{Polar-log transf. ($u=0.902$)}
    \label{fig_clean_noise_polar_PnP1}
  \end{subfigure}
\begin{subfigure}[b]{0.32\textwidth}
    \includegraphics[width=\textwidth]{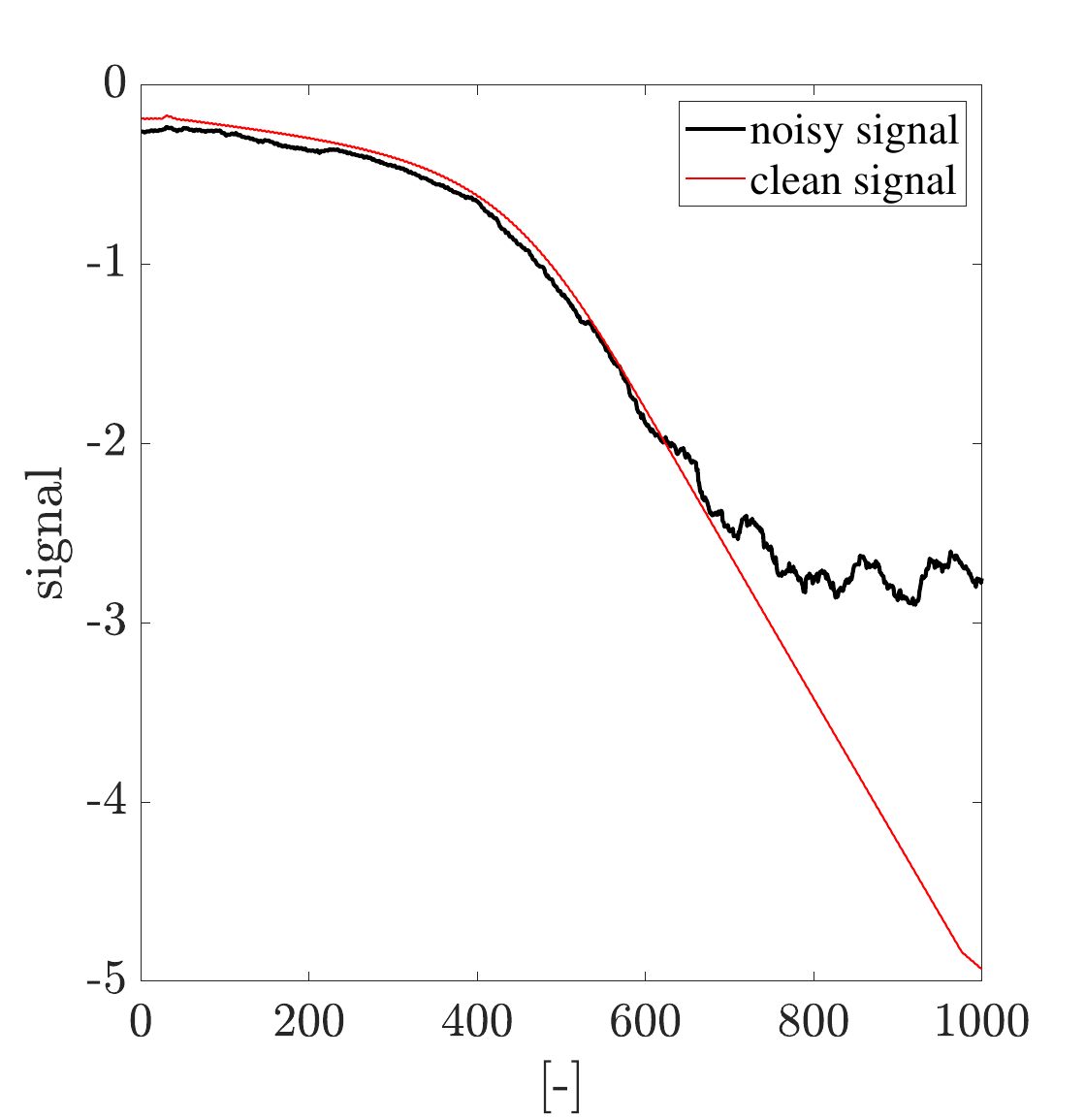}
    \caption{Polar-log transf. ($u=0.909$)}
    \label{fig_clean_noise_polar_PnP2}
  \end{subfigure}
\end{center}
\caption{\label{fig_noisy_signals}Comparison of the original time series obtained from direct numerical integration (red) and signal with added Gaussian noise (black) ($\text{SNR}=15$). (a) Signal from the pitch-and-plunge wing profile ($u=0.9024$); (b) signals in (a) after transformation into polar coordinates, normalization, and transformation into logarithmic scale; (c) signals as in (b) but with $u=0.90905$.}
\end{figure}

In this section, we investigate the performance of the trained network with noisy data. The analysis was limited to the most successful signal pre-processing strategy according to the analysis performed so far, namely polar transformation, normalization, transformation into natural logarithmic scale, and moving mean filtering (however, we note its poor performance in terms of overlapping region (Tab.~\ref{tab_overlapping})).
The same data utilized in the previous analysis was corrupted with a Gaussian noise using a signal-to-noise-ratio (SNR) of 15 dB. The corrupted signals were then post-processed as described in Sect.~\ref{sect_input}.
We note that the post-processing combines position and velocity signals. For this analysis, the noise was added independently to both position and velocity signals, which might generate artifact results; however, this is not expected to compromise the qualitative results of the analysis.
Noisy signals were obtained through the \texttt{MATLAB} built-in function \texttt{awgn(x,SNR,`measured')}, which automatically measures the signal average power and, accordingly, defines the power of the added noise. The signal power was measured independently for each signal, leading to higher absolute noise levels for signals with higher power.

A comparison of a signal with and without noise is depicted in Fig.~\ref{fig_clean_noise_original}.
The depicted signal refers to the pitch-and-plunge wing profile with $u=0.9024$, which is close to the fold bifurcation ($u^*=0.911$).
Despite the moderate SNR, it is observed that the original signal is practically unrecognizable for low-amplitude oscillations.

Figure \ref{fig_clean_noise_polar_PnP1} compares the clean and noisy signals provided to the network. The two signals are relatively close to each other at large amplitude, while the logarithmic scale enhances the difference between them at low amplitude.

\begin{figure}[h]
\begin{center}
\setlength{\unitlength}{\textwidth}
\begin{subfigure}[b]{0.35\textwidth}
    \includegraphics[width=\textwidth]{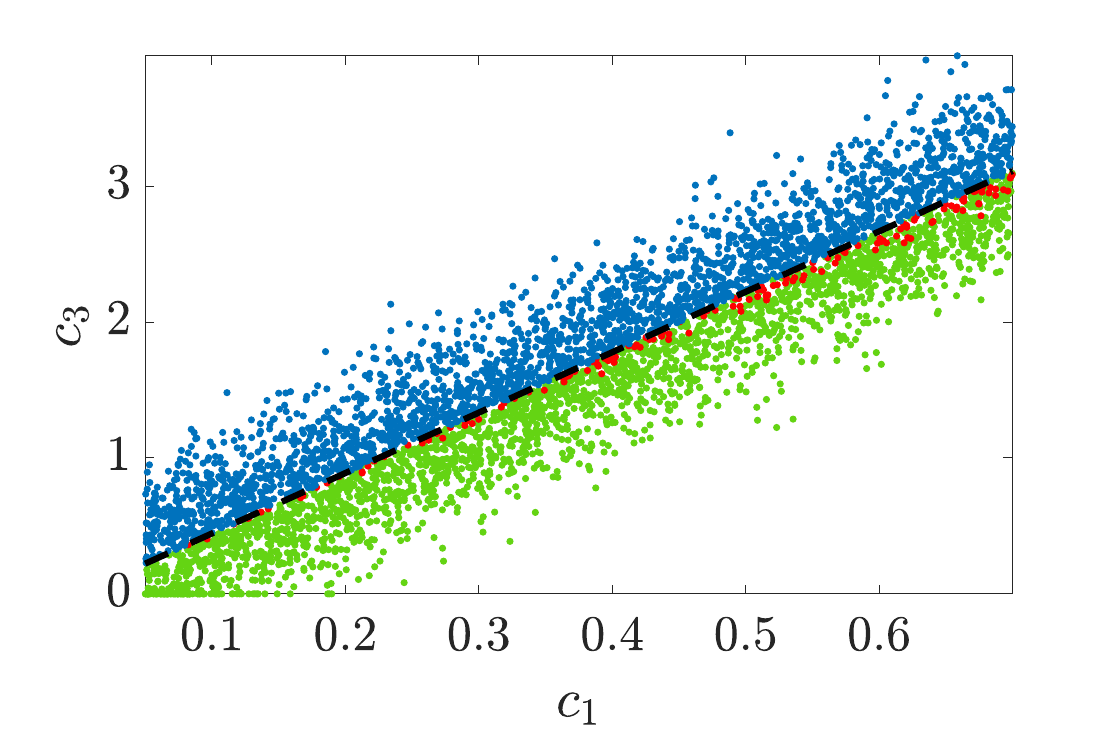}
    \caption{Nonlinear damping system}
    \label{fig_results_NLD_noisy}
  \end{subfigure}
\begin{subfigure}[b]{0.35\textwidth}
    \includegraphics[width=\textwidth]{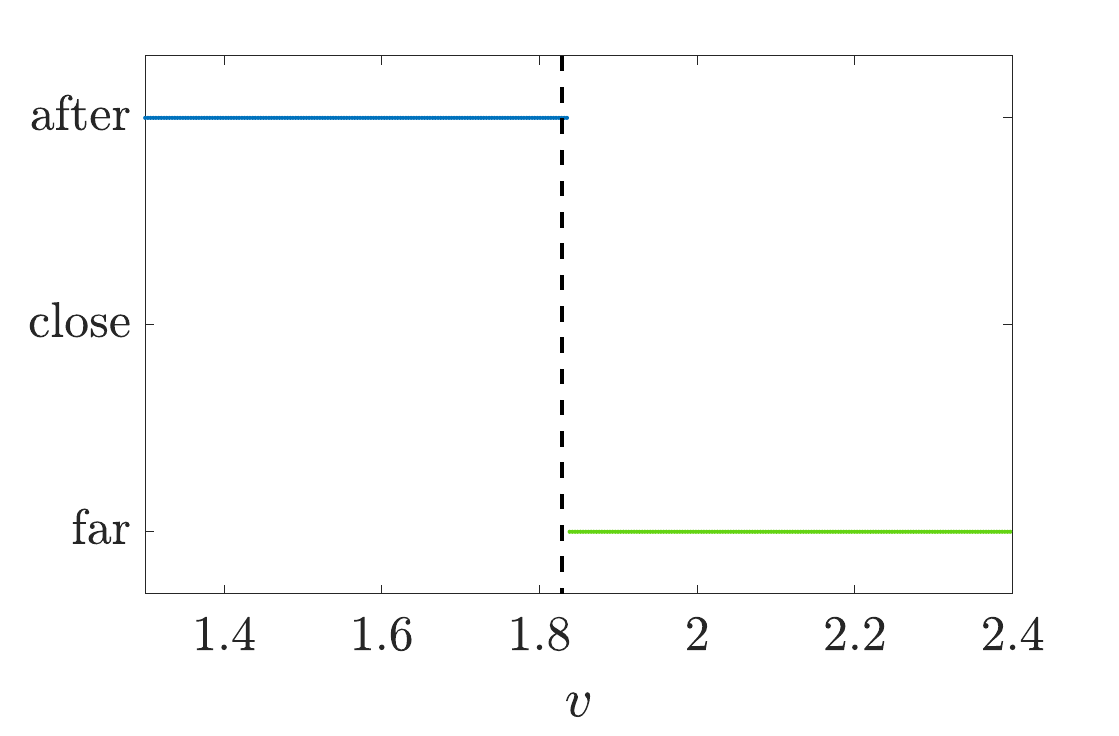}
    \caption{Mass-on-moving-belt}
    \label{fig_results_MOB_noisy}
  \end{subfigure}
\begin{subfigure}[b]{0.35\textwidth}
    \includegraphics[width=\textwidth]{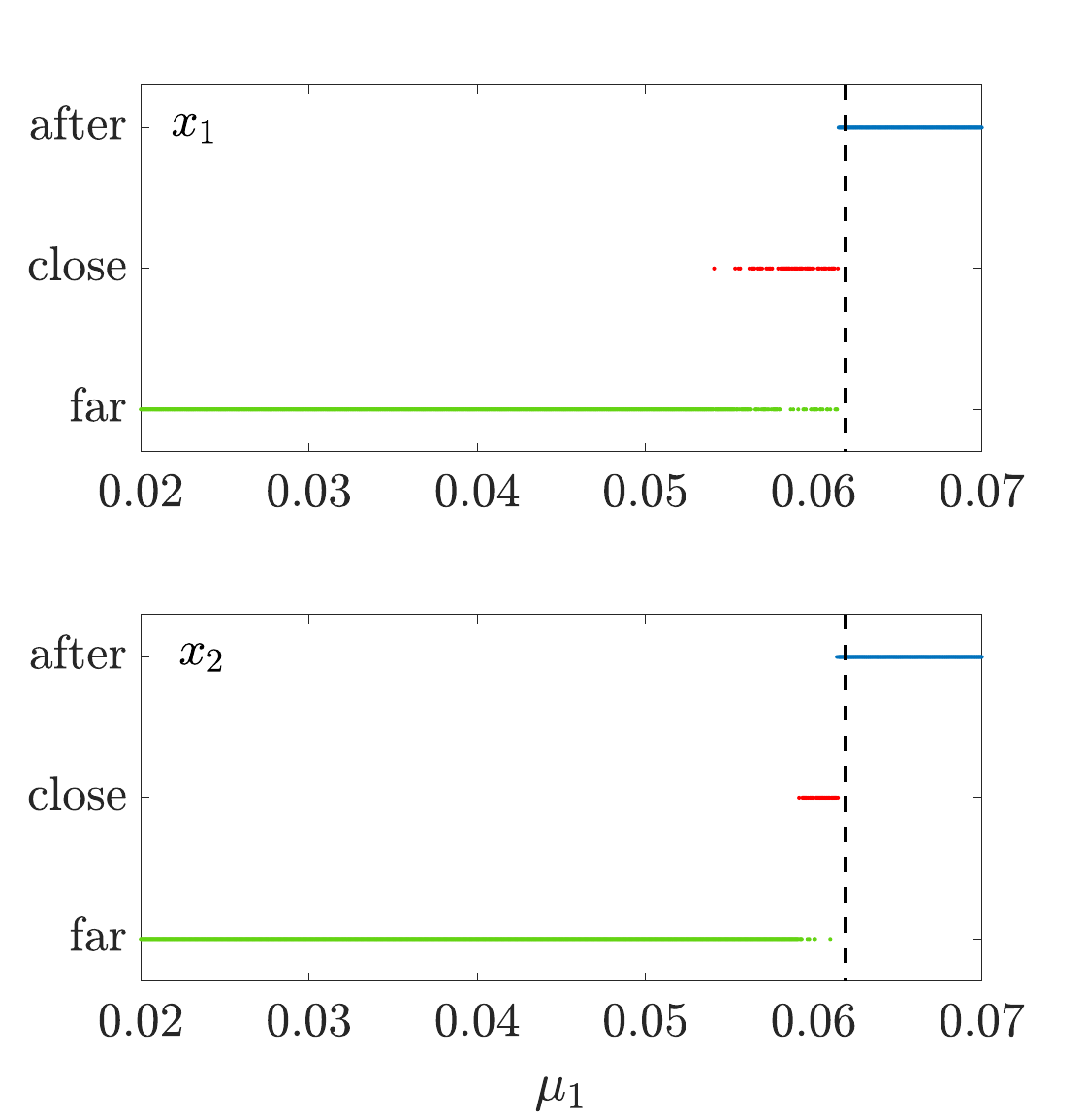}
    \caption{Van der Pol-Duffing oscillator}
    \label{fig_results_VDP_noisy}
  \end{subfigure}
\begin{subfigure}[b]{0.35\textwidth}
    \includegraphics[width=\textwidth]{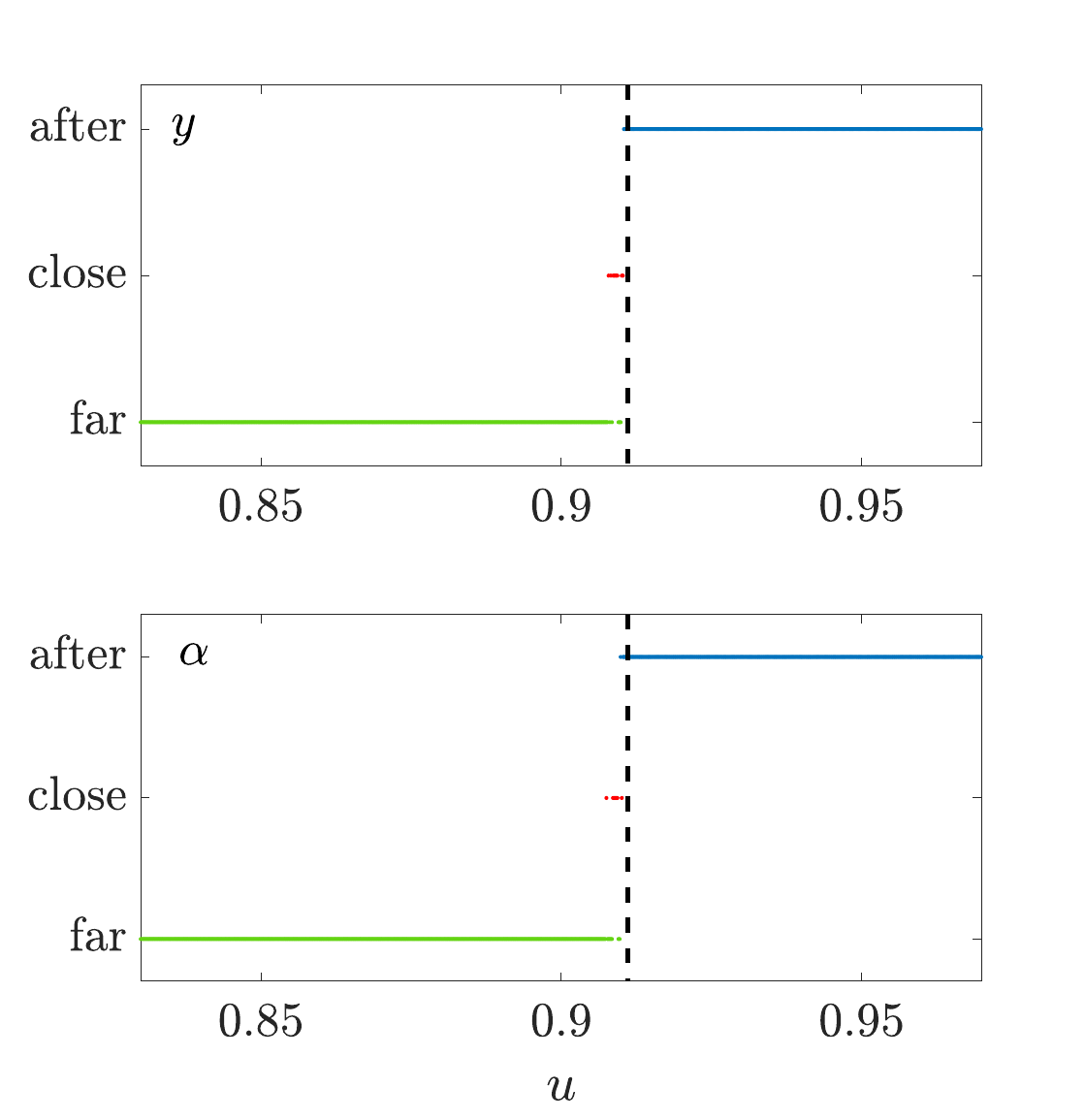}
    \caption{Pitch-and-plunge wing}
    \label{fig_results_PnP_noisy}
  \end{subfigure}
\end{center}
\caption{\label{fig_results_noisy}Prediction diagrams obtained by the CNN trained with data normalized in polar form and logarithmic scale and then filtered with a moving mean.
A Gaussian noise with an SNR of 15 was added to the test data (not the training data). Each point corresponds to a different time series. Dashed lines indicate the fold bifurcation. The colors mark the predicted classification.
Color interpretation: green-{\em far}, red-{\em close}, blue-{\em after}.}
\end{figure}

For time series of the same system and in the same parameter range of the training data (the oscillator with nonlinear damping with $c_1=0.5$), but with added noise, the accuracy of the NN decreased to 72.3~\% (it was 100~\% in the absence of noise). Testing the network for the same system, but with $c_1=0.1$, provided an accuracy of 71.5~\% (also 100~\% without noise). In both cases, all errors are relative to {\em close} trajectories erroneously classified as {\em far}.

Results of the analysis for the other systems are provided in Fig.~\ref{fig_results_noisy}.
As a general trend, we note that the range of solutions classified as {\em close} is significantly reduced. This phenomenon occurs for all studied systems. In particular, no solution is classified as {\em close} for the mass-on-moving-belt system. For all systems, trajectories that are no longer classified as {\em close} are now classified as {\em far}. The overlapping between {\em far} and {\em close} regions is also very large (Table \ref{tab_overlapping}).

Except for the mass-on-moving-belt system, this result cannot be strictly considered erroneous because of the arbitrary distinction between {\em far} and {\em close} cases. Nevertheless, from a practical perspective, it is unsatisfactory because it makes the prediction non-conservative concerning the fold bifurcation identification.

Because of the black-box nature of the method, it is not trivial to understand why the NN reduces the range of {\em close} solutions in the presence of noise. Looking at Figs.~\ref{fig_norm_far_polar_log} and \ref{fig_norm_close_polar_log}, it can be recognized that the NN distinguishes {\em far} and {\em close} trajectories from the bump present in {\em close} trajectories. Figure \ref{fig_clean_noise_polar_PnP1} depicts a trajectory classified either as {\em close} or {\em far} in the absence or presence of noise, respectively. Conversely, Fig.~\ref{fig_clean_noise_polar_PnP2} shows a trajectory closer to the fold, classified as {\em close} in both cases. We hypothesize that the noise reduces the slope of the signal, making the bump less visible. However, for trajectories sufficiently close to the fold, as the one in Fig.~\ref{fig_clean_noise_polar_PnP2}, the bump is pronounced enough to be recognized by the NN even in the presence of noise, which classifies the trajectory as {\em close}. Nevertheless, it is hard to validate this hypothesis. Additionally, we note that, since the noise level is proportional to the signal level, {\em close} trajectories present a relatively high noise level because their power level is initially high, making noise significant for the second part of the signal when the vibration amplitude decreases. This effect might contribute to the wrong classification of {\em close} trajectories.

For a more comprehensive assessment of the performance of the NN in the presence of noise, we trained another CNN with noisy data, adding the noise exactly as for test data. Since the noise increases the diversity of training data, we doubled the amount of provided data to a total of 6000 trajectories, 2000 for each class. Additionally, the range of {\em close} trajectories was extended from $c_3\in c_3^*\cdot\left(0.9,\,0.995\right)$ (see Sect.~\ref{sec_training}) to $c_3\in c_3^*\cdot\left(0.85,\,0.995\right)$, with the aim of facilitating the recognition of {\em close} trajectories. The hyperparameters of the NN were the same as in the previous cases (see Sect.~\ref{sect_architecture}).

The NN trained with noisy data provided a 100~\% accuracy for data in the same parameter range as the training data, confirming the 100~\% accuracy also for data from the same system, but with $c_1=0.1$. This result is particularly surprising and already represents a significant improvement compared to the NN trained with clean data.
It implies that, despite the presence of noise, the time series from the three categories are still different enough for the NN to correctly classify them with no error. Also in this case, if the gap between \emph{far} and \emph{close} categories was smaller, the accuracy would probably decrease. We note that the time series utilized for the validation are obtained for exactly the same parameter values as for the case without noise. The specific $c_3$ values utilized are illustrated in Fig.~\ref{fig_data_info}

\begin{figure}[h]
\begin{center}
\setlength{\unitlength}{\textwidth}
\begin{subfigure}[b]{0.35\textwidth}
    \includegraphics[width=\textwidth]{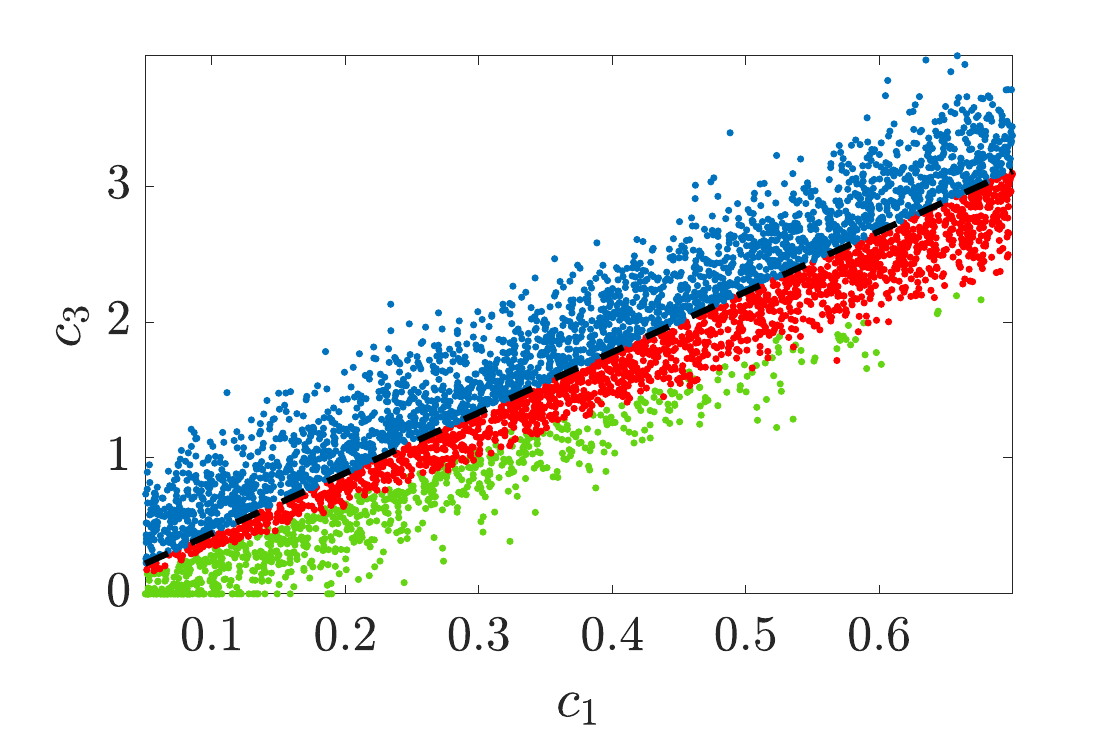}
    \caption{Nonlinear damping system}
    \label{fig_results_NLD_Tnoisy}
  \end{subfigure}
\begin{subfigure}[b]{0.35\textwidth}
    \includegraphics[width=\textwidth]{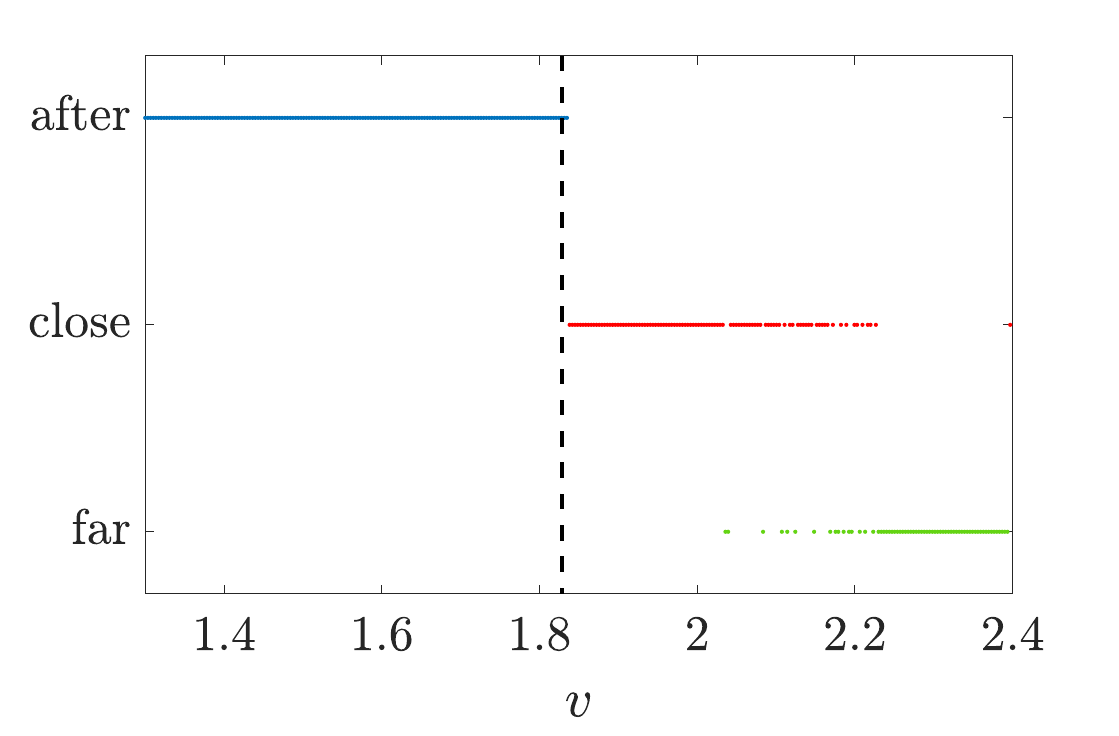}
    \caption{Mass-on-moving-belt}
    \label{fig_results_MOB_Tnoisy}
  \end{subfigure}
\begin{subfigure}[b]{0.35\textwidth}
    \includegraphics[width=\textwidth]{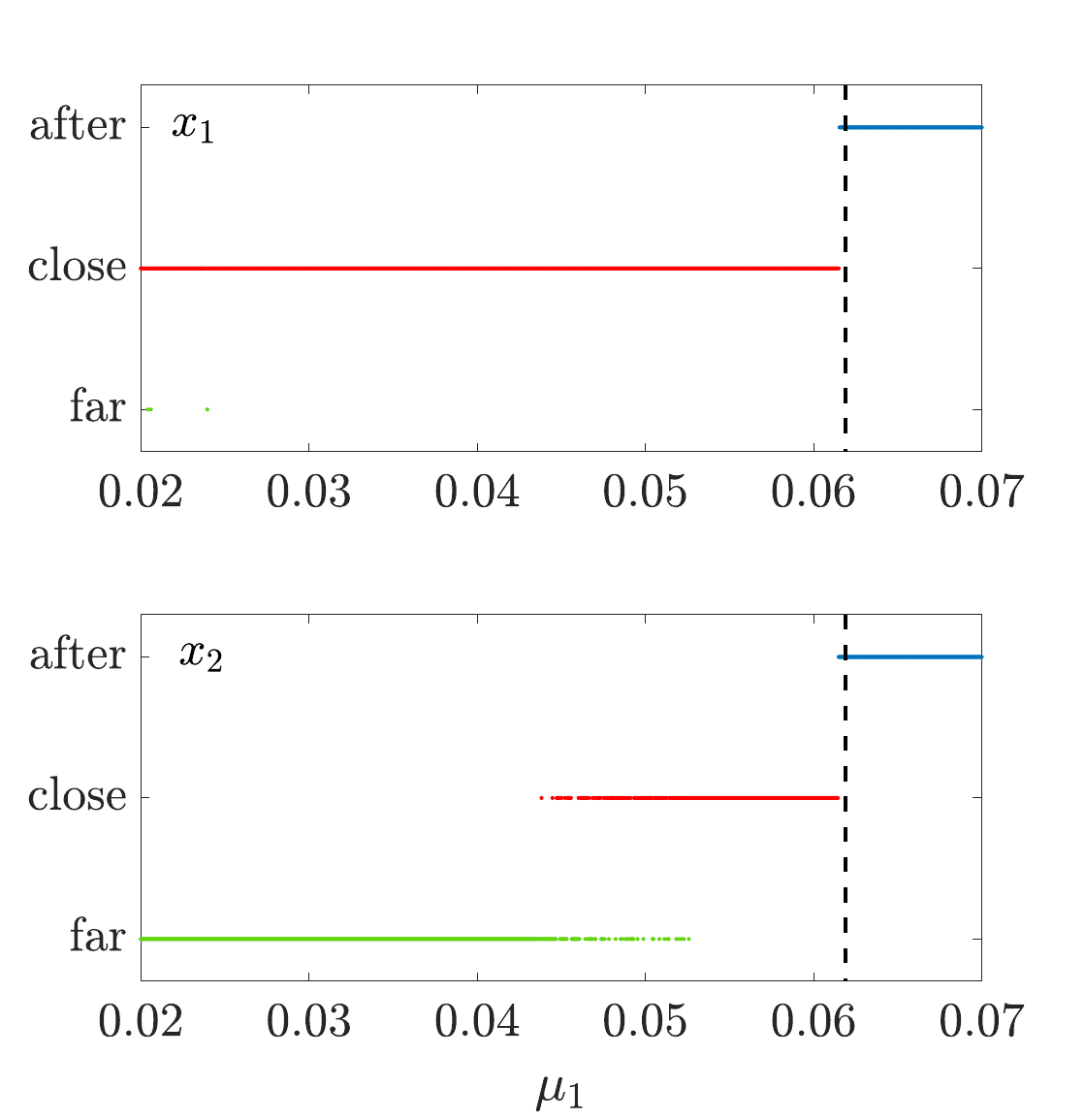}
    \caption{Van der Pol-Duffing oscillator}
    \label{fig_results_VDP_Tnoisy}
  \end{subfigure}
\begin{subfigure}[b]{0.35\textwidth}
    \includegraphics[width=\textwidth]{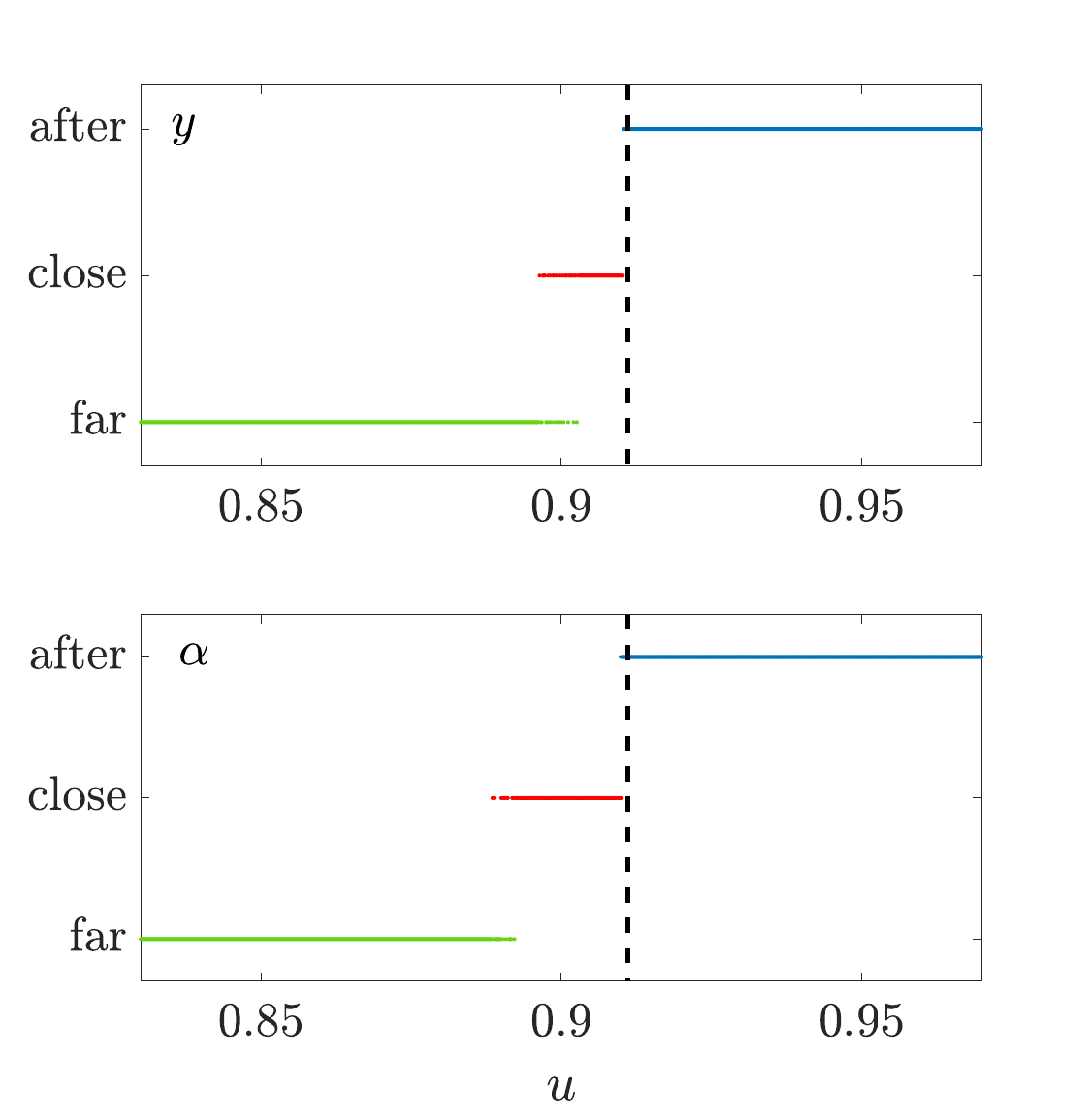}
    \caption{Pitch-and-plunge wing}
    \label{fig_results_PnP_Tnoisy}
  \end{subfigure}
\end{center}
\caption{\label{fig_results_Tnoisy}Prediction diagrams obtained by the CNN trained with data normalized in polar form and logarithmic scale and then filtered with a moving mean.
A Gaussian noise with an SNR of 15 was added to the training and test data. Each point corresponds to a different time series. Dashed lines indicate the fold bifurcation. The colors mark the predicted classification.
Color interpretation: green-{\em far}, red-{\em close}, blue-{\em after}.}
\end{figure}

Figure \ref{fig_results_Tnoisy} illustrates the predictions obtained by the NN for trajectories regarding the other systems. For the oscillator with nonlinear damping (Fig.~\ref{fig_results_NLD_Tnoisy}), results are satisfactory for the whole range of parameter values investigated. Also regarding the mass-on-moving-belt system (Fig.~\ref{fig_results_MOB_Tnoisy}), the NN was able to identify trajectories in all classes, with some overlapping between {\em close} and {\em far} trajectories. The same occurs for the pitch-and-plunge airfoil (Fig.~\ref{fig_results_PnP_Tnoisy}), with data from any coordinates. For this system, the overlapping region is particularly small (Table \ref{tab_overlapping}), which marks an improvement also with respect to the analysis with clean data. The only system for which the NN did not work well is the van der Pol-Duffing oscillator, where predictions obtained from $x_1$ confused almost all {\em far} trajectories as {\em close}. Interestingly, this error is opposite to the one done by the NN trained with clean data.

At last, the performance of the NN trained with noisy data was tested on trajectories without noise. The NN was able to correctly predict the classification for all the systems, including both coordinates of the van der Pol-Duffing oscillator. The relative figure is omitted for the sake of brevity.

Considering the limited data utilized for the training and the absence of specific measures for filtering the noise in the data, we believe that the results are encouraging concerning the effectiveness of the proposed methodology.

\section{Discussion and conclusions} \label{sect_discussion}

The performed analysis demonstrates that a properly trained neural network can recognize a trajectory passing close to a fold bifurcation.
However, in order for the network to capture the important features of such trajectories and provide correct classification for different systems, it is essential that training data are properly normalized.
This step eliminates inessential information from the data, such as oscillation frequency, allowing the network to focus on important ones, like the decrement of oscillation amplitude.
The results obtained with this method are remarkable.
In fact, a network trained with a single, very simple system performed very well on three different mathematical systems modeling mechanical systems of engineering relevance, thus exhibiting excellent extrapolation capabilities.
The most important step of the normalization was the transformation in polar coordinates, which eliminates almost completely information about oscillation frequency, thus highlighting amplitude.
Applying a logarithmic scaling, providing a physically more relevant scale, further improved the network performance.
Finally, introducing a moving mean filter and eliminating any residual information about variations in single oscillation periods enhanced even more correct classification capabilities.
Every step of data normalization is related to the physical understanding of the problem, and it is not standard for signal processing, except for the moving mean.

The present analysis does not rule out that a classical data normalization (with no physical basis) is sufficient for an NN to classify trajectories in different systems. However, it most probably requires that the training data involves several systems.

We observe a significant overlap in the regions where trajectories are classified as either {\em far} or {\em close}. This overlap arises primarily because some initial conditions result in dynamical behavior that is minimally affected by the fold, while others produce trajectories that are more significantly distorted by it. This phenomenon was clearly demonstrated in \cite{habib2023predicting}.
Furthermore, we note that the overlapping region is larger for the best-performing data pre-processing strategy. In contrast, pre-processing that involves only transformation into polar form, without applying a logarithmic scale, results in moderate overlap. This observation suggests that the benefit of the logarithmic transformation cannot be conclusively established.
Moreover, this might reflect an inherent limitation of the approach, which requires a compromise between achieving clear boundaries at the cost of higher misclassification rates, or tolerating fuzzier boundaries with improved accuracy.
To determine the optimal pre-processing strategy and to minimize fuzziness, more comprehensive testing is needed, including training on a diversified set of systems, and the analysis of various NN architectures.

Regarding the limitations of the approach, we remark that the range of the initial conditions significantly affects the network's accuracy.
In fact, referring to the best-performing case, the network has to distinguish between scenarios similar to those in Figs.~\ref{fig_norm_far_polar_log} and \ref{fig_norm_close_polar_log}; Fig.~\ref{fig_norm_close_polar_log} might be distorted if large initial conditions move down the bump, confusing the CNN.

%Additionally, we note that in the studied cases, the simulations were run until the systems reached extremely small oscillation amplitudes (about $e^{-10}$, see Fig.~\ref{fig_norm_far_polar_log}). In the case of a real system, and in the presence of noise, it is unreasonable to expect the system to reach such small amplitudes, which would modify the numerical values in the {\em far} and {\em close} cases, potentially generating wrong classifications. In particular, {\em far} and {\em close} trajectories might be classified as {\em after}.

These potential limitations can be overcome by increasing the heterogeneity of training data, for example, by considering different systems.
Another way to solve the issues is to use a sliding window convolutional neural network, which would analyze the signal by dividing it in time window frames. This way, the network would provide a list of outputs for each signal.
This output list could then be further analyzed to obtain a single classification output.

In terms of possible applications, this approach is implementable only on systems that can be forced to undergo, or naturally experience, large perturbations, such as, for example, an adaptive cruise control car-following system experiencing a sudden cut-in \cite{milanes2016handling}. Although this limitation is common for most methods available for fold bifurcation identification \cite{lim2011forecasting, habib2023predicting}, it still significantly limits the range of applications. This aspect should be addressed in future studies.
Besides, the bifurcation parameter is assumed quasi-static, i.e., constant for each trajectory. Trajectories with varying bifurcation parameters might include additional challenges.

Another limitation worth mentioning is that, in the bistable region, only trajectories approaching the periodic solutions were considered. Future developments of this study will include the analysis of trajectories converging to the equilibrium solution in the bistable region.

The trained NN exhibited a sensitivity to noise in the signals, which resulted in a degradation of accuracy; however, this sensitivity was significantly mitigated by including noise also in the training data.

This study does not investigate other machine learning algorithms which might provide similar or better results.
For instance, a random forests architecture might also lead to good results.
Alternatively, two autoencoders could be trained, one on {\em far} and one on {\em after} trajectories, in order to identify {\em close} ones.
Additionally, the convolutional layers utilized are relatively standard and not specifically designed for the accomplished task. A thorough analysis of the hyperparameters and alternative network architectures might provide significant improvement.

In the analysis, training data was relatively little and purposely limited to a single simple system.
However, training the network with more data obtained from diverse systems, including experimental ones, might enable it to correctly classify trajectories also in real-life problems.

Regarding the implementation of the method to other systems, a clear limitation is in the bifurcation diagram: the method can be applied only to systems having a fold bifurcation with properties similar to those of the systems discussed here. Besides, since the division between {\em close} and {\em far} trajectories is arbitrary, special care should be taken while applying the method to a new system.

At last, we note that this study is limited in its scope, as the network only distinguishes between trajectories {\em far}, {\em close}, or {\em after} the fold and does not try to estimate the position of the fold, as other deterministic methods do \cite{lim2011forecasting, habib2023predicting}.
While training the network to identify the position of the fold based on some time series computed before the fold is probably very challenging, it is likely feasible to add more classes to distinguish, for example, between {\em close} and {\em very close} trajectories.
Additionally, including trajectories starting at lower amplitude, it might be possible to distinguish also between trajectories far from, close to, or after the Andronov-Hopf bifurcation (if present).
However, these further developments probably require more diverse and abundant training data.

In general, we believe that this study advances the development of an algorithm for fold bifurcation prediction applicable to real-world problems, which could serve as a valuable tool in engineering, particularly for safety monitoring of dynamical systems. However, further research is needed to address the limitations discussed and to make the method practically implementable.

\section*{Data availability}
\noindent 
The code utilized for generating the results presented in the paper is publicly available at \href{https://gdrg.mm.bme.hu}{https://gdrg.mm.bme.hu}.

\section*{Acknowledgments}
\noindent 
The authors thank Dominik Wenesz for insightful discussions during the realization of the paper.

\section*{Funding}
\noindent 
The research reported in this paper has been supported by Project no.
TKP-6-6/PALY-2021 provided by the Ministry of Culture and Innovation of
Hungary from the National Research, Development and Innovation Fund,
financed under the TKP2021-NVA funding scheme and by the National Research, Development and Innovation Office (Grant no. NKFI-134496).

%\begin{figure}
%  \centering
%  \begin{subfigure}[b]{0.35\textwidth}
%    \includegraphics[width=\textwidth]{1DOF_NES.pdf}
%    \caption{}
%    \label{fig:compound}
%  \end{subfigure}
%  \begin{subfigure}[b]{0.55\textwidth}
%      \includegraphics[width=\textwidth]{stifness.pdf}
%      \caption{}
%      \label{fig:stifness}
%  \end{subfigure}
%  \caption{The primary system with an NES (a) and the connecting stiffness characteristic of conventional NESs and of the proposed periodically extended NES (b).}
%  \label{fig:system_stiffness}
%\end{figure}

    \bibliography{references}
    \setcounter{figure}{0}  % reset counter 
    
%    \appendix
%    \newpage
%    
%\section{...}
%
% 
%\section{...}

\end{document}